\documentclass[acmtog,accepted]{acmart_tmp}
\acmSubmissionID{670}

\usepackage[normalem]{ulem}
\usepackage{booktabs} 
\usepackage{xcolor}

\newcommand\edited[1]{{#1}}

\usepackage[ruled]{algorithm2e} 
\usepackage[noabbrev, capitalise]{cleveref}
\usepackage{caption}
\usepackage{subcaption}
\usepackage{dirtytalk}
\usepackage{selectp}
\usepackage{booktabs}
\usepackage{multirow}
\usepackage{fontawesome5}
\usepackage{mathtools}

\SetAlFnt{\small}
\SetAlCapFnt{\small}
\SetAlCapNameFnt{\small}
\SetAlCapHSkip{0pt}

\newcommand\numberthis{\addtocounter{equation}{1}\tag{\theequation}}

\newcommand{\algfull}{Conditional Adversarial Latent Models}
\newcommand{\alg}{CALM}

\acmJournal{TOG}

\copyrightyear{2023} 
\acmYear{2023} 

\begin{document}
\title{\alg: \algfull~ for Directable Virtual Characters}

\author{Chen Tessler}
\affiliation{%
 \institution{NVIDIA}
 \country{Israel}}
\email{ctessler@nvidia.com}
\author{Yoni Kasten}
\affiliation{%
 \institution{NVIDIA}
 \country{Israel}
}
\author{Yunrong Guo}
\affiliation{%
\institution{NVIDIA}
\country{Canada}}
\author{Shie Mannor}
\affiliation{%
 \institution{NVIDIA}
 \country{Israel}
}
\affiliation{%
 \institution{Technion Institute of Technology}
 \country{Israel}
}
\author{Gal Chechik}
\affiliation{%
 \institution{NVIDIA}
 \country{Israel}}
 \affiliation{%
 \institution{Bar-Ilan University}
 \country{Israel}}
\author{Xue Bin Peng}
\affiliation{%
 \institution{NVIDIA}
 \country{Canada}
}
\affiliation{%
 \institution{Simon Fraser University}
 \country{Canada}
}

\begin{abstract}
    \edited{In this work, we present \algfull~ (\alg), an approach for generating diverse and directable behaviors for user-controlled interactive virtual characters. Using imitation learning, \alg~ learns a representation of movement that captures the complexity and diversity of human motion, and enables direct control over character movements. The approach jointly learns a control policy and a motion encoder that reconstructs key characteristics of a given motion without merely replicating it. The results show that \alg~ learns a semantic motion representation, enabling control over the generated motions and style-conditioning for higher-level task training. Once trained, the character can be controlled using intuitive interfaces, akin to those found in video games.}

\end{abstract}

%
%


%
%


\begin{teaserfigure}
  \includegraphics[width=\textwidth]{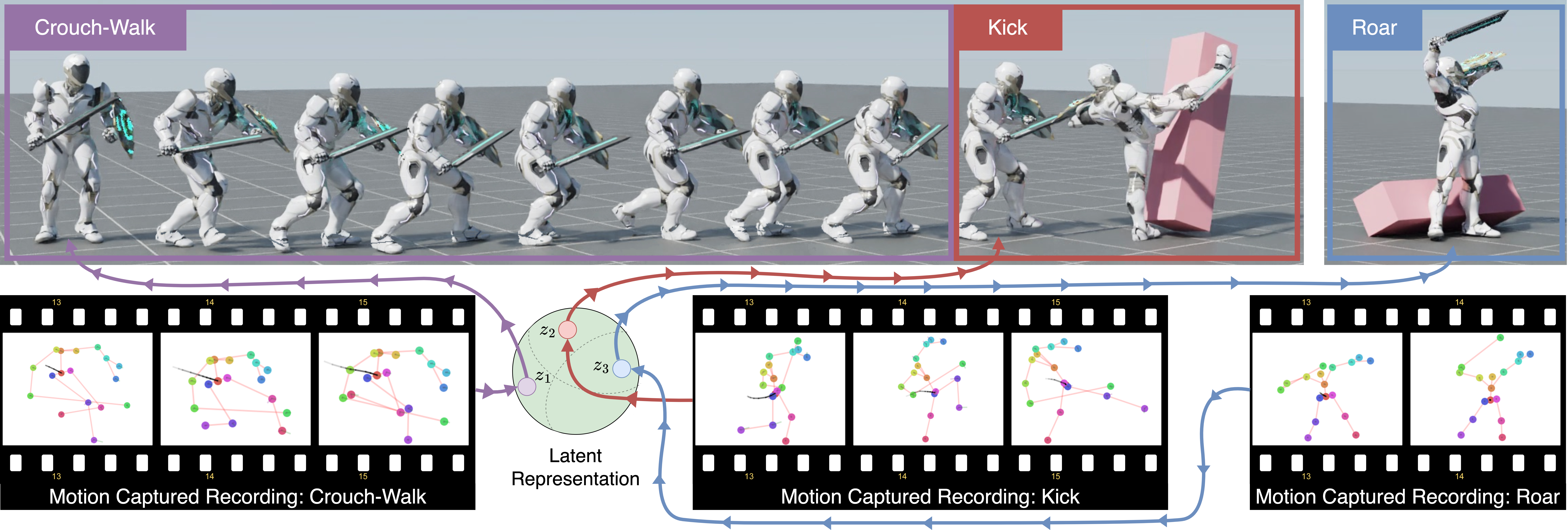}
  \caption{Our framework enables users to direct the behavior of a physically simulated character using demonstrations encoded in the form of low-dimensional latent embeddings of motion capture data. In this example, the character is instructed to crouch-walk towards a target, kick when within range, and finally raise its arms and celebrate.}
  \label{fig:teaser}
\end{teaserfigure}

\maketitle

\section{Introduction}
Virtual environments and interactive characters have become more prevalent and user-friendly, but creating realistic and diverse behaviors for these virtual agents remains a challenge due to the complexity of human motion. To create interactive and immersive experiences, virtual agents must adapt to different environments and user inputs in a life-like manner, and this requires the ability to perform a wide range of behaviors on demand. To that end, we need to develop control models that can generate complex and realistic behaviors, while taking into account the properties of the environment.
For example, in virtual reality games, players that interact with virtual characters and objects expect them to behave realistically. This includes responding to user commands and navigating through virtual environments. When virtual agents fail to respond naturally to user input, it can disrupt the immersive experience.

Recent advancements in machine learning and access to high-quality human motion capture data have led to the development of control policies that can replicate human behavior. Early studies in this field, such as \cite{peng2018deepmimic}, focused on imitating single motion clips. However, as each motion is learned using an independent controller, it does not effectively scale. Later research \cite[ASE]{peng2022ase} aimed to improve the diversity of generated motion by learning a latent-conditioned controller and maximizing a mutual information objective. When trained on a dataset of diverse motions, distinct behaviors emerged, however at the cost of losing the ability to control the generated motion.

Building on these previous works, we present \algfull~ (\alg), a method for learning a representation of movement that captures the complexity and diversity of human motion, while also providing a directable interface for controlling a character's movements. Given raw motion capture recordings, illustrated in \cref{fig:teaser}, \alg~ encodes them into a latent representation. \alg~ further decodes a given latent vector into a skill for a physically simulated character, enabling it to perform high-level tasks while being conditioned on a desired motion. As can be seen in \cref{fig:teaser}, without any further training, using only predefined raw motion data, the agent can solve challenging tasks using a set of desired skills. A key benefit of our framework is that the policy does not need to precisely replicate the original reference motion. Instead, it has the flexibility to produce diverse movements, as long as they resemble the distributional characteristics of the particular motion clip. This enables the policy to deviate from the motion data and generate new and diverse behaviors that appear natural even though they were not explicitly depicted in the original dataset.

Compared to prior work, we focus on unsupervised techniques. The data is unlabeled and we do not assume prior knowledge of semantic connections between motions. We present an end-to-end method that jointly learns a meaningful semantic representation of skills and a control policy capable of producing the selected skills. While previous work used language to derive semantic connections \cite{juravsky2022padl}, we demonstrate that semantic meaning can be directly inferred from the motions and similarity is determined in the context of the policy reproducing the motions. This ability to direct the character's motions then enables tasks to be solved using human-like control and without further re-training, demonstrating \alg's ability to generate interactive and directable virtual characters.
\edited{To conclude, our contributions are as follows:}
\begin{enumerate}
    \item \edited{We present a method for jointly training a generative motion controller and a motion encoder, from unlabeled motion capture data. The resulting policy can be directed to generate a motion $M$ via its encoding $z = E(M)$.}

    \item \edited{We introduce precision training, a way to reuse the pre-trained policy and leverage similarity within the learned latent space to enable control over the produced motion when solving high-level tasks, such as locomotion. }

    \item \edited{Finally, we show that combining steps 1 and 2 enables the design of simple FSMs to solve tasks without further training or meticulous design of reward functions or termination conditions.}
\end{enumerate}

\section{Related Work}

We aim to learn rich and reusable skill representations, from diverse unlabeled data, for character control. The field of data-driven motion generation can be broadly divided into kinematic models and physics-based models. Kinematic models directly generate pose trajectories without explicitly considering physical constraints, while physics-based models often use a physically simulated environment to enforce realistic dynamics during motion generation.

We differentiate between two methods of control for parametric models: direct prediction and latent/language-based control. In direct prediction, the model learns to generate motions for a pre-defined task, while latent and language-based control involves learning a generative model for motions. As our focus is on creating interactive, controllable characters, we build upon the literature of physics-constrained generation and focus on latent-based control.

\subsection{Physics-constrained motion generation}

In physics-constrained generation, the model predicts motor actuations. These are fed into the simulator, which then produces the next state by emulating the character's motion while adhering to the various governing laws (such as gravity and friction). Recent advances in deep learning, specifically deep reinforcement learning, have enabled a leap forward in generation quality. Such an example is DeepMimic \cite{peng2018deepmimic}, in which they defined a motion-tracking reward that is used for training a policy to imitate specific motions. However, these schemes focused on imitating single, pre-defined, motion clips.

\paragraph{Direct prediction.} In direct prediction, the goal is to directly learn to solve a downstream task. For instance, the agent may be tasked with reaching a goal location. Here, AMP \cite{peng2021amp} combines adversarial imitation learning \cite[GAIL]{ho2016generative} with classic RL. The agent attempts to balance between maximizing the task reward, whilst successfully fooling a discriminator. However, when the demonstration data distribution does not fit the task, the resulting behavior is unsatisfactory, and the resulting model is unable to generalize to new tasks -- requiring re-optimizing the provided data and re-training per each task.

\paragraph{Latent-based control.} Here the task is to learn a behavior manifold that can be sampled from. \citet{park2019learning} learn to predict future states and then a controller to track that behavior. \citet{won2022physics} learn a conditional variational auto-encoder (VAE), conditioned on the next state. Finally, closest to our work, ASE \cite{peng2022ase} presents an unsupervised discriminative learning procedure. By maximizing the mutual information between the latent space and the produced next state, in addition to optimizing the imitation learning objective, the agent learns to generate diverse motions.

These methods share a common theme -- the resulting latent space is complex to control. In this work, we learn a dense representation of human motion jointly with a directable latent-conditioned policy.

\subsection{Representation Learning}

Representation learning, the task of capturing the underlying structure of data, has been a significant area of focus in the field of machine learning, particularly in the representation of images and videos. There are different ways to define similarity between data points, but one approach that is closest to our method is by directly utilizing a downstream task. For instance, generative adversarial networks \cite[GAN]{goodfellow2020generative} and VAEs \cite{kingma2013auto} learn a low-dimensional latent representation in order to recreate data from the reference distribution.

In the context of learning representations for motions, VAE and adversarial-based training can be differentiated. Motion VAEs \cite{ling2020character,won2022physics} enable jointly learning to represent and generate motion. While VAE-based methods are typically easier to train, their pitfall is that they are driven by a reconstruction loss, preventing the ability to deviate from the data distribution.

The benefit of adversarial methods is in their ability to generate new motions that are likely under the reference data distribution. This is important in the context of controllable characters. The character must transition naturally between every motion pair, even when it is not provided with an explicit demonstration.

As shown in ASE \cite{peng2022ase}, the policy learns to generate diverse behaviors, generalizing beyond motion reconstruction. However, their resulting latent space lacks global semantic structure. As a result, ASE tends to mode-collapse and does not provide an easy way to map motions to the latent space, an important requirement for directability and control.
Alternative work such as PADL \cite{juravsky2022padl} utilized supervision from natural language.
 These methods are able to train language-aligned motion representations that can then generate behaviors based on natural language commands. However, they require access to labeled data.

In this work, the data is unlabeled and the representation is learned end-to-end with the imitation learning policy. Hence, the semantic connections between motions are determined in the context of the policy reproducing the motions.

\subsection{Hierarchical Reinforcement Learning}

Latent generative models can be seen as part of the options/skills framework \cite{sutton1999between}. The options framework differentiates between a high and low-level controller. The low-level controller produces micro-actions, which are high-frequency actions capable of generating diverse behaviors. On the other hand, the high-level controller often plans at a lower frequency. At each time step, it selects which skill to play. Skills are temporally extended actions, or, in our context, long-term motions.

Prior efforts in hierarchical reinforcement learning (HRL) have shown that utilizing meaningful skills can not only speed up training \cite{tessler2017deep}, but also enforce the solution to reside within the support of the data \cite{peng2022ase}.

\section{Reinforcement Learning Background}

In this work, both the pre-training and the downstream tasks are modeled as reinforcement learning problems, where an agent interacts with an environment according to a policy $\pi$. At each step $t$, the agent observes a state $s_t$ and samples an action $a_t$ from the policy $a_t \sim \pi (a_t | s_t)$. The environment then transitions to the next state $s_{t+1}$ based on the transition probability $p(s_{t+1} | s_t, a_t)$. The goal is to maximize the discounted cumulative reward, defined as
\begin{equation}
    J = \mathbb{E}_{p(\tau|\pi)} \left[ \sum_{t=0}^T \gamma^t r_t \middle| s_0 = s \right] \,,
\end{equation}
where $p(\tau|\pi) = p(s_0) \Pi_{t=0}^{T-1} p(s_{t+1}|s_t,a_t) \pi(a_t | s_t)$ is the likelihood of a trajectory $\tau = [s_0, a_0, r_0, \ldots, s_{T-1}, a_{T-1}, r_{T-1}, s_T]$, and $\gamma \in [0, 1)$ is the discount factor that determines whether the agent is short-sighted or considers longer-term outcomes.

\begin{figure}
    \centering
    \includegraphics[width=\linewidth]{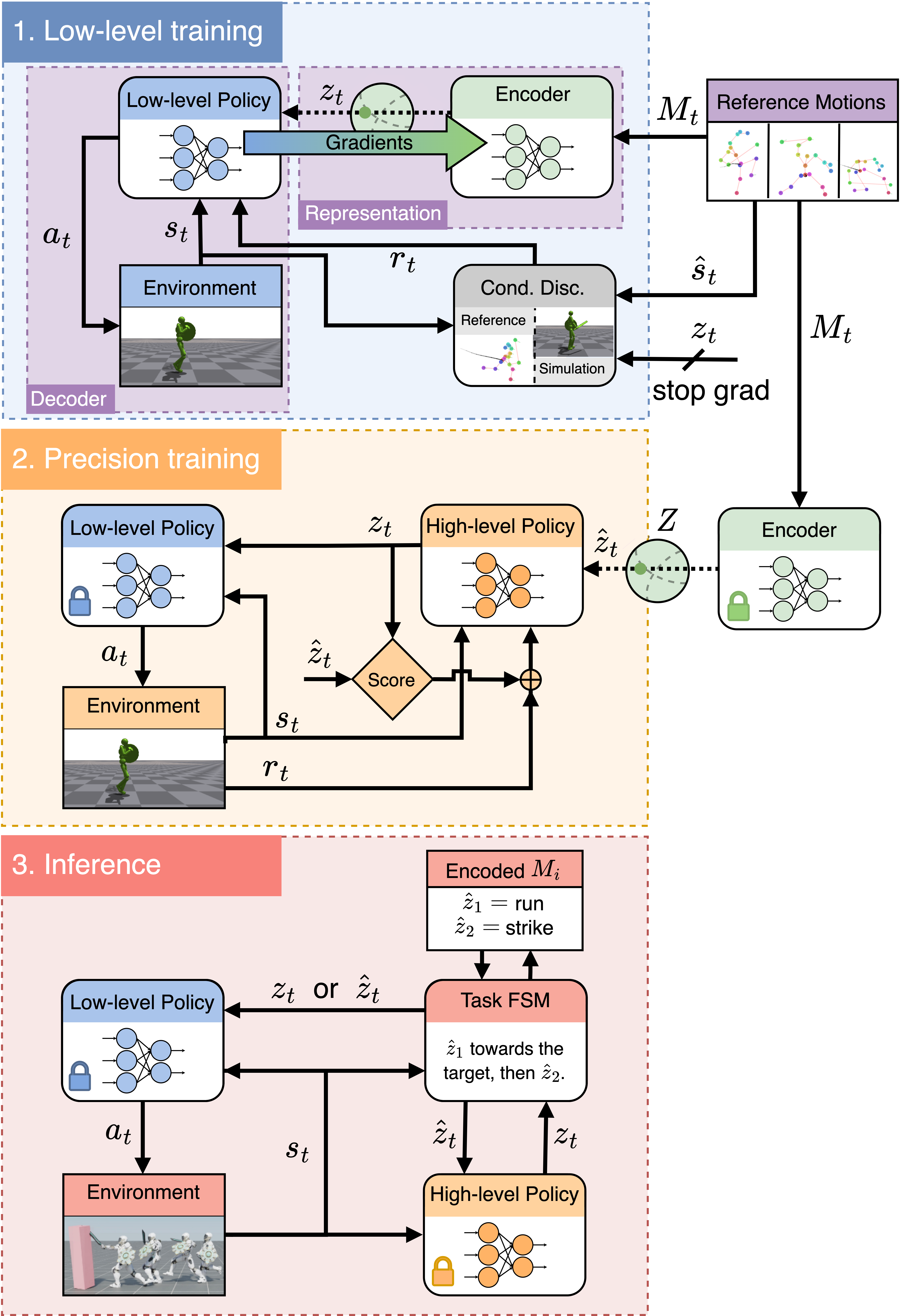}
    \caption{\textbf{The \alg~ framework} consists of three phases. \textbf{1)} In the initial training phase, both the encoder and low-level policy are developed by utilizing feedback from a conditional discriminator. The encoder learns to create a condensed representation that encapsulates the core of the motion, while the low-level policy, which interacts with the environment, serves as a decoder. \textbf{2)} Once the pre-training phase has concluded, the encoder and low-level policy are frozen. The next step is \textbf{precision training}. Here, a high-level policy is trained to control the low-level policy. \textbf{3)} In the last phase, \textbf{inference}, the high-level policy is also frozen. Solutions to complex tasks are then described using finite state machines (FSM), rule-based systems that do not require training, instead of rewards. Based on the state of the task, the FSM either provides a command to the high-level policy or provides a latent directly to the low-level policy.}
    \label{fig: motion ae}
\end{figure}

\section{Overview}

In this paper, we introduce \algfull~ (\alg), a scalable, data-driven approach for creating directable controllers for physically simulated characters. These characters can be controlled in a similar way to how players control virtual characters in games, by providing a sequence of instructions for movement and actions. \cref{fig: motion ae} outlines our framework. It consists of three steps. (1) Low-level training of a motion encoder and motion generator. (2) Precision training using a high-level policy. (3) Inference, during which tasks are solved without further training.

During low-level training, \alg~ learns an encoder $E$. It takes a motion $M$ from a reference dataset of motions $\mathcal{M}$, a time-series of joint locations, and maps it into to a low-dimensional latent representation $z \in \mathcal{Z}$. Additionally, \alg~ also jointly learns a decoder. The decoder is a low-level policy $\pi (a | s, z)$ that interacts with the simulator and generates motions similar to the reference dataset. This policy produces a variety of behaviors on demand, but is not conditioned on the directionality of the motion. For example, it can be instructed to walk, but does not enable intuitive control over the direction of walking.

Next, to control motion direction, we train a high-level task-driven policy to select latent variables $z_t$. These latents are provided to the low-level policy which generates the requested motion. Here, the motion encoder is used to constrain the latents $z_t$ to be close to pre-specified motions $\hat{z}_t$, thus guiding the high-level policy to adopt a desired behavioral style. For example, move in a given direction while performing a crouch-walking motion.

Finally, the previously trained models are combined to compose complex movements without additional training. To do so, the user produces a finite-state machine (FSM) containing standard rules and commands. These determine which motion to perform, similar to how a user controls a video game character. For example, they determine whether the character should perform a simple motion, performed directly using the low-level policy, or a directed motion requiring high-level control. As an example, one may construct an FSM like (a) "crouch-walk towards the target, until distance < 1m", then (b) "kick", and finally (c) "celebrate" (\cref{fig:teaser}).

\section{\algfull}\label{sec: low-level}

Learning a rich reusable skill representation enables characters to generate a wide variety of motions on demand, opening a range of potential applications such as games and visual effects. In \alg~, these skills are modeled using a motion-conditioned policy $\pi(a | s, z)$, where motions $M$ are encoded into a latent variable $z \in \mathcal{Z}$. This mapping is modeled using a motion encoder $z = E(M)$, and learned by solving a conditional imitation learning objective over motions $M$ sampled uniformly from a reference dataset $\mathcal{M}$:
\begin{equation}\label{eqn: objective}
    \max_\pi - \mathbb{E}_{M \in \mathcal{M}} \left[ D_\text{JS} \left( d^\pi \left(s, s' | z \right) \big|_{z = E(M)} \Big|\Big| d^M \left(\hat{s}, \hat{s}' \right) \right) \right]\,,
\end{equation}
where $D_\text{JS}$ is the Jensen-Shannon divergence, and $d^\pi \left(s, s' | z \right) \big|_{z = E(M)}$ and $d^M \left( \hat{s}, \hat{s}' \right)$ are respectively the state transition distribution of the policy and reference motions.

This objective is related to prior work in imitation learning literature, namely learning from observations (\textit{LfO}) \cite{torabi2018generative}. In this setup, the agent is provided with a demonstration dataset that consists only of state transitions/observations, without the underlying actions that produced those transitions. Generative Adversarial Imitation from Observation \cite[GAIfO]{torabi2018generative} jointly trains a policy and a discriminator. The policy generates state transitions $(s, s')$, while the discriminator tries to distinguish them from demonstrations sampled from the data $(\hat{s}, \hat{s}')$. However, provided a variety of motions, GAIfO is prone to mode-collapse, where the policy does not model all of the different behaviors from a large dataset, and the resulting model does not explicitly learn a skill embedding that can be used to direct the policy on downstream tasks.

In previous research, ASE \cite{peng2022ase} attempted to address these problems by introducing a latent variable and maximizing mutual information. They demonstrated that the model can produce diverse behavior when provided with randomly sampled variables. However, solely relying on mutual information loss is not enough. To avoid mode-collapse, ASE also employs a diversity loss, which states that similar latent variables should produce similar action distributions. Our study shows that this is particularly important in terms of directability. The encoder learns an imprecise mapping between motions and latents, resulting in a controller that does not produce similar motions when conditioned on the same latent, making it challenging to direct the policy to perform specific motions.

\alg~ mitigates the issues from prior methods by using a conditional discriminator that forces the policy to reproduce each motion in the dataset. At each iteration, a random motion $M$ is sampled from the reference dataset. The encoder maps the motion to a latent encoding $z = E(M)$. Both the policy and the conditional discriminator are then conditioned on this latent $z$. Conditioning the discriminator on the latent helps to mitigate mode collapse, by forcing the policy to produce motions that are similar to the corresponding motion of a given latent. This leads to the following discriminator loss
\begin{align*} \label{eqn: discriminator_loss0}
    \mathcal{L}_\mathcal{D} = - \mathbb{E}_{M \in \mathcal{M}} \Bigg[ &\mathbb{E}_{d^M(\hat{s}, \hat{s}')} \Big(\log \mathcal{D} ( \hat{s}, \hat{s}' | z ) \Big) \numberthis \\
    & + \mathbb{E}_{d^{\pi(z)}(s, s')} \Big( \log ( 1 - \mathcal{D} ( s, s' | z ) ) \Big) \Bigg| z = E(M) \Bigg] \,,
\end{align*}
and policy objective
\begin{align*}
    &J = \mathbb{E}_{M \in \mathcal{M}} \left[ \mathbb{E}_{p(\tau|\pi,z)} \left( \sum_t \gamma^t r(s_t, s_{t+1}, z) \Big| s_0 = s \right) \Bigg | z = E(M) \right] \,, \numberthis \label{eqn: reward}
\end{align*}
with rewards $r(s_t, s_{t+1}, z) = - \log \left(1 - \mathcal{D}\left(s_t, s_{t+1} | z \right) \right)$.

\subsection{Practical Considerations}

In this section, we present design decisions needed for a practical instantiation of \alg, as well as in-depth implementation details.

\subsubsection{Encoder}

An ideal representation is one that is optimized for the control policy, rather than relying on auxiliary objectives to drive the structure of the latent representation, as in contrastive learning methods \cite{oord2018representation}. We show that an effective encoder can be trained in an end-to-end fashion using gradients from the policy when optimizing the pre-training objective \cref{eqn: reward}. Furthermore, this approach results in a latent space with a clear semantic structure, where similar motions are grouped closely, enabling interpolation in a semantically meaningful manner.

The encoder's output is projected onto the $l_2$ unit hypersphere. This constraint is inspired by prior work in the field \cite{parkhi2015deep,bojanowski2017unsupervised,wang2017normface,xu2018spherical,chen2020simple,wang2020understanding,peng2022ase}, and has several benefits. For example, fixed-norm vectors are known to improve training stability in machine learning, where dot products are commonly used \cite{wang2017normface,xu2018spherical}. In the context of motion generation, the structure imposed on the latent space by the unit $l_2$ norm constraint reduces the likelihood of unnatural behaviors arising from sampling out-of-distribution latents during inference, as demonstrated by \citet{peng2022ase}.

As motion clips can be arbitrarily long, we split the data into overlapping sub-motions of 2 seconds. This results in motions that are long enough to present distinct and coherent characteristics. \edited{Additionally, sub-motions with close temporal proximity are likely to have similar characteristics. We leverage this understanding to improve the latent space structure by applying an alignment and uniformity loss on the predicted embeddings \cite{wang2020understanding}.
\begin{equation}
    \mathcal{L}_\text{align} = \mathbb{E}_{(M, M') \stackrel{\text{overlapping}}{\sim} \mathcal{M}} \left[ \Big|\Big| E(M) - E(M') \Big|\Big|_2^2 \right] \,,
\end{equation}
\begin{equation}
    \mathcal{L}_\text{uniform} = \log \mathbb{E}_{(M, M') \stackrel{\text{i.i.d}}{\sim} \mathcal{M}} \left[ \exp \left(-2 \Big|\Big| E(M) - E(M') \Big|\Big|_2^2 \right) \right] \,,
\end{equation}
where \textit{overlapping} corresponds to two-second sub-motions $(M, M')$ from the same original motion sequence with non-zero overlap, and \textit{i.i.d} to sub-motions randomly sampled from the data.}

\subsubsection{Skill Transitions}

When performing new tasks, agents often need to sequentially transition between behaviors. Even in simple tasks, like reaching a location, the character needs to utilize a combination of skills like walking, turning, and standing. To enable smooth and robust transitions between motions, we explicitly train the model to transition between different motions by changing the conditional motion $M$ at random timesteps. This teaches the policy to successfully transition between disparate motions.

\subsubsection{Discriminative Loss}

In adversarial imitation learning the discriminative reward provides the learning signal to the agent. To improve training dynamics, \cite{peng2021amp,peng2022ase,juravsky2022padl} propose to use a gradient penalty regularizer \edited{and negative sampling}. This regularization helps mitigate discriminator overfitting and produces a smoother optimization landscape for the policy. The result is improved training stability and overall quality of the generated motion. In addition, as our goal is to learn a motion encoding optimal for the control task, we prevent gradients from flowing from the discriminator's objective into the encoder. This results in the following objective:
\begin{align*}
    \mathcal{L}_\mathcal{D} = &- \mathbb{E}_{M \in \mathcal{M}} \Big( \mathbb{E}_{d^\pi (s, s' | z)} \left[ \log \left( 1 - \mathcal{D} (s, s' | z) \right) \right] \numberthis \label{eqn: discriminator loss} \\
    &\edited{+ \mathbb{E}_{d^M (\hat{s}, \hat{s}')} \left[ \log \mathcal{D} (\hat{s}, \hat{s}' | z) + \log \left( 1 - \mathcal{D} (s, s' | z' \sim \mathcal{Z}) \right) \right]} \\
    &+  w_\text{gp} \mathbb{E}_{d^\mathcal{M} (\hat{s}, \hat{s}')} \left[ || \nabla_\theta \mathcal{D}(\theta) |_{\theta = (\hat{s}, \hat{s}' | z)} ||^2 \right] \Big | z = \text{stop grad}(E(M)) \Big) \,.
\end{align*}
Here, $w_\text{gp}$ is the gradient penalty coefficient.

\section{High-level control}

Once the low-level controller has been trained, it is used as a motion generator, grounding the motions of the character to those seen in the data. In this section, we present how to train a high-level policy to control the direction in which motions are performed, which can then be leveraged to solve complex tasks in varying forms without specifically training on them.

\subsection{Precision training}\label{subsec: hrl precision}

Provided a motion encoding $\hat z = E(M)$, the low-level policy generates a matching motion, however, providing the low-level policy with the encoding for "run" does not control the running direction. To provide better control we train a high-level policy to generate motion in the requested form and direction.

Specifically, the high-level policy produces latent variables $z_t$. These are then provided to the low-level policy which controls the character. The character is tasked with moving in a specified direction $\mathbf{d}_t^*$ while crouch-walking $M_1$, or alternatively sprinting $M_2$. We achieve this by combining a task reward with a latent similarity loss. Specifically, given a motion encoding $\hat{z} = E(M)$, we train the high-level policy with the following reward
\begin{equation}
    r_t^\text{locomotion} = \text{exp}\! \left(\!- 0.25 \Bigg|\Bigg|\mathbf{d}^*_t - \frac{\dot{\mathbf{x}}_t^\text{root}}{||\dot{\mathbf{x}}_t^\text{root}||} \Bigg|\Bigg|^2 \right) + \text{exp}\! \left(\!- 4 || z_t - \hat z ||^2 \right) \,,
\end{equation}
where $\dot{\mathbf{x}}_t^\text{root}$ is the character's velocity.

\subsection{Exemplar guidance}\label{method: fsm}

Given a pre-trained encoder and low-level policy (\cref{sec: low-level}) and a pre-trained high-level policy (\cref{subsec: hrl precision}), the task designer describes how a task should be solved, in a natural and intuitive way, overcoming the fragility of reward design. Here, the task designer provides demonstrations for the various motions the agent should perform, enclosed within a finite-state machine (FSM) that determines when to transition between behaviors. For instance, the task of striking an object can be broken down into three phases "run towards the object", once within 0.5 meters then "perform an attack", and then "stand idle" until the next command.

Achieving such a level of control requires a combination of versatility, provided by the low-level policy, and precision, provided by a pre-trained high-level policy. During phases requiring precision, such as moving in a specified direction, the FSM provides the high-level policy with a requested motion embedding $\hat{z}$ and a direction in which this motion should be performed. When transitioning to isolated motions, such as a specific sword swipe, the FSM provides the motion encoding directly to the low-level policy.

This enables re-usability without re-training and resembles how a user would interact with the character given a game controller. A single combination of (a) low-level policy, (b) encoder, and (c) high-level policy, are used to solve unseen tasks in varying forms.

\section{Experiments}

\textbf{The data:}
To acquire diverse motion control capabilities, the low-level policy is trained using 160 motion clips totaling over 30 minutes in duration \cite{reallusion}. Each motion clip is broken down into 2-second continuously overlapping sub-sequences\edited{, oblivious to transition boundaries. Hence, if a motion clip contains multiple skills, the 2-second clips may contain motions from several skills.}. This includes basic movements such as various forms of walking, as well as more complex motions such as sword-strike combinations. This is explained in further detail in the supplementary material.

\textbf{Training workflow:}
Throughout the pre-training process, the agent interacts with the environment for rollouts of $K$ steps. At random timesteps, a random motion is sampled from the reference dataset, it is then encoded and provided to the low-level policy. The rollout then consists of the observed states resulting from the low-level policy, conditioned on the latent $z$, interacting with the environment. The low-level policy is then trained with respect to the collected rollout using PPO \cite{schulman2017proximal}.

We parallelize training over 4096 Isaac Gym \cite{makoviychuk2021isaac} environments on a single A100 GPU, for a total of 5 billion steps. The low-level (high-level) policy takes decisions at a rate of 30 (6) Hz. The latent space $\mathcal{Z}$ is defined as a 64D hypersphere. 

\textbf{Model architecture}
The encoder is a standard MLP, mapping $E(M) \mapsto z$, the policy and conditional discriminator each contain an additional input head $H(z)$ for latent parsing, followed by an MLP $\pi(s, H(z)) \mapsto a$, where $a \in \mathbb{R}^{31}$.

\section{Results}

We tested the effectiveness of our method by using \alg~ to learn skill embeddings that allow a simulated humanoid to perform various motion control tasks. We demonstrate that \alg~ learns a semantically meaningful latent representation of diverse human motion and a directable policy. We then trained a high-level policy to control the direction in which these motions are performed. Finally, we show how these low and high-level policies can be re-used for solving unseen tasks without further training. See the motions produced by \alg~ in the provided supplementary video \footnote{\edited{See \cref{sec: limitations} for an explanation on rendering artifacts within the visualization process.}}.


\begin{table}[t]
    \centering
    \begin{tabular}{l|c|c|c}
         & Encoder quality $\downarrow$ & Diversity $\uparrow$ & Controllability $\uparrow$ \\
         \toprule 
        \textbf{\alg} & \textbf{0.23} & \textbf{19.8$ \pm $0.1} & \textbf{78\%} \\
        \textbf{ASE} & 0.68 & 18.6$\pm$0.4 & 35\% \\
        \hline
    \end{tabular}
    \caption{\textbf{Pre-training:} Quantitative evaluation of the learned encoder and low-level policy. We measure Fisher's concentration coeff. (Encoder quality), Inception score (Diversity), and Generation accuracy (Controllability).}
    \label{tab: LLC results}
\end{table}

\subsection{Controllable motion generation}
\label{sec: results llc}
We begin by analyzing three aspects of \alg: (1) the \textit{encoder quality}, (2) \textit{diversity} of the low-level controller, and (3) \textit{controllability} of the combined system. Results are reported in \cref{tab: LLC results}. 
We focus our comparison on ASE \cite{peng2022ase}, a latent generative model which learns to map arbitrary latent variables to motions.

\textit{Experiment 1: Encoder quality. }
Using Fisher's class separability metric \cite{bishop1995neural} over the representation learned by the encoder, we measure the separability between the motion classes within the latent space, where a motion class is defined as sub-motions within a single motion file. As shown in \cref{tab: LLC results}, \alg~ learns to encode motions into representations with much better separation.

\textit{Experiment 2: Diversity. } We trained a classifier using the reference dataset from \cref{sec: low-level} to map a motion sequence to the originating motion index. We report the Inception Score \cite{salimans2016improved} over generated motions, when generated from randomly sampled latents $z \sim \mathcal{Z}$. As seen in \cref{tab: LLC results}, \alg~ significantly improves the diversity of generated motion.

\textit{Experiment 3: Controllability.} Finally, we quantified how well \alg~ generates the requested motions, using a user study. Provided a reference motion and a textual description (taken from \citet{juravsky2022padl}) raters were asked to classify the generated motion as similar or not. For each model, we presented raters with 40 reference motions and 3 generations per reference. We report the accuracy, measured as the percentage of accurate generations by the controller. The results show that \alg~ enables better control over the generated motions, compared to ASE, increasing the accuracy of perceived generation from $35\%$ to $78\%$. 

This was enabled by performing end-to-end learning of both the representation (encoder) and the generative motion model (low-level policy) using a conditional discriminative objective. As a result, \alg~ learns to encode motion onto a semantically meaningful representation and a controller capable of generating motion with similar characteristics to the demonstration.

\begin{table}[t]
    \centering
    \begin{tabular}{l|cc|cc|cc}
        \multirow{2}{*}{\textbf{Motion}} & \multicolumn{2}{c|}{\textbf{Heading}} & \multicolumn{2}{c|}{\textbf{Location}} & \multicolumn{2}{c}{\textbf{Strike}} \\ \cline{2-7}
            & Style & Score & Ending & Score & Ending & Score \\
         \toprule 
        \multirow{3}{*}{Run} & \multirow{3}{*}{1} & \multirow{3}{*}{0.92} & \faIcon{male} & 0.98 & \faIcon{shoe-prints} & 0.96 \\
            & & & \faIcon{child} & 0.99 & \faIcon{shield-alt} & 1 \\
            & & & \faIcon{frog} & 0.96 & \faIcon{gavel} & 0.99 \\ \cline{2-7}
            \\[-1em]
        \multirow{3}{*}{Walk} & \multirow{3}{*}{0.81} & \multirow{3}{*}{0.92} & \faIcon{male} & 0.99 & \faIcon{shoe-prints} & 0.98 \\
            & & & \faIcon{child} & 1 & \faIcon{shield-alt} & 1 \\
            & & & \faIcon{frog} & 0.97 & \faIcon{gavel} & 0.96 \\ \cline{2-7}
            \\[-1em]
        \multirow{3}{*}{\shortstack[l]{Crouch \\ Walk}} & \multirow{3}{*}{0.94} & \multirow{3}{*}{0.91} & \faIcon{male} & 0.98 & \faIcon{shoe-prints} & 0.99 \\
            & & & \faIcon{child} & 0.99 & \faIcon{shield-alt} & 1 \\
            & & & \faIcon{frog} & 0.97 & \faIcon{gavel} & 0.99 \\ \hline
    \end{tabular}
    \caption{\edited{Quantitative evaluation of \textbf{directional motion control} (heading) and \textbf{zero-shot task solution}. We consider three forms of locomotion: \textit{run}, \textit{walk}, and \textit{crouch-walk}, each characterized by a different speed and style. For each task, we consider various finishing motions. For the location task, we consider \faIcon{male} (stand idle), \faIcon{child} (celebrate, arms-up), and \faIcon{frog} (crouch idle). In the strike task, we consider: \faIcon{shoe-prints} (kick), \faIcon{shield-alt} (shield charge), and \faIcon{gavel} \text{(sword swipe)}.}}
    \label{tab: FSM results}
\vspace{-0.5cm}
\end{table}

\subsubsection{Qualitative analysis}

In \cref{fig: llc behaviors}, we show motions generated by \alg~
. Throughout a single episode, the conditional motions were changed, resulting in human-like transitions between the requested motions. Additionally, to illustrate the semantic structure of the latent space, we encode two semantically connected motions "sprint" and "crouching idle" and interpolate between their encodings over time. As shown in \cref{fig llc: interpolate}, \alg~ smoothly transitions between the two motions, decreasing both speed and height while continuously performing a form of walking motion.


\subsection{Solving downstream tasks}

Using the encoder and low-level policy from \cref{sec: results llc}, we show how they can be used to compose motions for solving unseen tasks using commands akin to video game control.

\subsubsection{Directional motion control}\label{subsubsec: precision training}

\begin{figure}[!ht]
     \centering
     \begin{subfigure}[b]{0.33\textwidth}
         \centering
         \includegraphics[width=0.2475\textwidth]{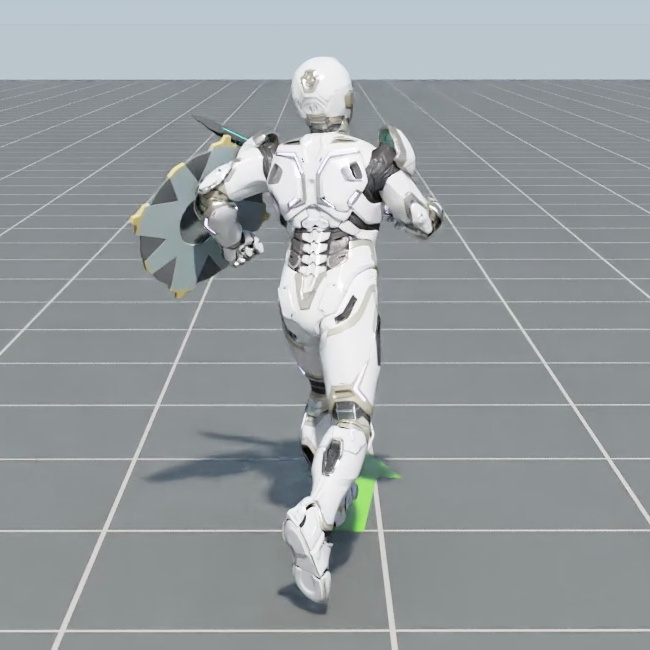}\hfill
         \includegraphics[width=0.2475\textwidth]{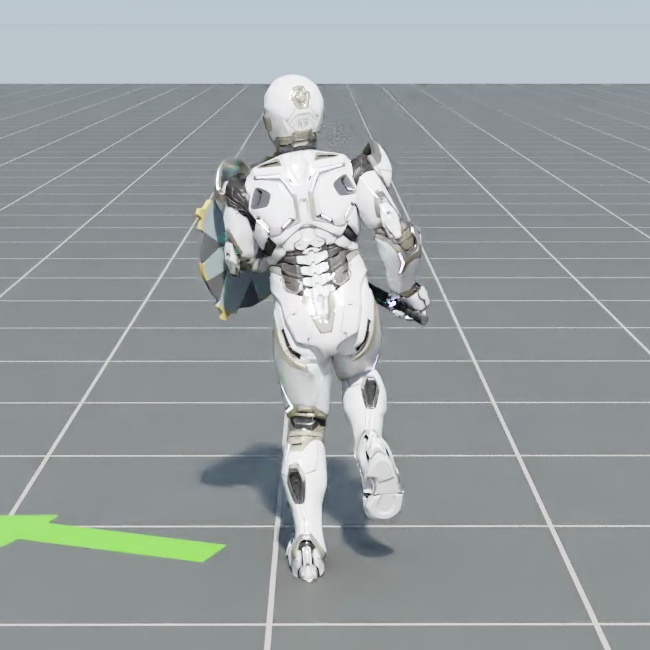}\hfill
         \includegraphics[width=0.2475\textwidth]{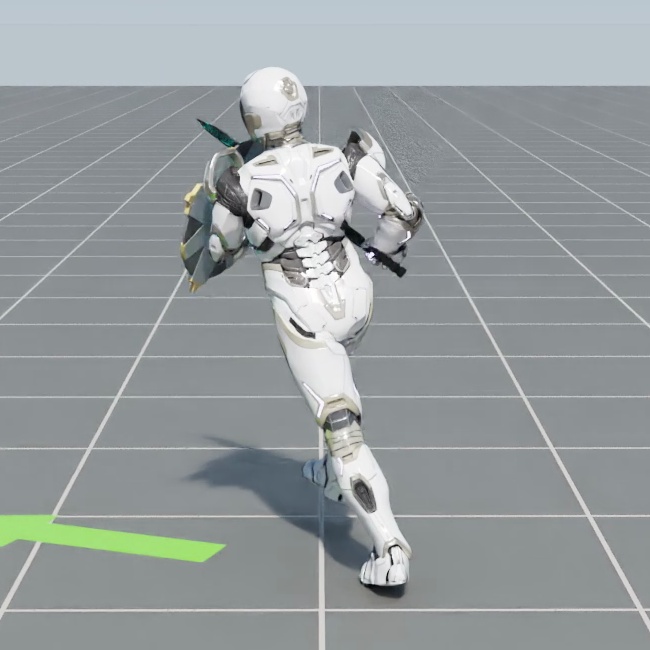}\hfill
         \includegraphics[width=0.2475\textwidth]{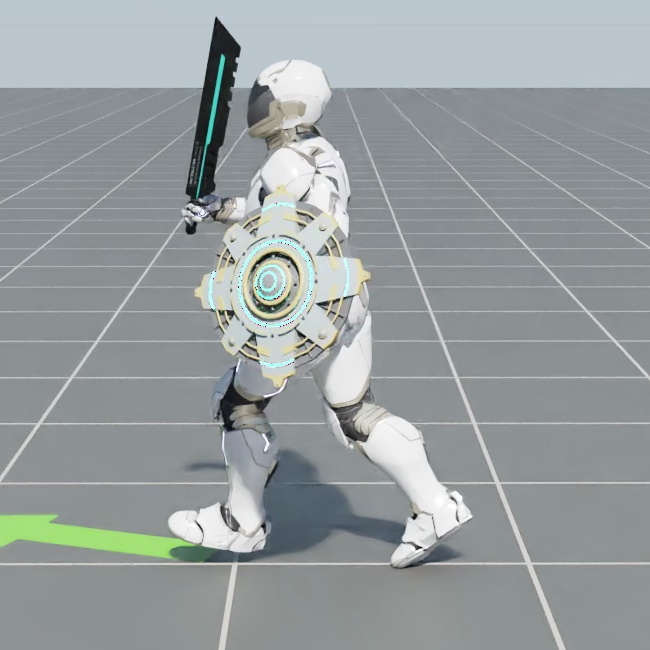}
         \caption{Heading: Run}
         \label{fig hrl: heading run}
     \end{subfigure}
     \hfill
     \begin{subfigure}[b]{0.33\textwidth}
         \centering
         \includegraphics[width=0.2475\textwidth]{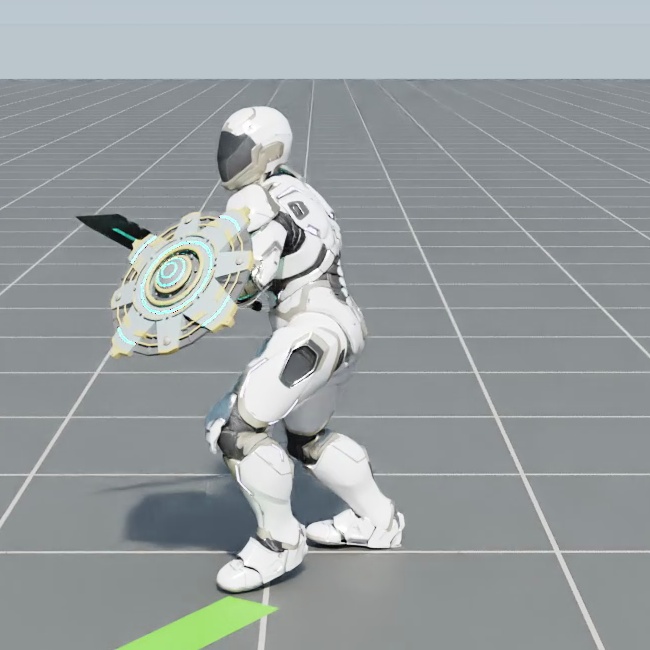}\hfill
         \includegraphics[width=0.2475\textwidth]{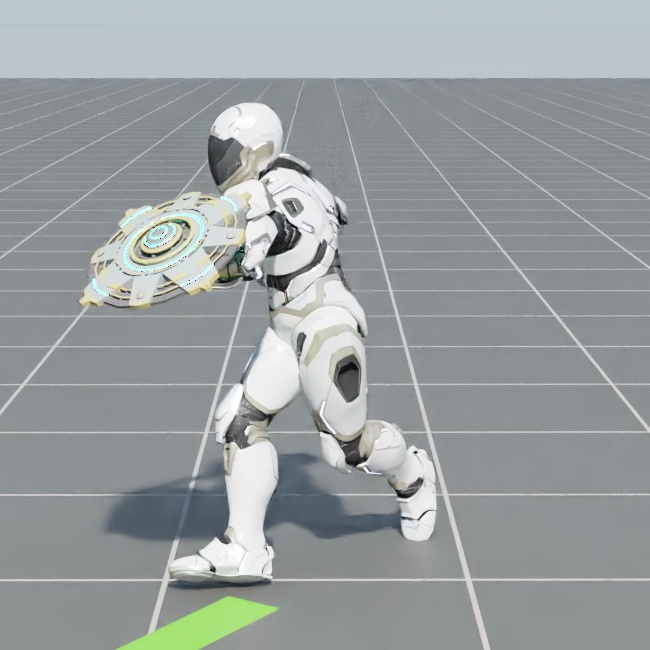}\hfill
         \includegraphics[width=0.2475\textwidth]{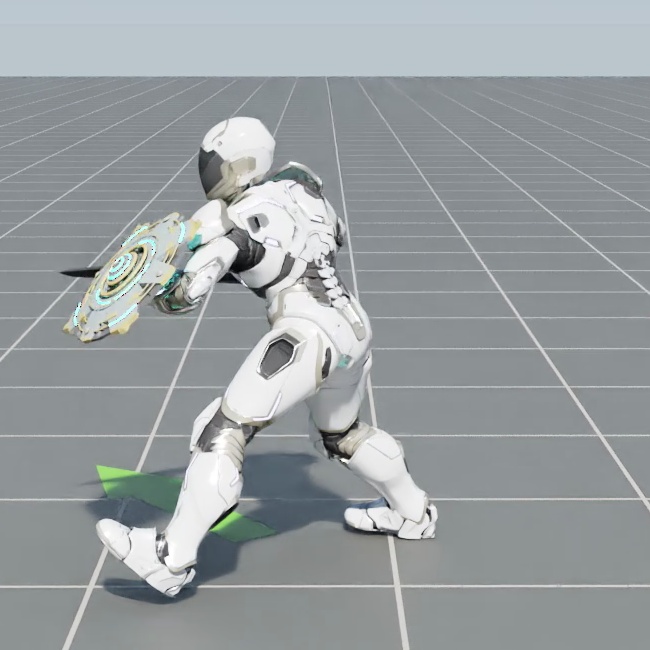}\hfill
         \includegraphics[width=0.2475\textwidth]{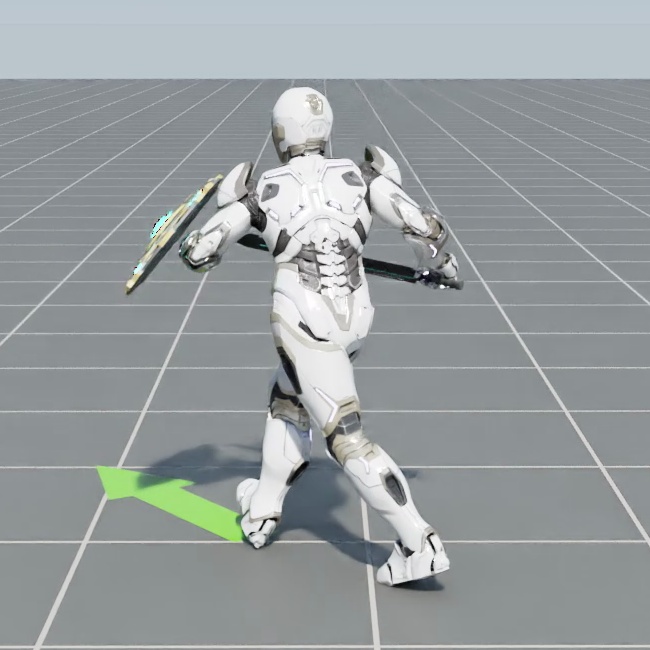}
         \caption{Heading: Walk, shield up}
         \label{fig hrl: heading walk}
     \end{subfigure}
     \hfill
     \begin{subfigure}[b]{0.33\textwidth}
         \centering
         \includegraphics[width=0.2475\textwidth]{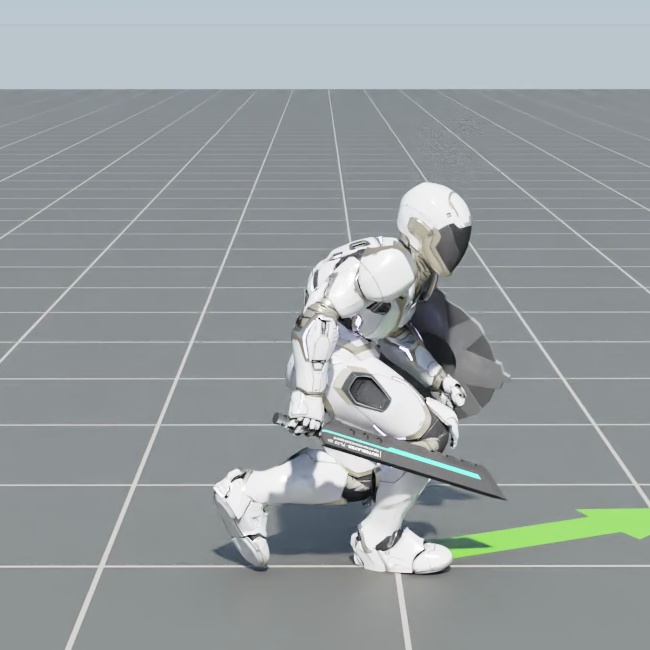}\hfill
         \includegraphics[width=0.2475\textwidth]{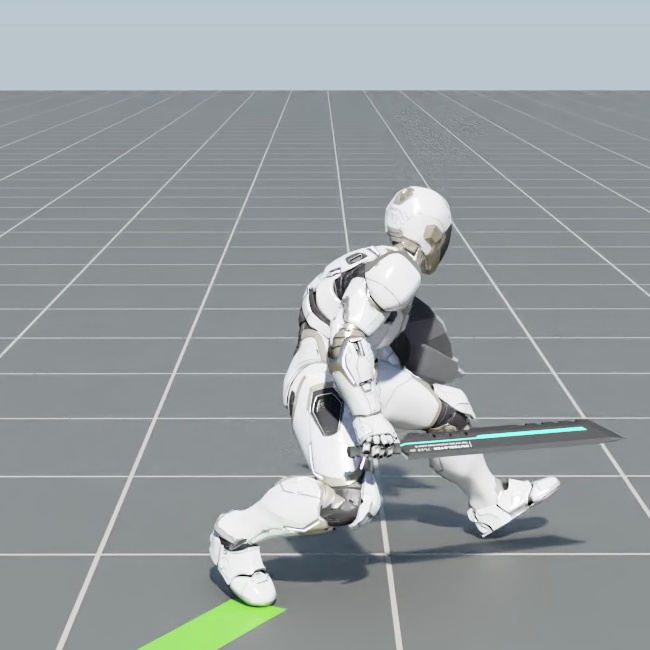}\hfill
         \includegraphics[width=0.2475\textwidth]{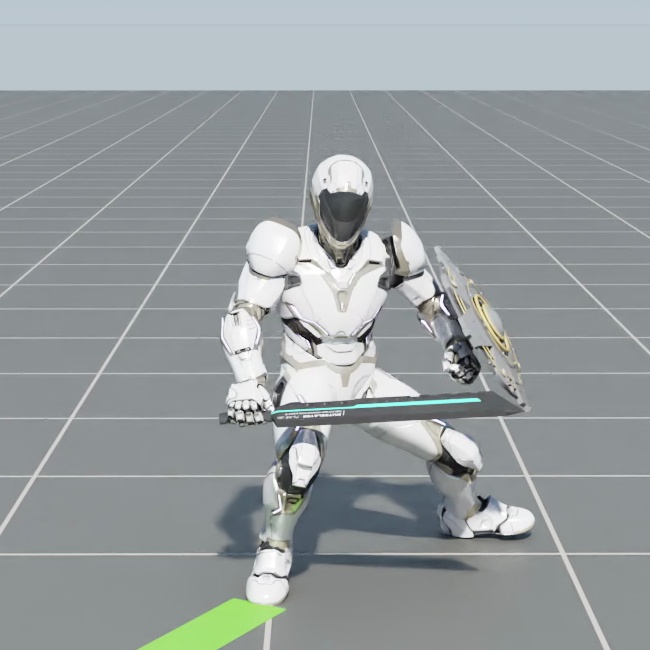}\hfill
         \includegraphics[width=0.2475\textwidth]{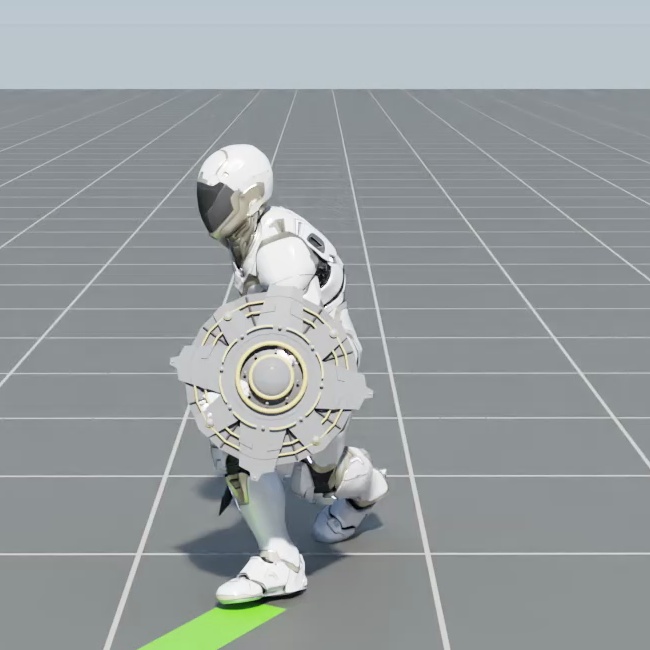}
         \caption{\edited{Heading: Crouch-walk}}
         \label{fig hrl: heading crouch-walk}
     \end{subfigure}
    \caption{\textbf{Precision training:} A single high-level controller is trained to generate style-constrained locomotion via reward guidance.}
    \label{fig: hrl precision}
\end{figure}

First, we show that provided a reference motion $M$ and a direction $d^*$, a high-level policy can learn to control the low-level policy. We refer to this task as Heading. The character should produce motions with similar characteristics in the requested direction. We demonstrate the learned motions in \cref{fig: hrl precision} and quantify them in \cref{tab: FSM results} under the Heading column. Here, a high-level policy was conditioned on jointly learning "run", "walking with a raised shield", and "walk crouching".

\edited{During the evaluation, the high-level policy is conditioned on a fixed style and the direction is changed at random timesteps. We report the success in generating the requested style, measured using human raters, and the direction of motion, measured as the cosine distance between the requested and actual movement direction.}

\newpage
\onecolumn
\begin{figure}[!ht]
     \centering
     \begin{subfigure}[b]{0.498\textwidth}
         \centering
         \includegraphics[width=0.165\textwidth]{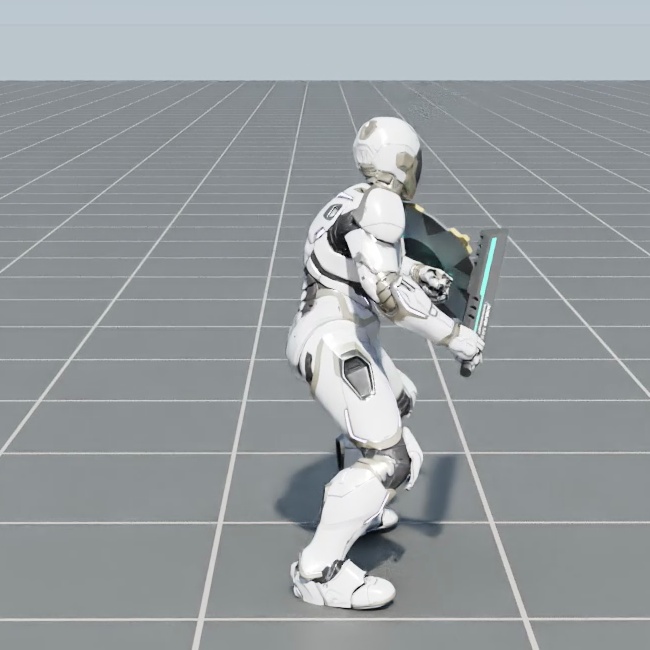}\hfill
         \includegraphics[width=0.165\textwidth]{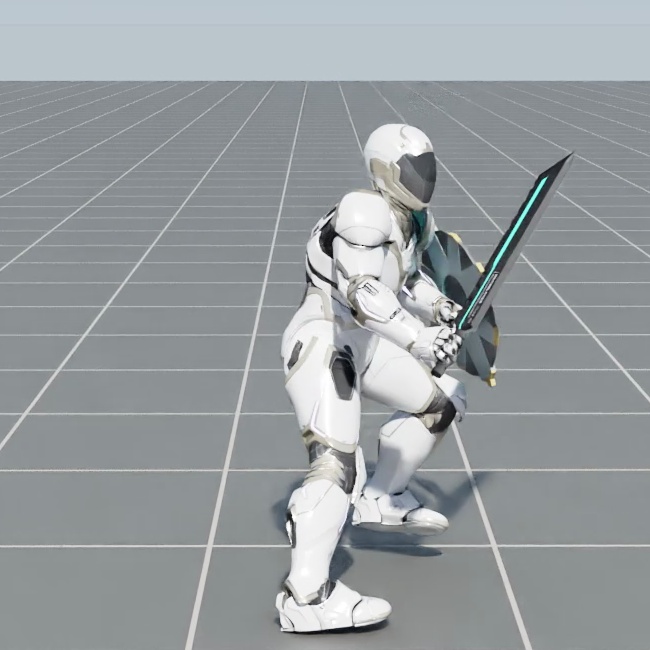}\hfill
         \includegraphics[width=0.165\textwidth]{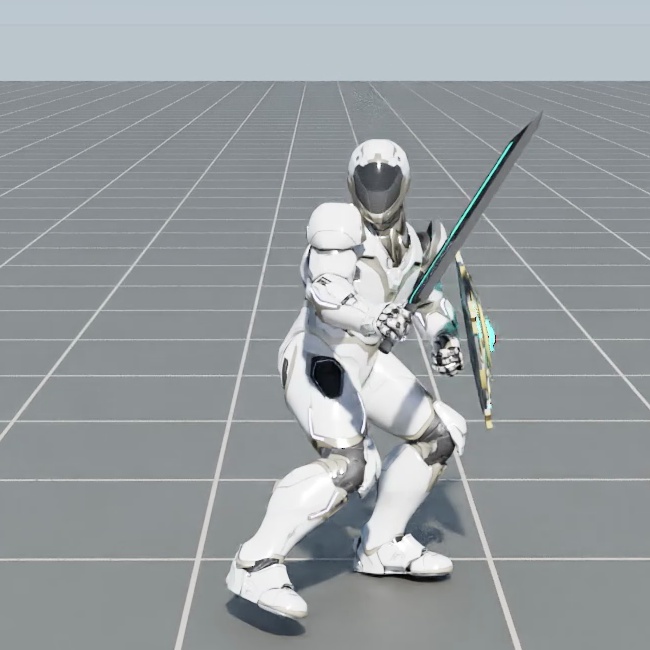}\hfill
         \includegraphics[width=0.165\textwidth]{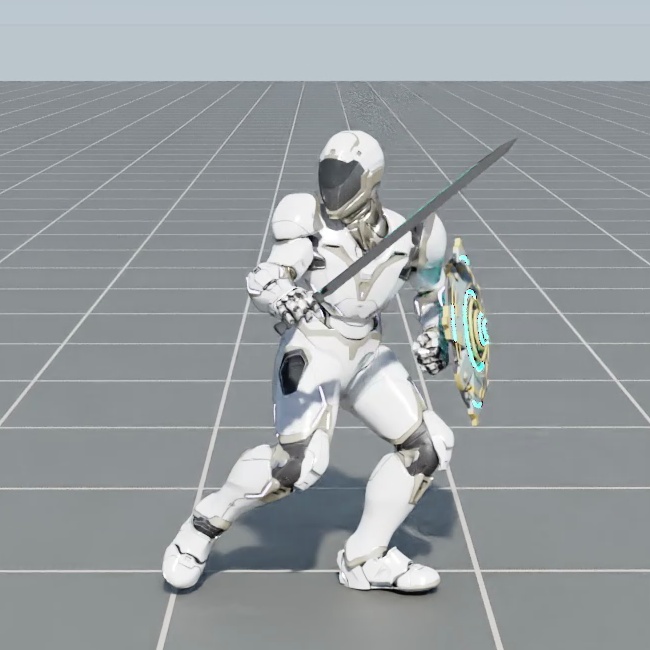}\hfill
         \includegraphics[width=0.165\textwidth]{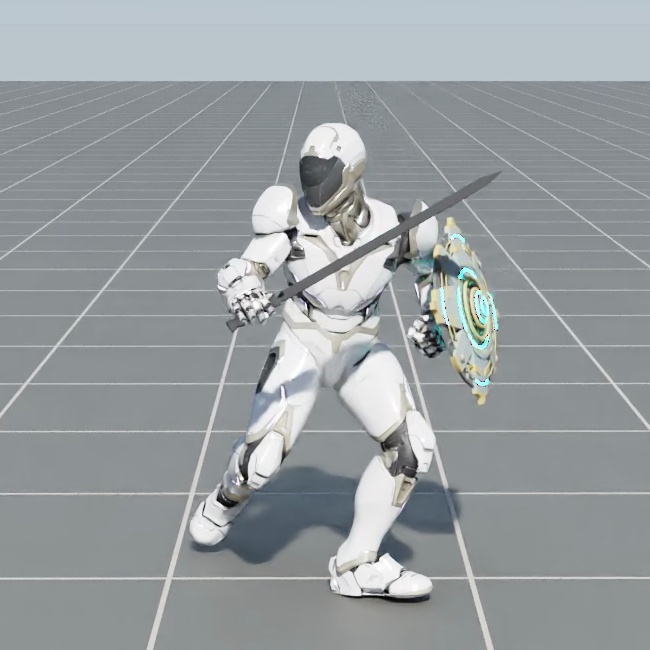}\hfill
         \includegraphics[width=0.165\textwidth]{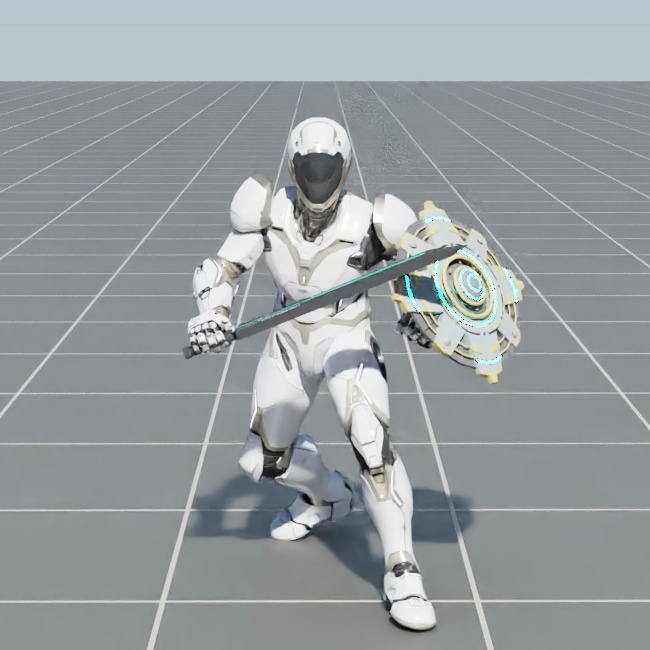}
         \caption{Turn $180^{\circ}$}
         \label{fig llc: turn}
     \end{subfigure}
     \hfill
     \begin{subfigure}[b]{0.498\textwidth}
         \centering
         \includegraphics[width=0.165\textwidth]{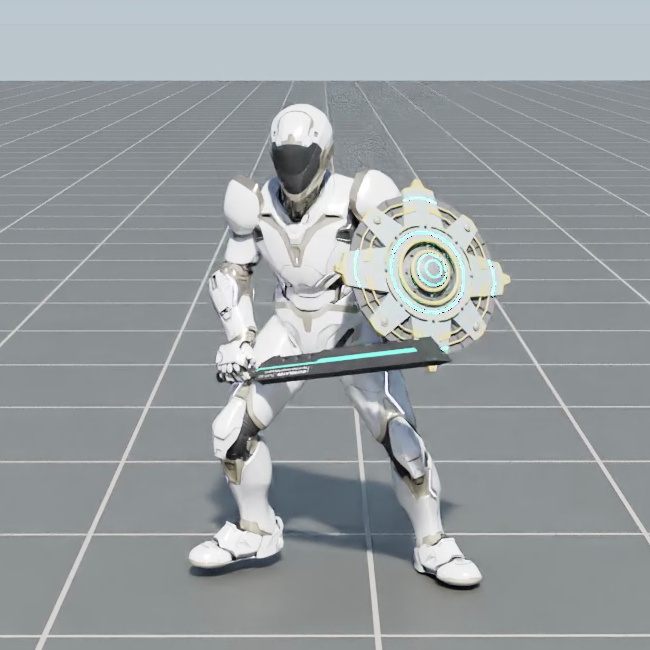}\hfill
         \includegraphics[width=0.165\textwidth]{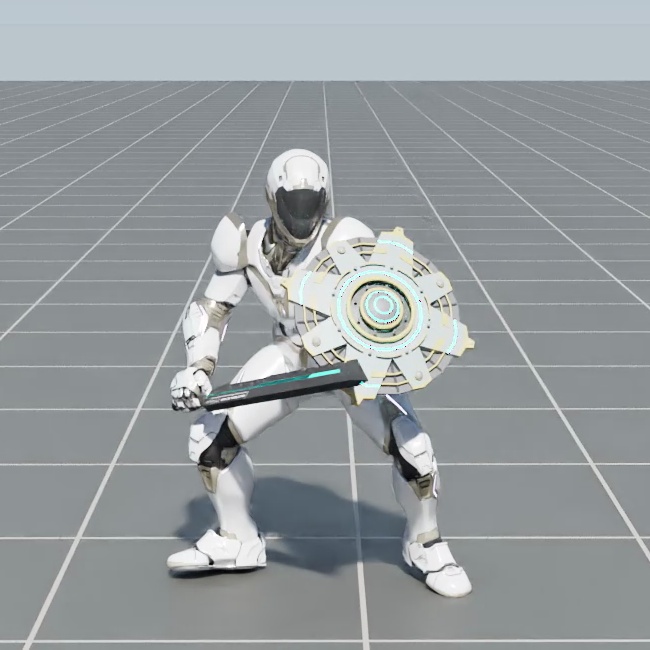}\hfill
         \includegraphics[width=0.165\textwidth]{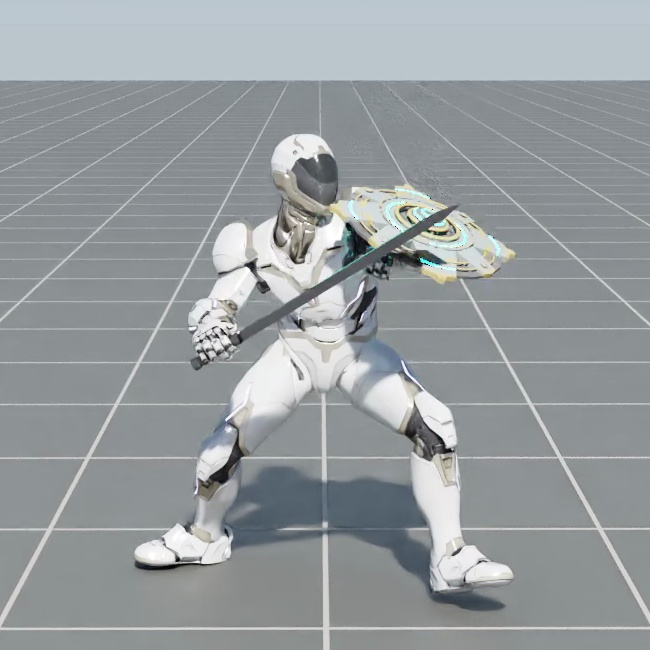}\hfill
         \includegraphics[width=0.165\textwidth]{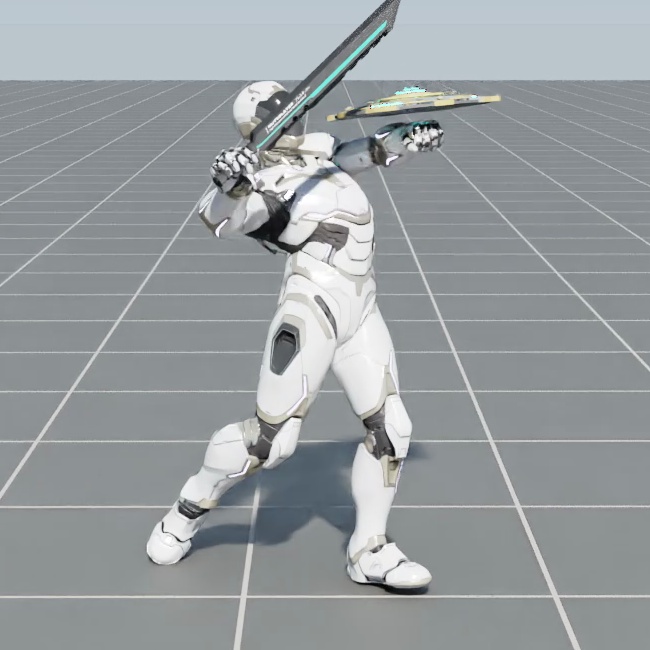}\hfill
         \includegraphics[width=0.165\textwidth]{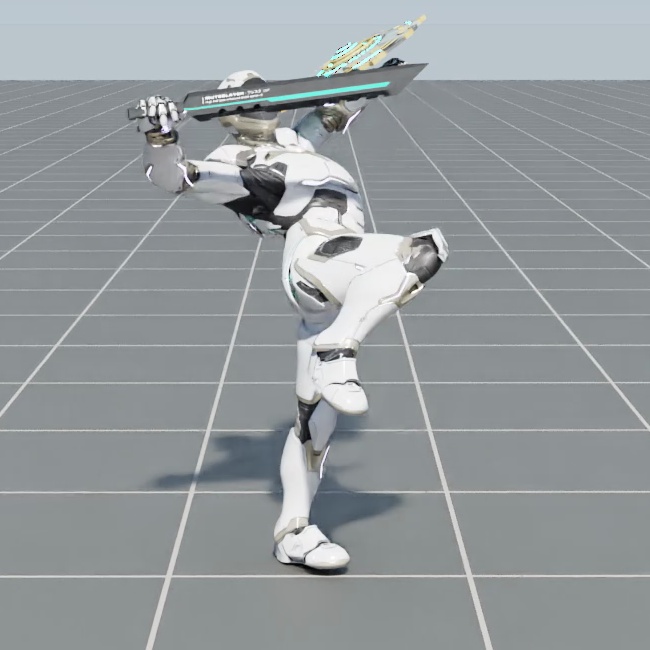}\hfill
         \includegraphics[width=0.165\textwidth]{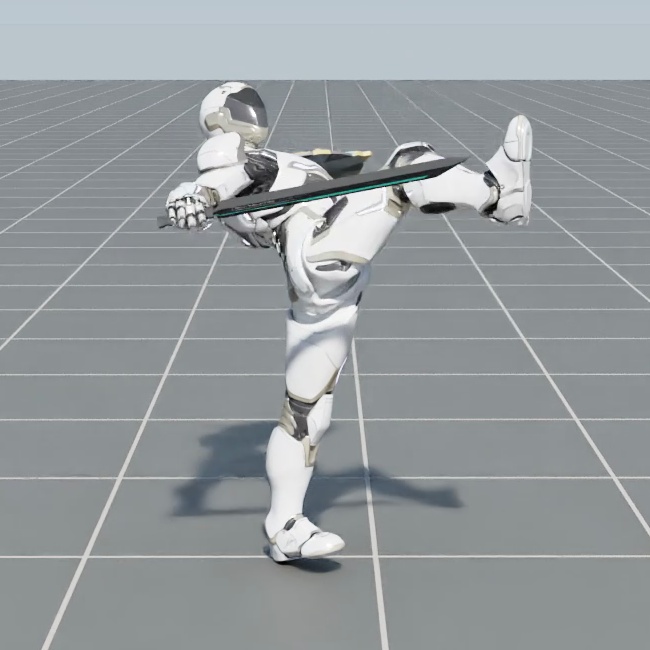}
         \caption{Kick}
         \label{fig llc: kick}
     \end{subfigure}\\
     \begin{subfigure}[b]{0.498\textwidth}
         \centering
         \includegraphics[width=0.165\textwidth]{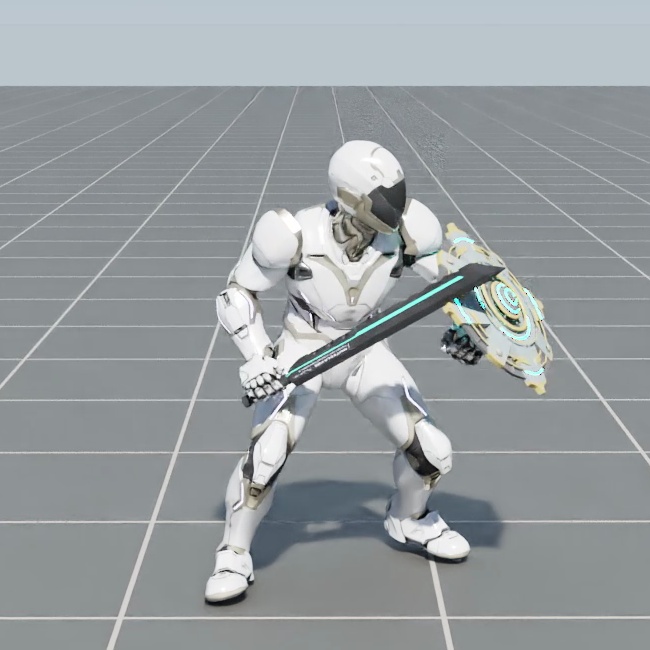}\hfill
         \includegraphics[width=0.165\textwidth]{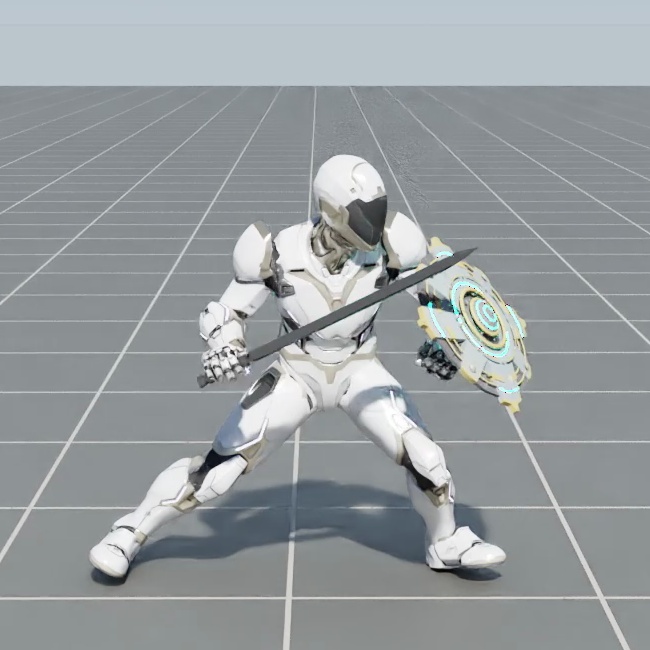}\hfill
         \includegraphics[width=0.165\textwidth]{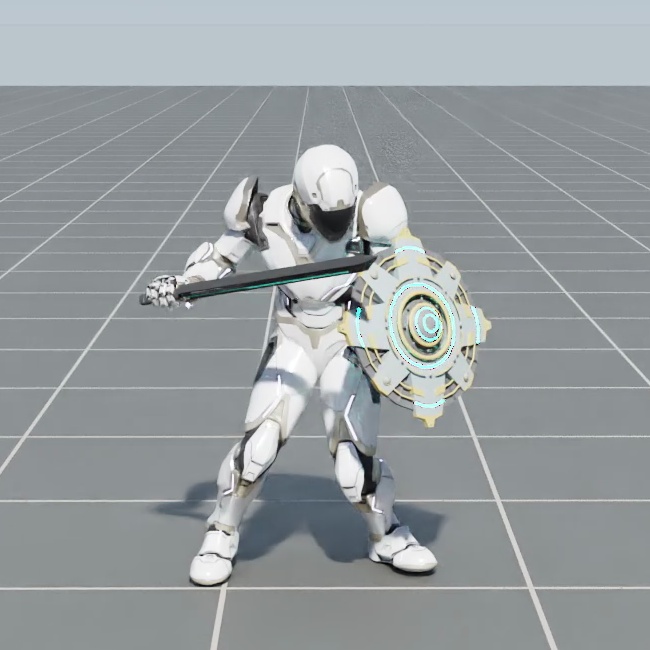}\hfill
         \includegraphics[width=0.165\textwidth]{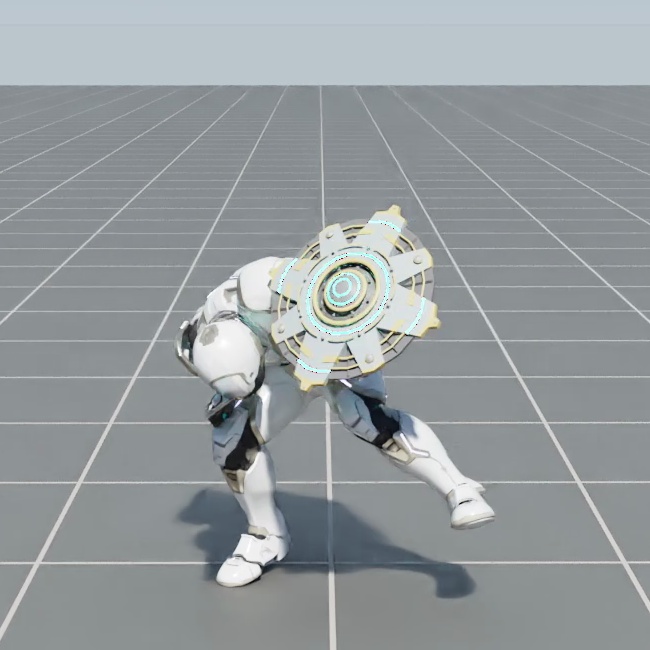}\hfill
         \includegraphics[width=0.165\textwidth]{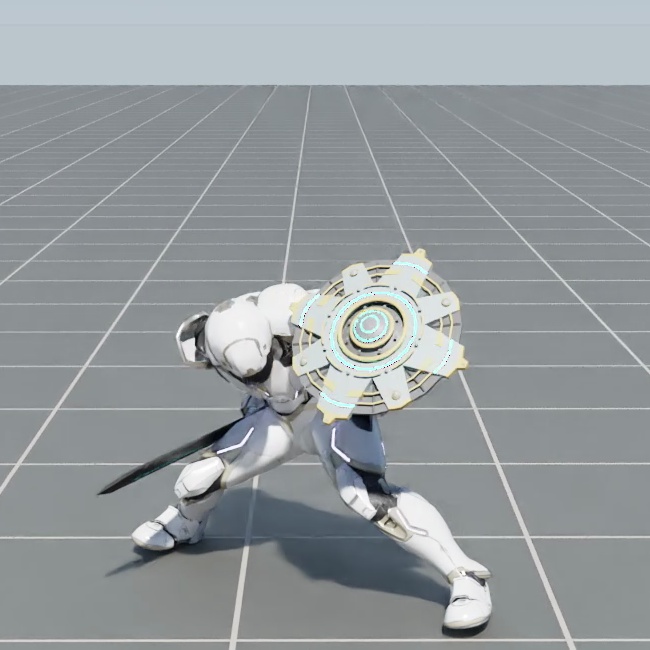}\hfill
         \includegraphics[width=0.165\textwidth]{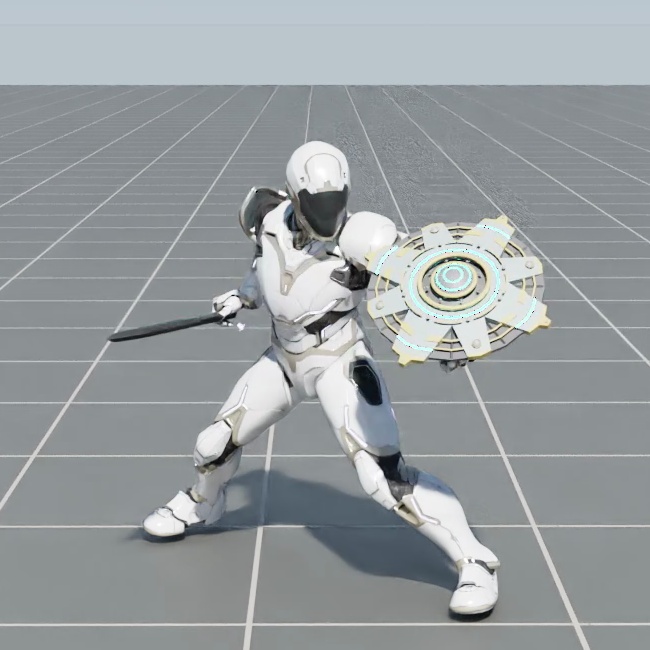}
         \caption{Shield Charge}
         \label{fig llc: shieldcharge}
     \end{subfigure}
     \hfill
     \begin{subfigure}[b]{0.498\textwidth}
         \centering
         \includegraphics[width=0.165\textwidth]{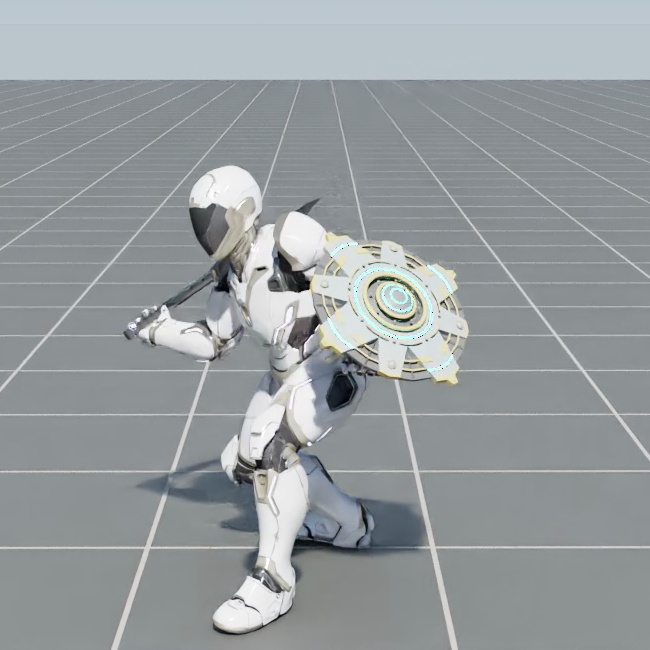}\hfill
         \includegraphics[width=0.165\textwidth]{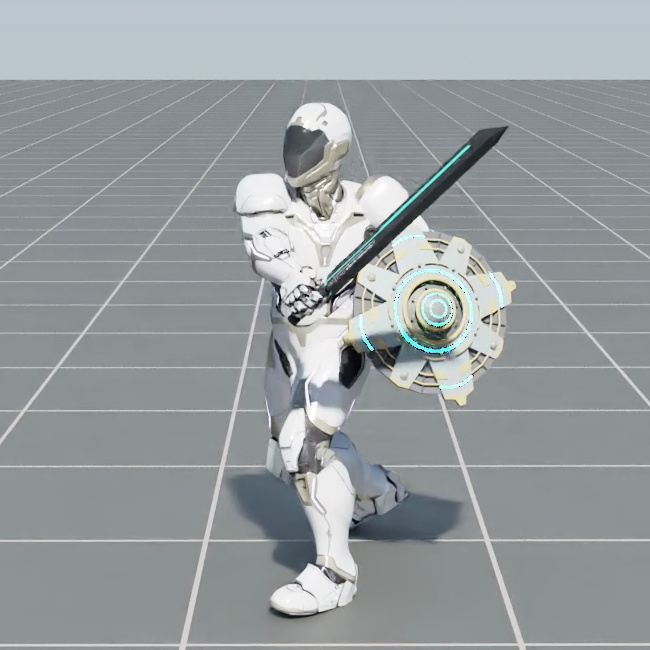}\hfill
         \includegraphics[width=0.165\textwidth]{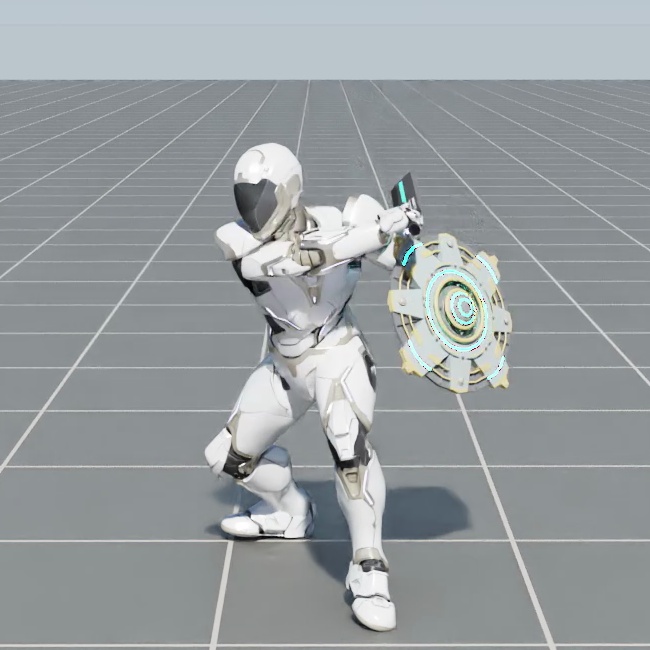}\hfill
         \includegraphics[width=0.165\textwidth]{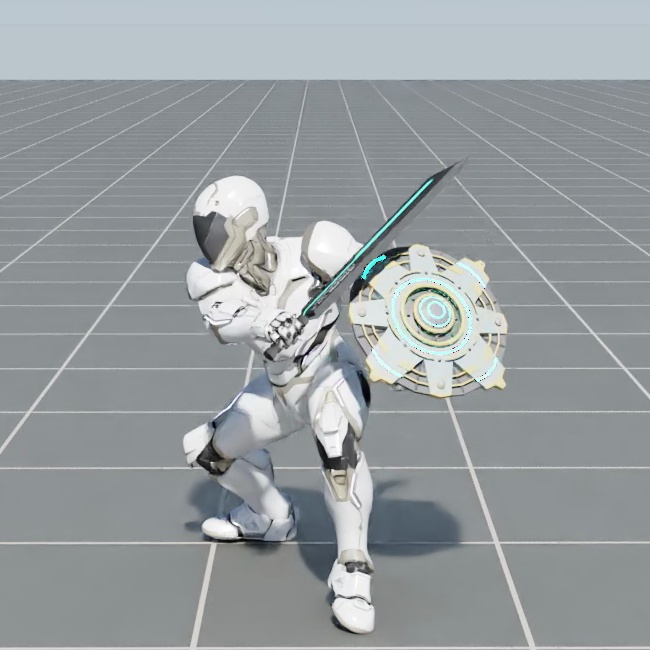}\hfill
         \includegraphics[width=0.165\textwidth]{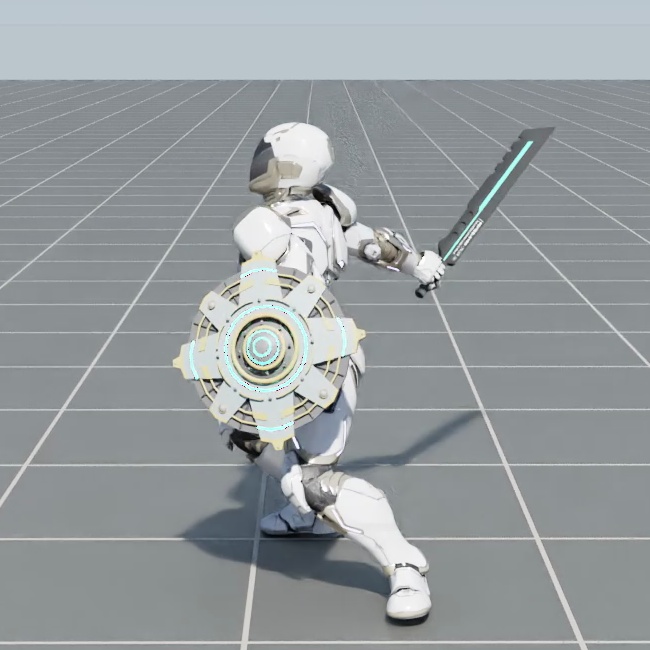}\hfill
         \includegraphics[width=0.165\textwidth]{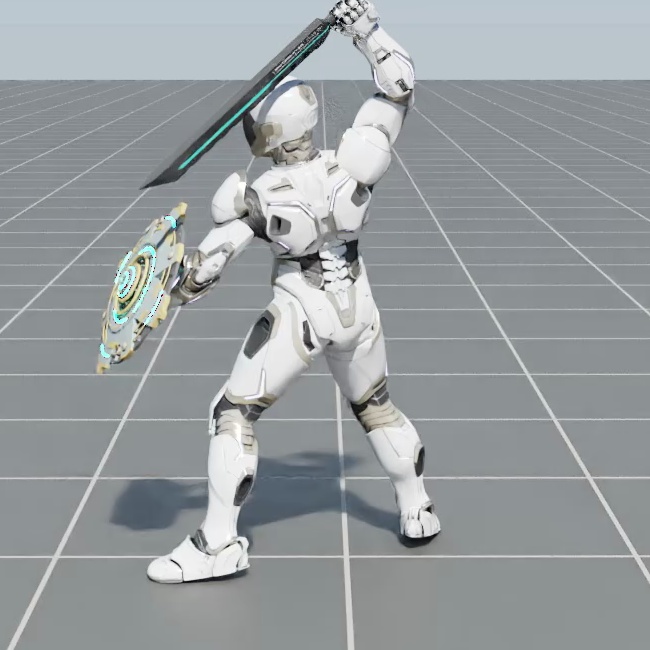}
         \caption{Double-Swipe Sword Combo}
         \label{fig llc: combo}
     \end{subfigure}
     \begin{subfigure}[b]{\textwidth}
         \centering
         \includegraphics[width=0.082\textwidth]{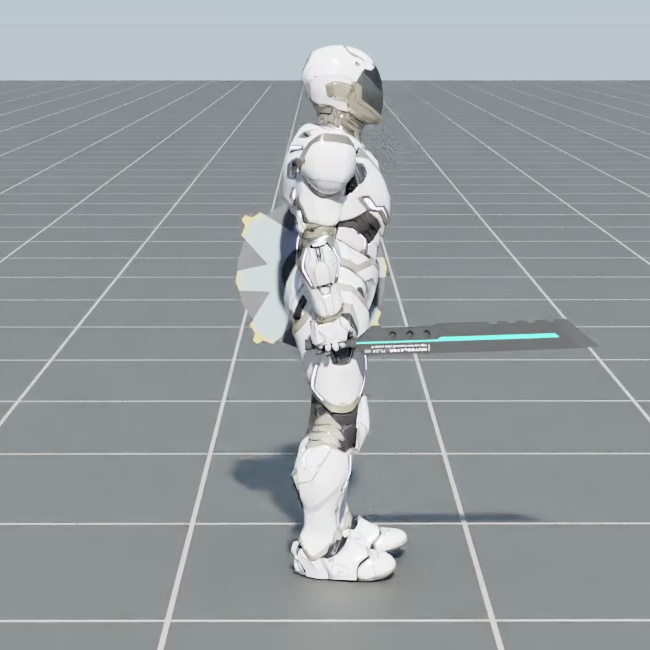}\hfill
         \includegraphics[width=0.082\textwidth]{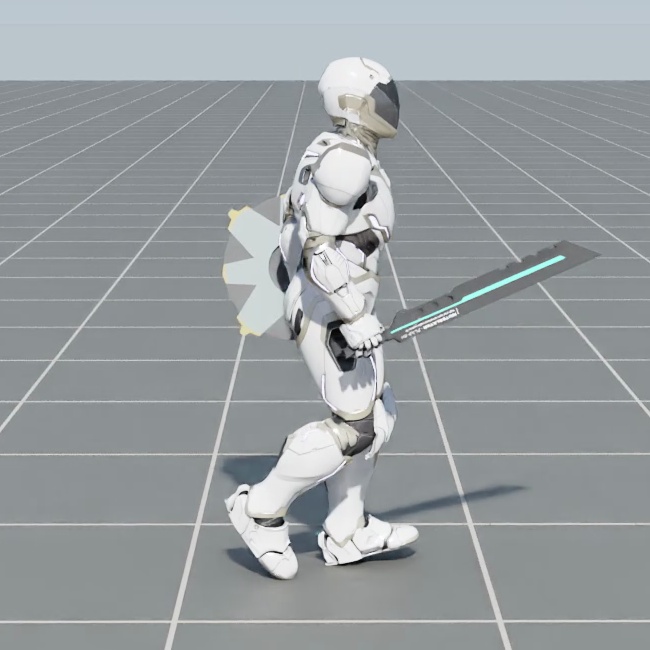}\hfill
         \includegraphics[width=0.082\textwidth]{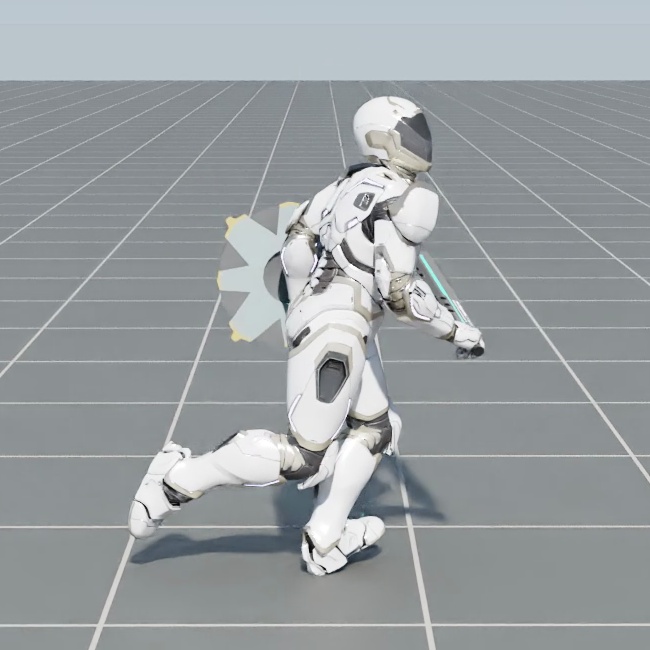}\hfill
         \includegraphics[width=0.082\textwidth]{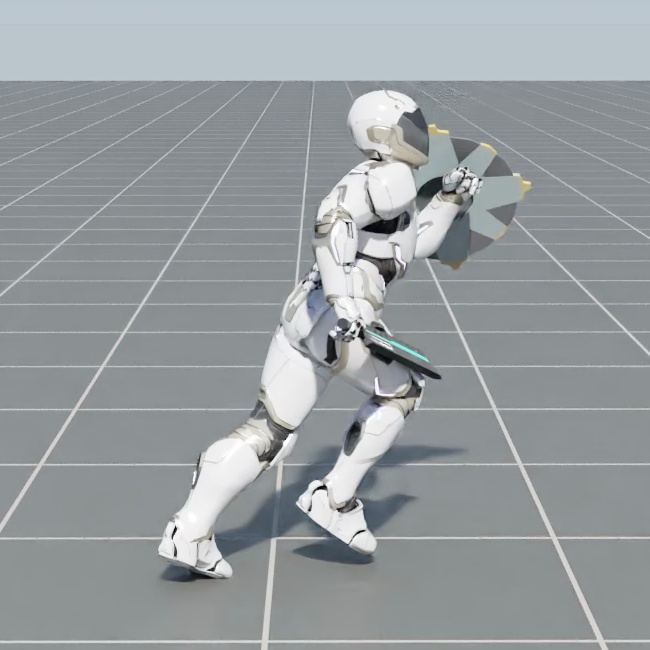}\hfill
         \includegraphics[width=0.082\textwidth]{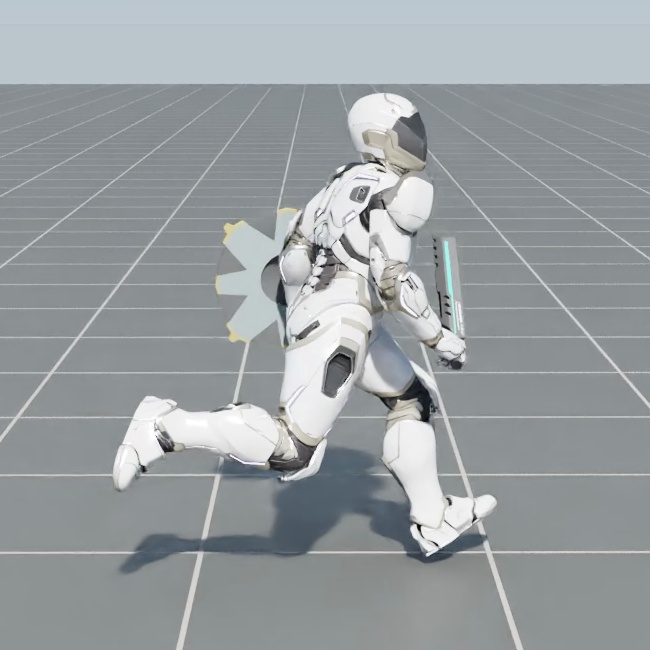}\hfill
         \includegraphics[width=0.082\textwidth]{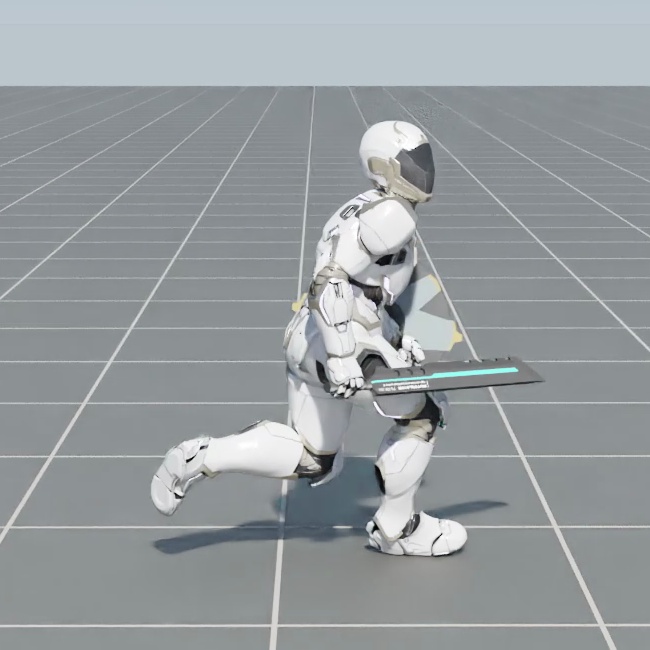}\hfill
         \includegraphics[width=0.082\textwidth]{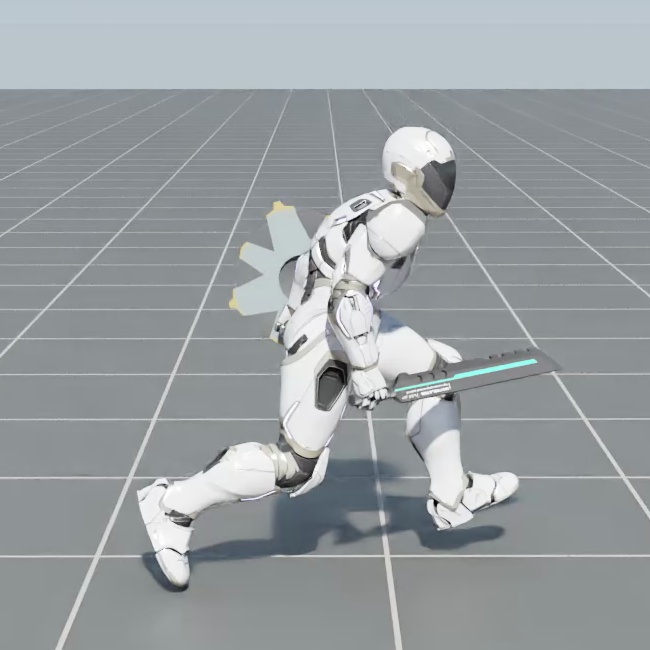}\hfill
         \includegraphics[width=0.082\textwidth]{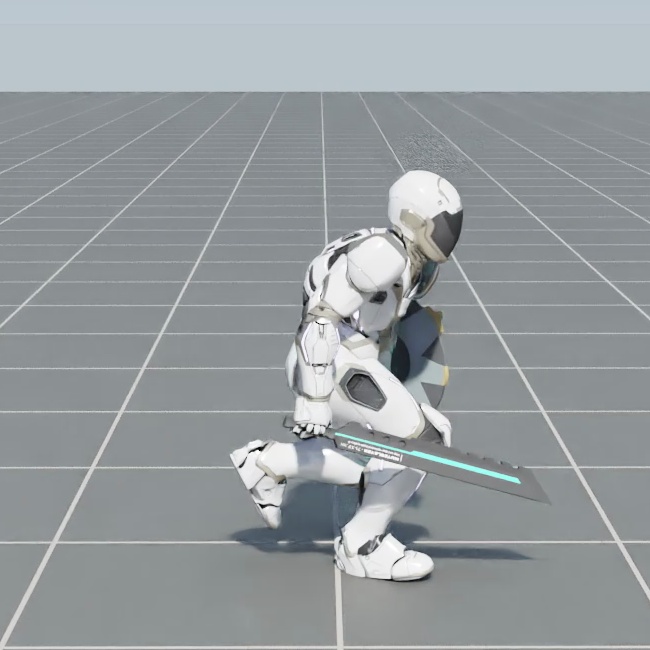}\hfill
         \includegraphics[width=0.082\textwidth]{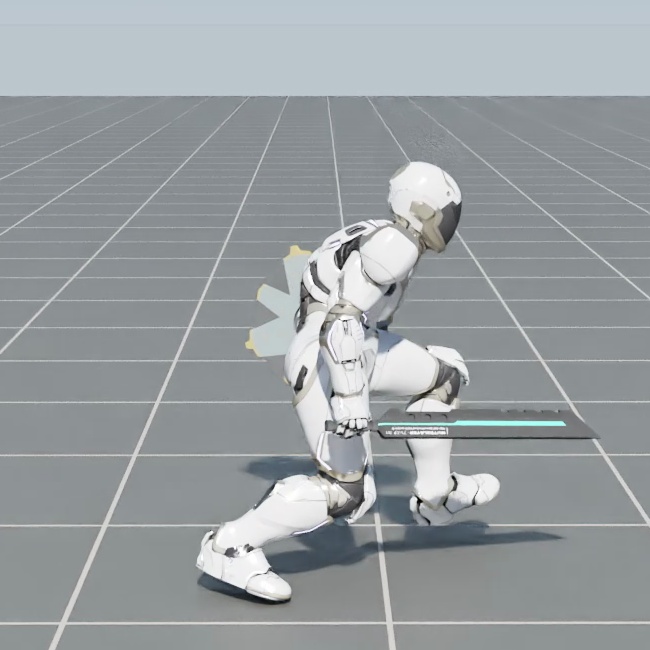}\hfill
         \includegraphics[width=0.082\textwidth]{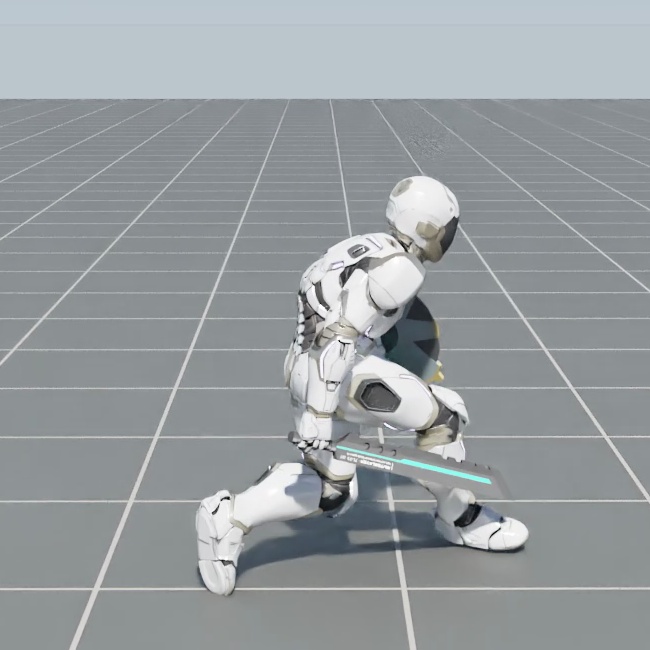}\hfill
         \includegraphics[width=0.082\textwidth]{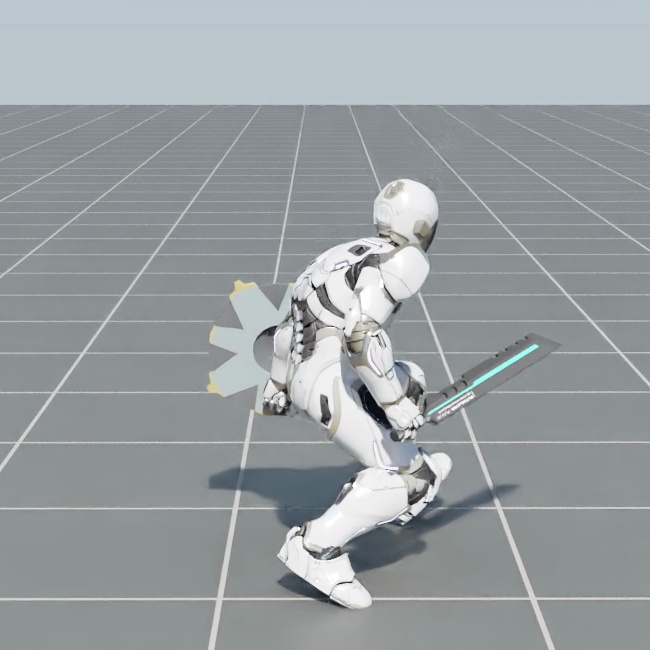}\hfill
         \includegraphics[width=0.082\textwidth]{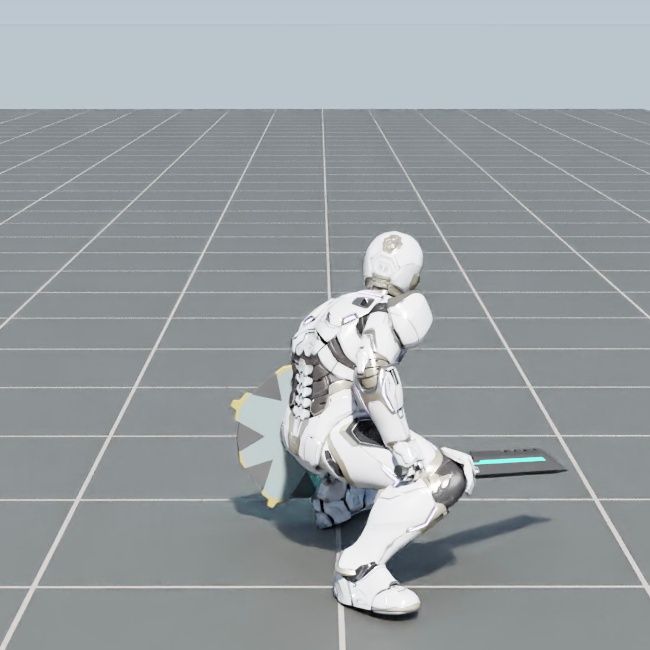}\\
         \includegraphics[width=0.082\textwidth]{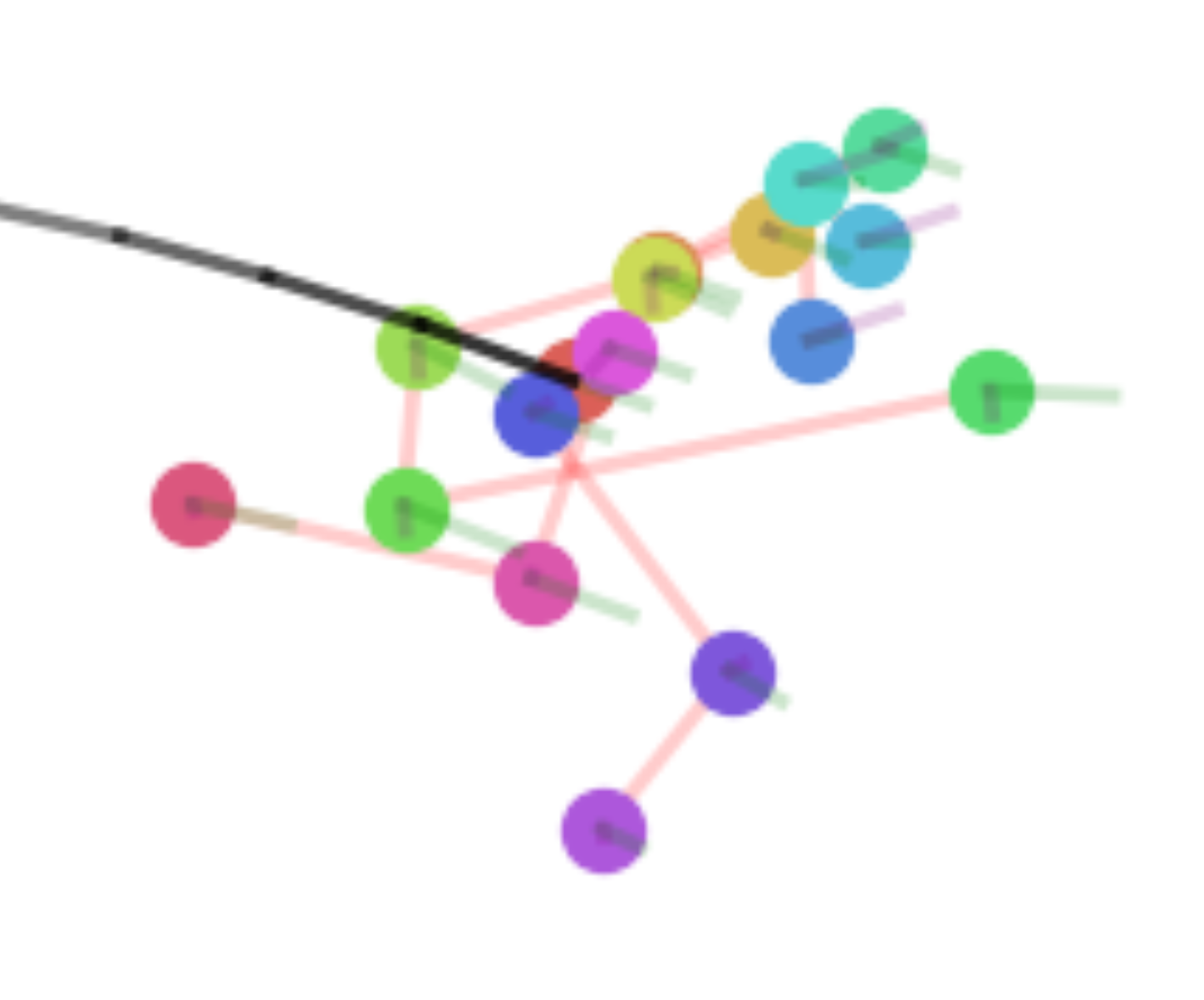}\hfill\includegraphics[width=0.83\textwidth]{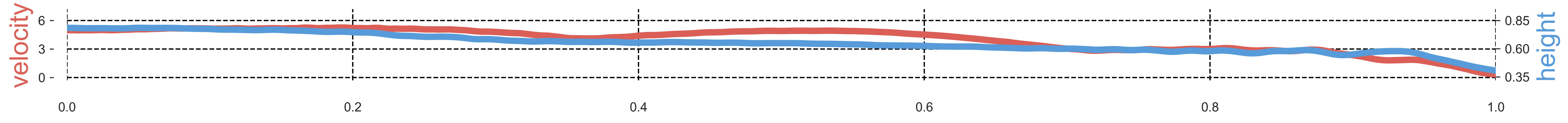}\hfill\includegraphics[width=0.082\textwidth]{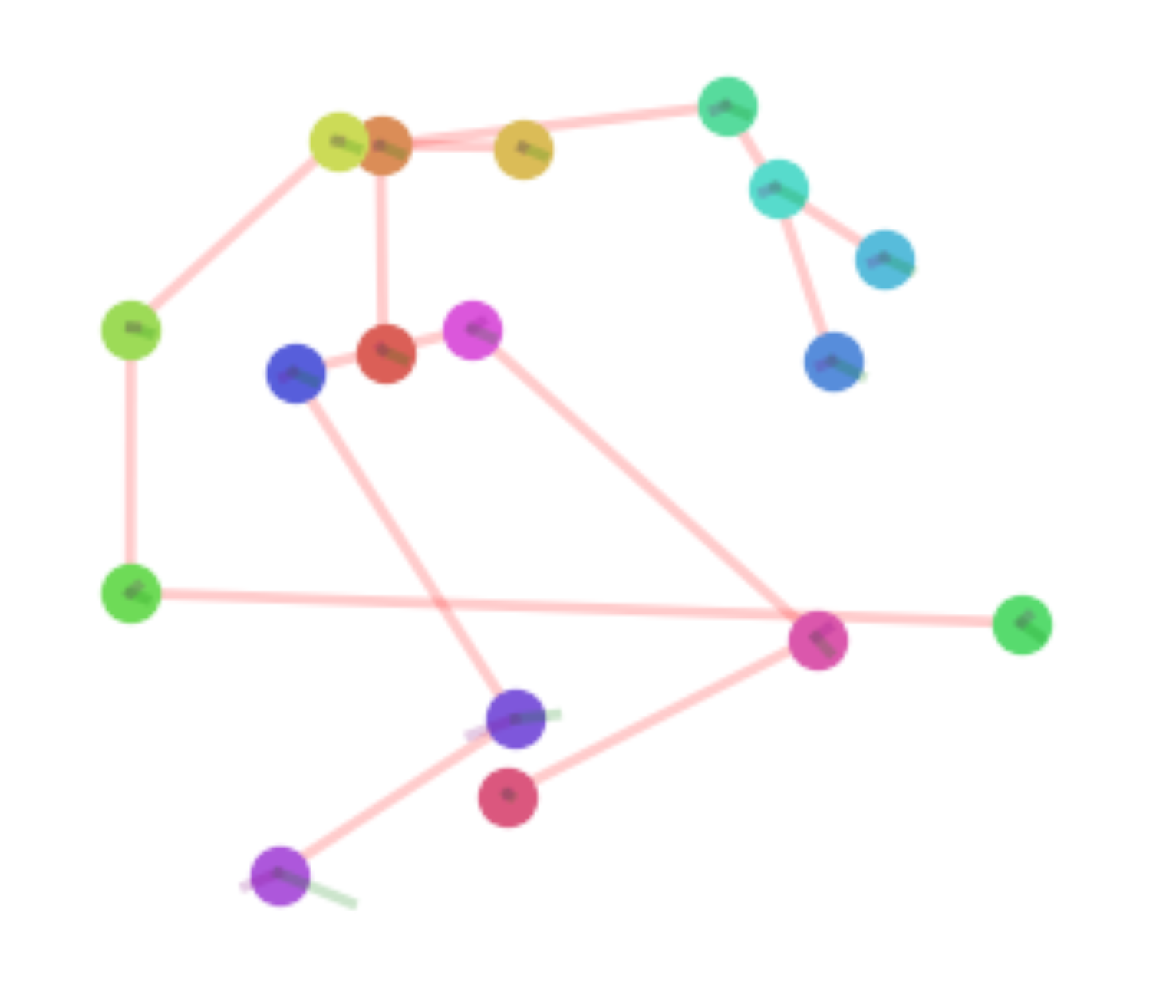}
         \caption{Interpolation over time: from sprinting to crouching-idle}
         \label{fig llc: interpolate}
     \end{subfigure}
    \caption{\textbf{Low-level training:} Skills generated by a low-level controller conditioned on the encoding of a demonstrated motion.}
    \label{fig: llc behaviors}
\end{figure}

\begin{figure}[!ht]
     \centering
     \begin{subfigure}[b]{0.498\textwidth}
         \centering
         \includegraphics[trim={7cm 2.4cm 0 3.9cm},clip,width=\textwidth]{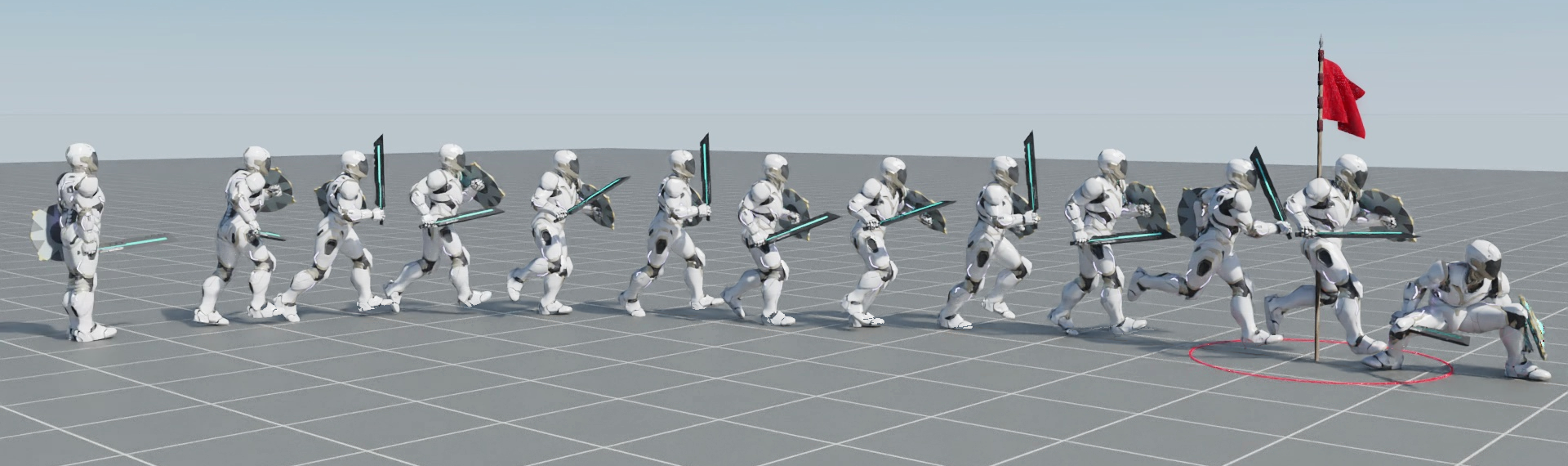}
         \caption{Location: Run, then Crouch-idle}
         \label{fig hrl: run location}
     \end{subfigure}\hfill
     \begin{subfigure}[b]{0.498\textwidth}
         \centering
         \includegraphics[trim={0 1.2cm 0 3.5cm},clip,width=\textwidth]{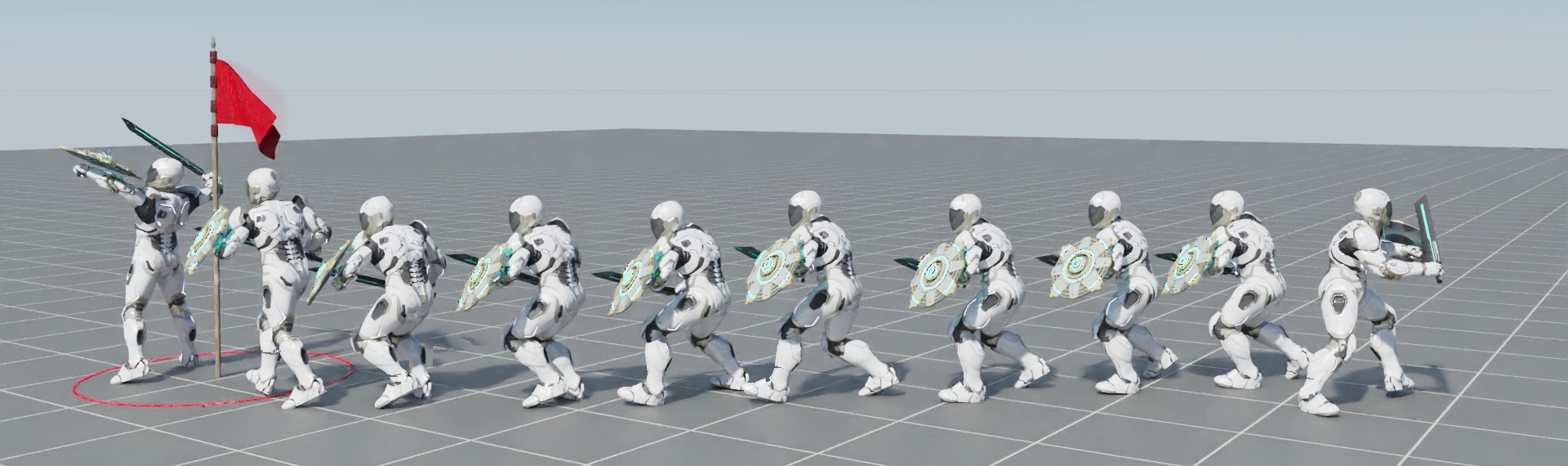}
         \caption{Location: Crouch-walk, then Roar}
         \label{fig hrl: crouch walk location}
     \end{subfigure}\\
     \begin{subfigure}[b]{0.498\textwidth}
         \centering
         \includegraphics[trim={0 1cm 0 4.6cm},clip,width=\textwidth]{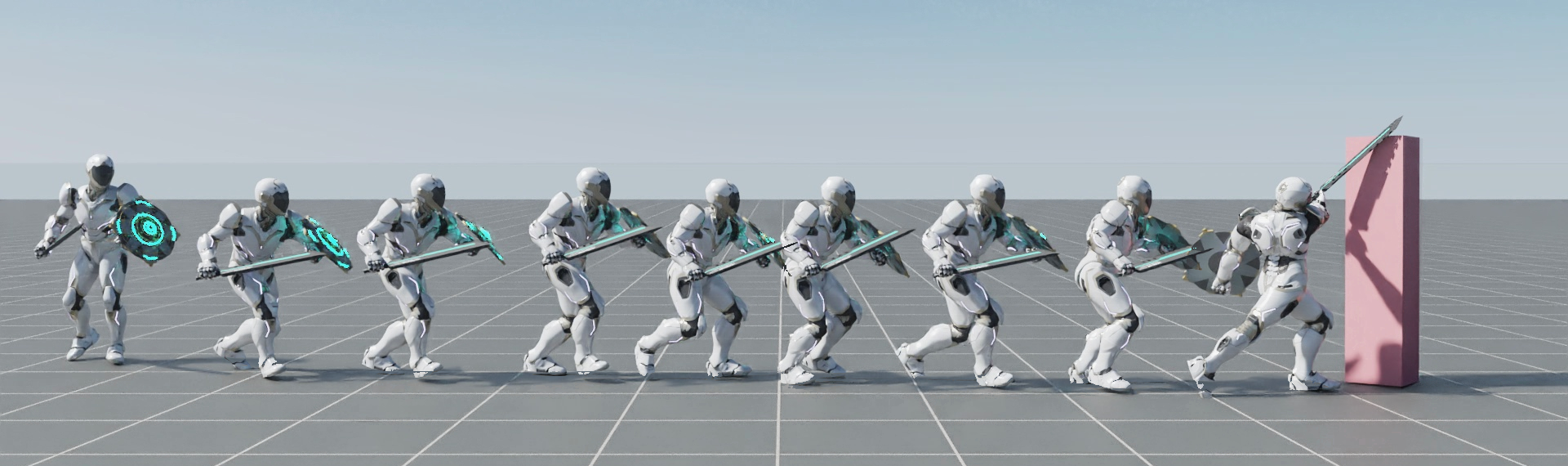}
         \caption{Strike: Crouch-walk, then Sword-swing left}
         \label{fig hrl: crouch walk sword swing strike}
     \end{subfigure}\hfill
     \begin{subfigure}[b]{0.498\textwidth}
         \centering
         \includegraphics[trim={0 1cm 0 4.6cm},clip,width=\textwidth]{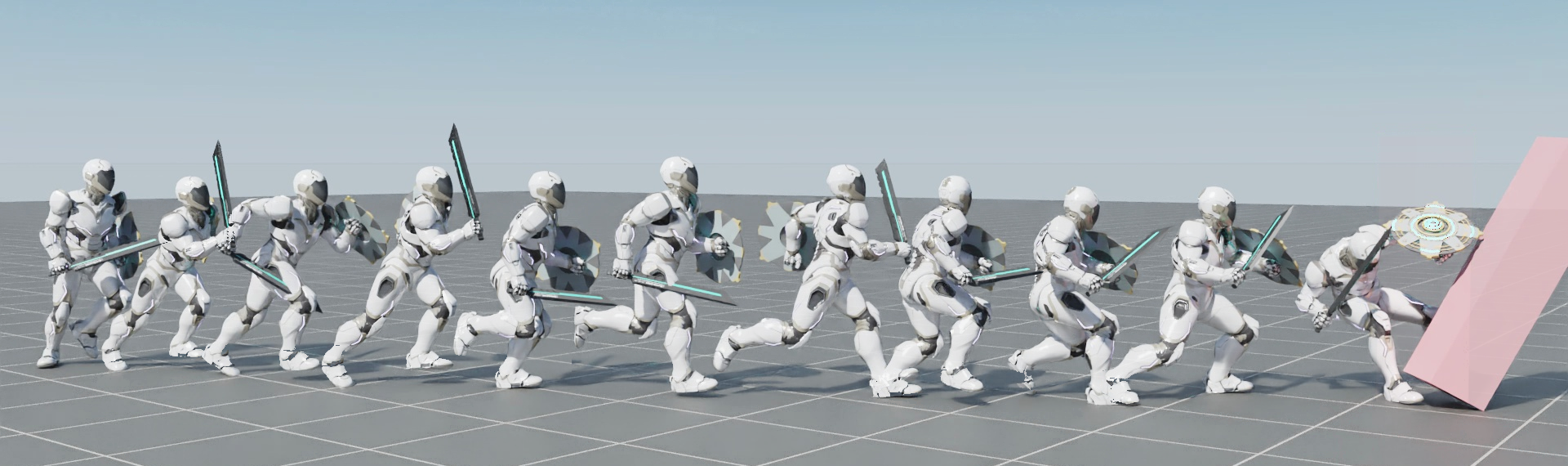}
         \caption{Strike: Run, then Shield-charge}
         \label{fig hrl: run shield charge strike}
     \end{subfigure}\\
     \begin{subfigure}[b]{0.498\textwidth}
         \centering
         \includegraphics[trim={0 1cm 0 4.6cm},clip,width=\textwidth]{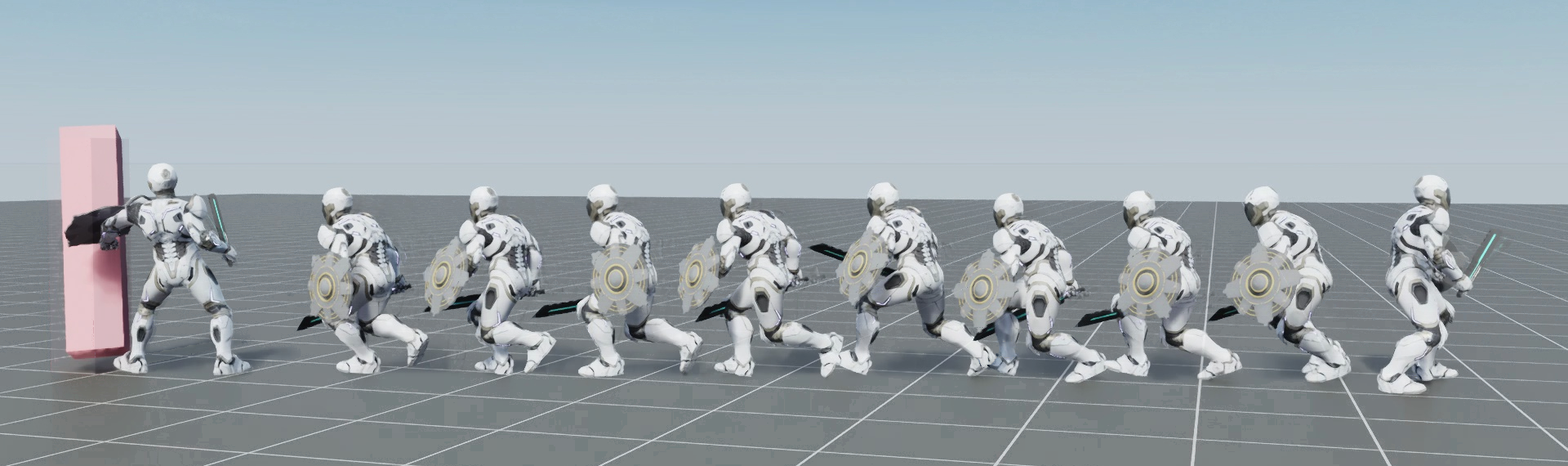}
         \caption{\edited{Strike: Crouch-walk, then Shield-swipe}}
         \label{fig hrl: crouch walk shield swipe strike}
     \end{subfigure}\hfill
     \begin{subfigure}[b]{0.498\textwidth}
         \centering
         \includegraphics[trim={0 1cm 0 4.6cm},clip,width=\textwidth]{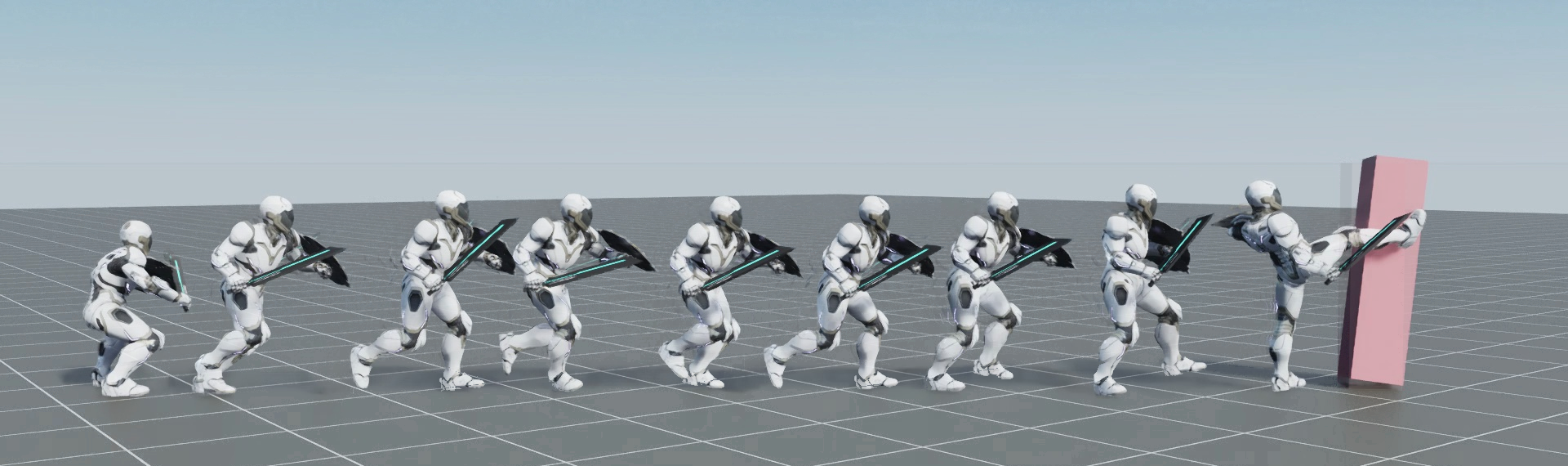}
         \caption{\edited{Strike: Walk, then Kick}}
         \label{fig hrl: walk kick strike}
     \end{subfigure}
    \caption{\textbf{Inference:} Without any further training, \alg~ solves multiple different tasks and the same task in multiple different forms.}
    \label{fig: hrl exemplar}
\end{figure}
\twocolumn

The results show that by constraining the latents to reside close to the reference motion encoding, the high-level policy is capable of generating motion in the specified style while ensuring it moves in the requested direction.


\subsubsection{Solving tasks without further training}\label{subsubsec: exemplar}

In our final experiment, we combine the directable low-level controller together with the high-level locomotion policy to provide zero-shot solutions to unseen tasks. We consider two tasks, location and strike. For the location task, the agent should reach and remain within the goal position -- illustrated as a circle around the flagpost. Strike, a more complex task, requires the agent to reach the target and strike it down. In both cases, the character is controlled by conditioning a sequence of reference motions. To do so, the direction vector is provided to the target location, represented in the character's local coordinate frame. Once within range, the low-level policy is directly provided the latent corresponding to the requested action, e.g., kick, shield charge, or sword swipe.

As shown qualitatively in \cref{fig: hrl exemplar} and \edited{quantitatively in \cref{tab: FSM results}}, \alg~ can be used to solve tasks similarly to how a human, given a game controller, would solve them. Thanks to the controllability aspect of \alg, without any further training or task-specific reward design, the FSM sequentially orders the character to transition between motions. The result is a composition of human-like motion that solves the task.


\section{Limitations}\label{sec: limitations}

\edited{The results in \cref{tab: LLC results,tab: FSM results} show how \alg~ learns a diverse repertoire of motions without sacrificing controllability. This can then be leveraged to learn style-conditioned locomotion and finally to compose motions for solving multi-step, unseen, tasks. In this section, we highlight open challenges and questions that arise from this work.}

\edited{Pre-training -- mode collapse: We have shown that our algorithm \alg~ improves the controllability of generated motions compared to the existing approach, ASE, with a significant boost in performance from 35\% up to 78\%. However, mode collapse remains an open challenge. For instance, we have observed that conditioning the low-level controller on idle-motions can lead to unrealistic micro-motions that slowly move the character. Although our approach addresses the problem of mode collapse to a large extent, there remains room for further improvements.}

\edited{Pre-training -- unseen motions: Our work focuses on learning a latent generative motion controller for in-distribution motions. However, when conditioning the character on encodings from unseen motions, we cannot guarantee the quality of the generated motions. While we have observed that some unseen motions map to semantically similar motions from the data, such as tip-toe mapping to bounce-walk, we anticipate that the model may fail as the motions become increasingly out of distribution.}

\edited{Precision-training -- beyond locomotion: In \cref{tab: FSM results} our approach is shown to leverage the learned latent space and achieve style-conditioned locomotion. However, controlling intricate movements such as the path of a sword or shield in an attack may require additional innovations in the pre-training phase, such as learning motions with a larger distributional discrepancy to the data.}

\edited{FSM -- robustness: Our approach has demonstrated the ability to solve tasks using classic tools from the gaming and animation industry, such as FSM and behavior trees, presenting a game-controller-like interface, as shown in \cref{fig: hrl exemplar}. However, we anticipate that the policy's robustness envelope may limit its ability to solve tasks with vastly different dynamics from those seen during training, such as climbing stairs or walking on uneven terrain. Therefore, further innovations and training may be required to solve such tasks.}

\edited{Rendering -- artifacts: To illustrate how our work can be integrated into the gaming industry, we visualize the motions using high-resolution characters, rendered within Omniverse (OV). Although physically accurate motion is recorded in IsaacGym (IG) where physical constraints are maintained without visual artifacts, the visualization character in OV is not subject to these constraints during rendering. Consequently, due to the difference in character geometry, the visualization character may exhibit penetrations and other visual artifacts that do not occur in IG. It is worth noting that the issue of rendering artifacts is not an inherent problem with our proposed algorithm \alg~, but due to the differences in character geometry. One way to minimize these artifacts is by ensuring that the simulated character's geometry is closer to that of the visualization. Another solution is robustifying the training process to handle varying character morphologies to directly control the visualization character while enforcing physical constraints.}

\section{Discussion and Future Work}

In this work, we presented \alg, a framework for learning reusable and directable motor skills for physics-based character animation. Our model enables the character's behaviors to be directed using motion clips. Given an unlabeled motion dataset, \alg~ learns both an encoder and a low-level controller. The encoder maps motions onto a semantically meaningful low-dimensional representation and a low-level controller takes the role of a decoder and produces motions with similar characteristics to those encoded within the learned representation. These reference motions can be used both for controlling low-level skills and to guide higher-level controllers and specify which motions the character should use when solving complex tasks. The ability to control the generated motion of the character enables zero-shot solutions to complex multi-step tasks, a step towards real integration of interactive virtual characters.

Our motion-constrained training enabled guiding the solution towards utilizing pre-specified motions. However, successfully learning to produce the requested motion in the specified direction required delicate tuning of the reward parameters. We are interested in exploring ways for disentangling the representation of direction from the representation of the content motion. Such disentanglement will enable high-level policies to learn with simplified rewards, while ensuring that they produce motion with the desired characteristics. Finally, some motions require coordinated interaction with the environment. For instance, aerobatic motions like handsprings and vaults can only be performed while interacting with a vaulting table/elevated box. We intend to investigate automation methods for understanding the motion-object pairs, and the integration of such objects throughout training for learning the respective motions.

\newpage

\bibliographystyle{ACM-Reference-Format}
\bibliography{bibliography}

\clearpage
\appendix

\section{State and action space}

In this work, we consider a 3D physically-simulated humanoid character wielding a sword and a shield, with 37 degrees of freedom. A similar character was used in \citet{peng2022ase}. To encode the state, we follow the same representation technique from \citet{peng2021amp}. The agent observes:
\begin{itemize}
    \item Root (character's pelvis) height.
    \item Root rotation with respect to the character's local coordinate frame.
    \item Local rotation of each joint.
    \item Local velocity of each joint.
    \item Positions of hands, feet, shield, and the tip of the sword, all in the character's local coordinate frame.
\end{itemize}

The character's local coordinate frame is defined with the origin located at the root, the x-axis oriented along the root link's facing direction, and the y-axis aligned with the global up vector. The 3D rotation of each joint is encoded using two 3D vectors, corresponding to the tangent \textbf{u} and the normal \textbf{v} of the link's local coordinate frame expressed in the link parent's coordinate frame \cite{peng2021amp}. In total, this results in a 120D state space.

\subsection{Low-level policy}

In addition, the low-level policy observes a 64D latent representation of motion.

To control the character, the agent provides an action $a$ that represents the target rotations for PD controllers, which are positioned at each of the character's joints. Similar to \citet{peng2021amp,peng2022ase,juravsky2022padl}, the target rotation for 3D spherical joints are encoded using a 3D exponential map \cite{grassia1998practical}, resulting in a 31D action space.

\subsection{Encoder}

The encoder learns to map fixed-length motions onto a low-dimensional representation. We consider 2-second motions. As the low-level controller operates at 30Hz, this results in 60 frames. During training, we randomly sample motion clips. If the motion clip is longer than 2 seconds, we randomly sample a continuous 2-second window and interpolate between the discrete supports of the motion. On the other hand, if the motion is shorter than 2 seconds, we apply zero padding.

\subsection{Discriminator}

The discriminator is trained similarly to ASE \cite{peng2022ase}. Given a 2-second motion clip $M$, the discriminator is conditioned on the corresponding latent encoding $z = E(M)$ and tasked with differentiating between $(\hat{s}_1, \ldots, \hat{s}_{10})$ randomly sampled from the reference motion $\mathcal{M}$ and the transition sequence $({s}_1, \ldots, {s}_{10})$ generated by the policy, also conditioned on the same latent $z$.

\subsection{High-level policy}

The high-level policy observes additional task information and produces latent variables $z \in \mathcal{Z}$ that are provided to the low-level policy.

\subsubsection{Block}
In the block task, a projectile is launched toward the agent. The agent is required to block the projectile using its shield. In each episode, the projectile has two phases. During the warmup phase, it remains static, followed by a launch phase in which it travels towards a target area in the character's body.

The high-level policy observes
\begin{itemize}
    \item The relative progress within this process.
    \item The location of the origin, in the character's local coordinate frame.
    \item The local of the projectile, in the character's local coordinate frame.
    \item The angle to the projectile.
    \item The projectile's velocity and its angular velocity.
\end{itemize}
this results in an additional 18D task-specific observation.

\subsubsection{Reach}
The goal of reach is to place the tip of the sword in a specified location and maintain this position. Here, the agent observes the 3D position of the goal location, in the character's local coordinate frame.

\subsubsection{Locomotion}
The locomotion task is a motion-conditioned task. Provided a set of motions $\tilde{\mathcal{M}}$, at time $t$, the high-level policy is tasked with moving in a specified direction, while utilizing motions with similar characteristics as $M_t \in \tilde{\mathcal{M}}$. To do so, it observes
\begin{itemize}
    \item The target direction, in the character's local coordinate frame.
    \item A one-hot encoding $|\tilde{\mathcal{M}}|$ of the current specified motion $M_t$.
\end{itemize}

\section{Architecture details}

In this work, all networks are composed of fully connected layers. The encoder and high-level policies are standard MLP networks. Prior to concatenating the latent representation with the observation, the low-level policy and the conditional discriminator also incorporate a pre-processing MLP for the latent variable $z$.

\section{Comparison of \alg~ to other physics-constrained motion generation frameworks}

Prior work has also considered the challenge of physical character animation. In this section, we compare our method \alg~ with prior work, highlighting the differences both in the method and the results.

\subsection{Motion matching}

Initial efforts in generating complex character motion focused on heuristics for motion matching. In DeepMimic \cite{peng2018deepmimic}, a reward is formulated for the deviation between the character's current pose and the corresponding pose in the data. This was later extended with motion VAEs enabled a certain degree of control by developing character controllers using the VAE paradigm. Notable efforts are \citet{ling2020character} and \citet{won2022physics} that learned, respectively, an unconditional and conditional motion VAE for motion reconstruction.

Citing \citet{won2022physics}, we highlight the main challenge in motion-matching schemes:

\say{our controller can also have the sinking problem mentioned in [Ling et al. 2020], where it fails to transition among different behaviors if the input trajectories lack such transitions. Using datasets composed of many heterogeneous behaviors where individual recordings are relatively short might cause this problem. Augmenting datasets by constructing motion graphs, which we used in our experiments, might be a quick remedy, however, preparing datasets rich in transitions would create the most natural-looking motions.}

As the low-level controller learns based on a state-reconstruction loss, it lacks the built-in ability to generalize beyond what was seen in the data. A core strength of adversarial techniques is that they force the behavior to be likely under the reference data distribution, enabling the emergence of new and novel movements such as transitioning between motion classes, as seen in \cref{fig: motion transitioning}.

On the other hand, a core benefit of methods aiming to directly mimic demonstrations is that the per-trajectory loss is highly informative of the agent's ability to reconstruct a given motion. This, which may be non-trivial in adversarial methods, can easily be leveraged for adaptive sampling \cite{park2019learning}, thus reducing mode-collapse and improving overall quality.

\subsection{Adversarial techniques}

More similar to our line of work are adversarial techniques. Here, the motion-matching reward is replaced with a discriminative signal. The discriminator produces a signal corresponding to how likely the generated motions are, given the data.

Initially, AMP \cite{peng2021amp} focused on solving single tasks using an unconditional policy and unconditional discriminator. The lack of conditioning prevents the ability to direct the behavior. Hence, optimizing the adversarial objective results in generating the entire reference data distribution. To force human-like solutions, the reference data distribution is carefully tuned to match the expected task statistics.

This effort was later extended by ASE \cite{peng2022ase} to split learning into two phases -- low-level training, followed by high-level training. First, a low-level conditional policy is trained with an unconditional discriminator on a diverse dataset with an information maximization term. This enabled learning diverse behaviors that can be controlled by fixing the conditioning variable $z$. However, the lack of correspondence between motions and latents results in mode-collapse, requiring tricks such as a diversity loss in order to improve the quality. As a result, controlling the motions generated by ASE is not trivial.

Finally, closest to our work, PADL \cite{juravsky2022padl} introduced a scheme composing both a conditioned policy and a discriminator. The main differences between \alg~ and PADL are that (1) PADL requires labeled data, which is both costly to acquire and less expressive than demonstrations. We do not assume any supervision beyond the ability to obtain sequences of motions. (2) PADL learns the representation separately from the control. We jointly learn the control and representation. (3) PADL directly learns a task-driven controller whereas we learn a general low-level conditional skill generator that can be re-used without further re-training.

As a result, \alg~ learns without supervision (1) a well-structured latent representation that captures the semantic meaning of motions. This is seen in the interpolation experiment \cref{fig llc: interpolate} where the character transitions smoothly through a semantically meaningful path. (2) \alg~ provides a more scalable approach, where the same low-level controller can be re-used for multiple tasks as shown in \cref{fig: hrl exemplar}.

\section{Qualitative analysis of the low-level controller}

In this section, we present some additional information regarding the quantitative results from \cref{sec: results llc}.

To analyze what \alg~ has learned, we performed two tests. The first aims to evaluate the encoder and what it has learned. We report the Fisher's distance, which measures the concentrability. Given a motion file $M$ and 2-second sliding window portions of it $M_i$
\begin{equation}
    coeff(M) = \frac{\sum_{M_i, M_j \in M}\frac{||E(M_i) - E(M_j) || }{|M|^2}}{\sum_{M_i \in M, M_k \in \mathcal{M}} \frac{||E(M_i) - E(M_k) || }{|\mathcal{M}| \cdot |M|}}
\end{equation}
and the score is defined as the average concentrability coefficient over motion files.

Next, we tested the ability of the low-level controller to generate motions on demand. We randomly selected a subset of 40 motions from the dataset. We conditioned \alg~ and ASE, each with their own respective encoder, to generate motion with similar characteristics. The initial state was randomly sampled from the data. Human raters were tasked with classifying each generated motion, where we generated 3 motions per reference clip.

\begin{figure}
    \centering
    \includegraphics[width=0.5\textwidth]{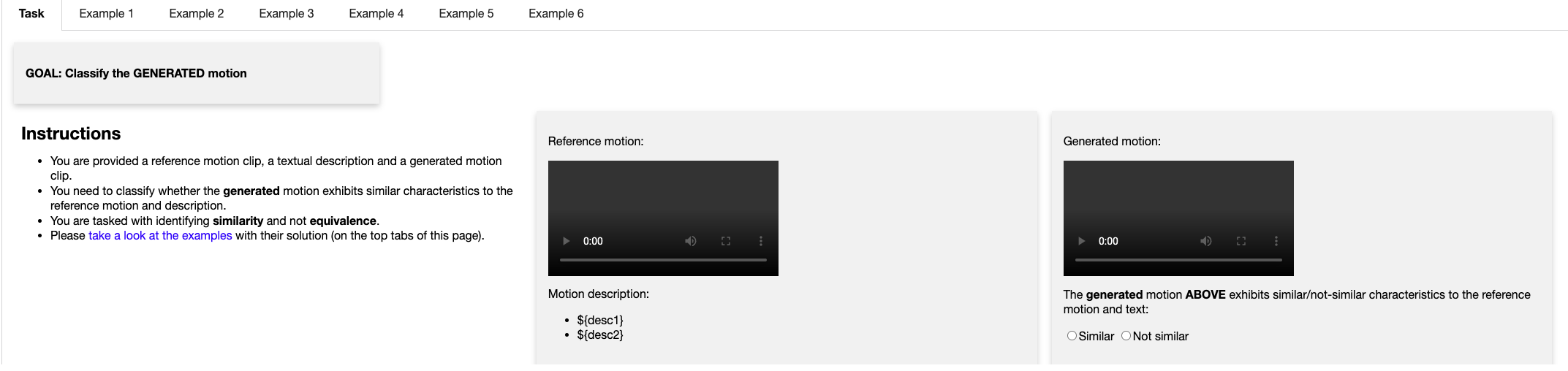}
    \caption{Motion classification system}
    \label{fig: screenshot - amt}
\end{figure}

We report the mean accuracy over all generated motions. In addition, we present a screenshot from our questionnaire in \cref{fig: screenshot - amt}.

\section{Additional experiments}

Complementing the experiments in the main paper, we present some additional results. The motions are best seen in the supplementary video.

\subsection{Ablation analysis}

\edited{To analyze the importance of the various design decisions for the pre-training phase, we perform an ablation experiment. We compare the full model, with removing negative samples from the discriminator training, and with removing the latent space regularization term.}

\edited{As shown in \cref{tab: ablation}, adding each design element further improves the generation quality of the model.}

\begin{table*}[t]
    \centering
    \begin{tabular}{l|c|c|c}
         & Concentration $\downarrow$ & Generation $\uparrow$ & Accuracy $\uparrow$ \\
         \toprule 
        \textbf{\alg} & \textbf{0.23} & \textbf{19.8$ \pm $0.11} & \textbf{78\%} \\
        \textbf{w/o negative samples} & 0.24 & 15.7$\pm$0.07 & 62\% \\
        \textbf{w/o negative samples, w/o regularization} & 0.35 & 12.8$\pm$0.05 & 61\% \\ \hline
    \end{tabular}
    \caption{\edited{\textbf{Pre-training ablation:} We perform ablation for various design choices for the pre-training phase.}}
    \label{tab: ablation}
\end{table*}

\subsection{Skill transitions}

To test the ability of the agent to transition between skills in a natural manner and on demand, we run the low-level policy on a long episode. During the episode, we sample a motion clip from the reference dataset and condition the agent on the corresponding latent.

\begin{figure*}[!ht]
     \centering
     \begin{subfigure}[b]{0.33\textwidth}
         \centering
         \includegraphics[width=0.333\textwidth]{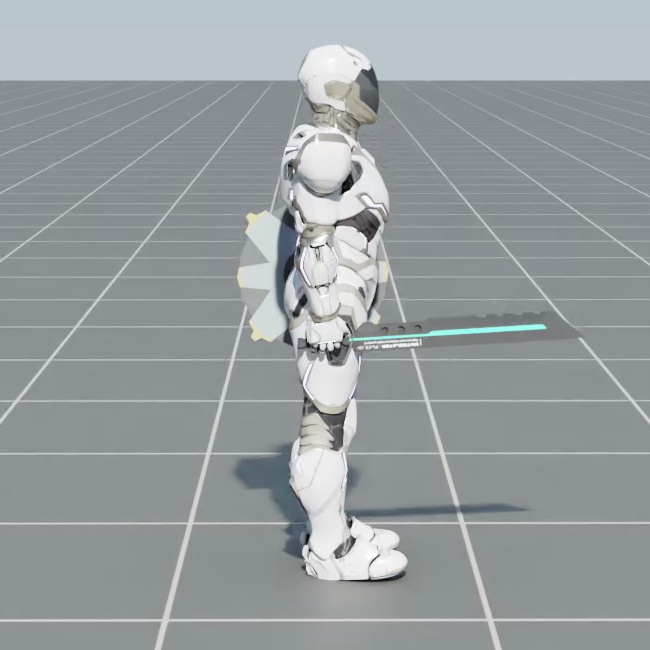}\hfill
         \includegraphics[width=0.333\textwidth]{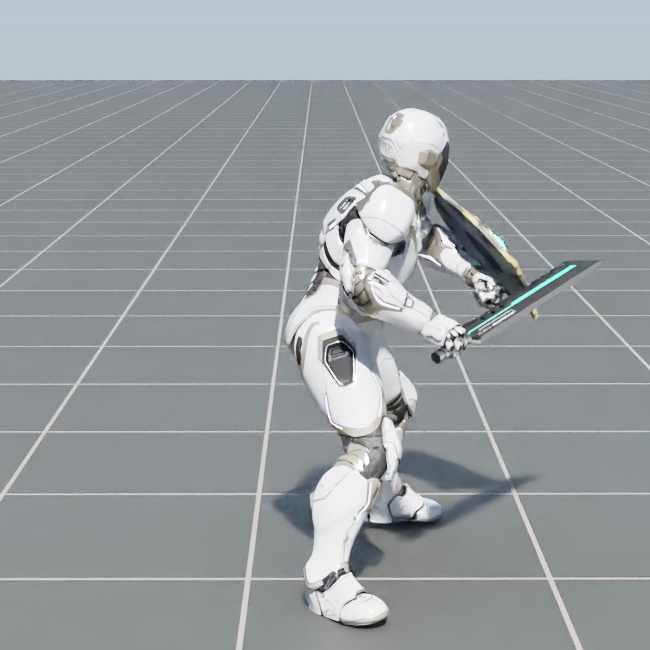}\hfill
         \includegraphics[width=0.333\textwidth]{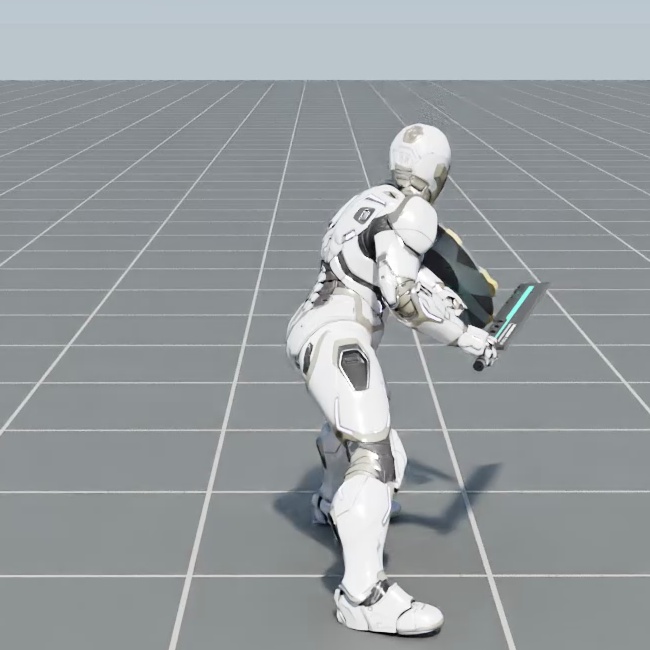}
         \caption{Idle, battle ready}
         \label{fig llc: apndx 1}
     \end{subfigure}\hfill
     \begin{subfigure}[b]{0.33\textwidth}
         \centering
         \includegraphics[width=0.333\textwidth]{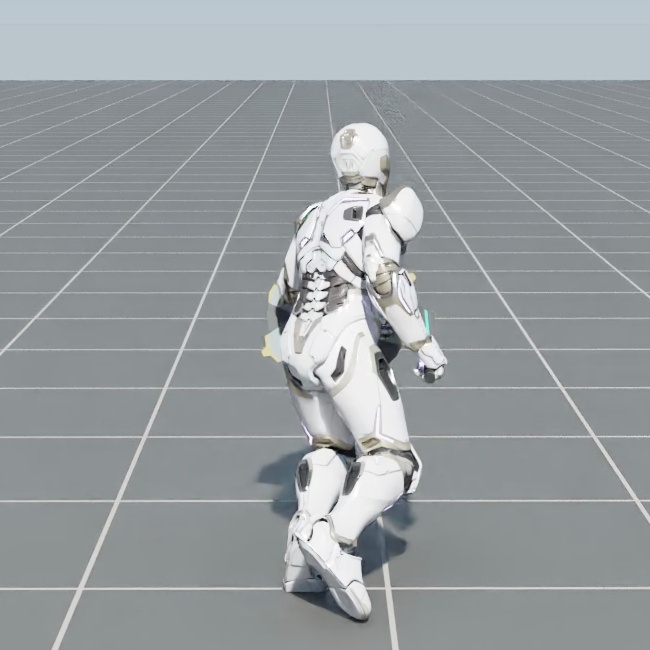}\hfill
         \includegraphics[width=0.333\textwidth]{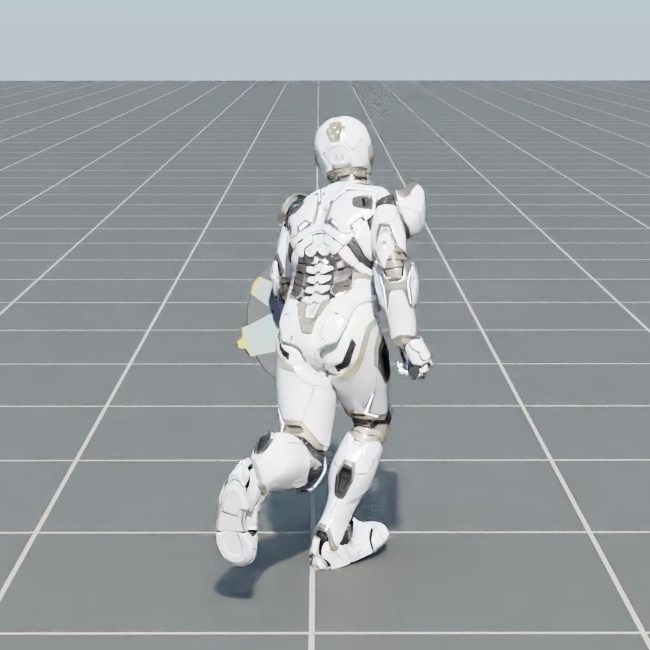}\hfill
         \includegraphics[width=0.333\textwidth]{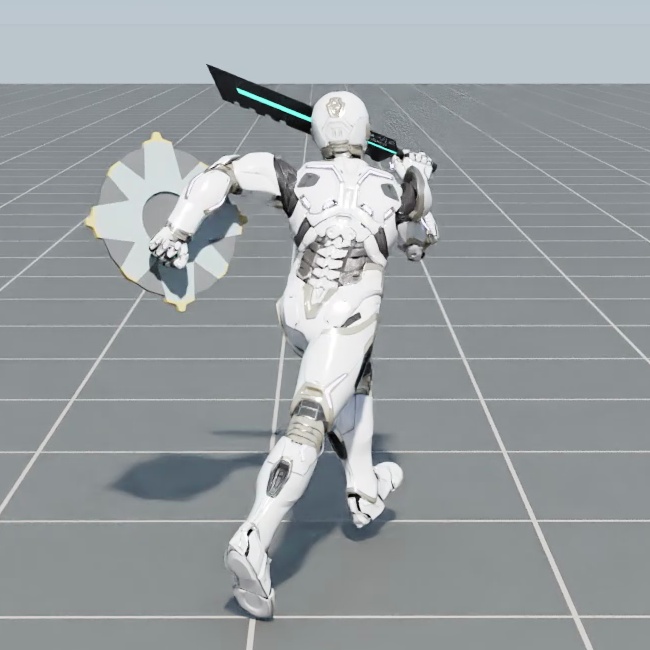}
         \caption{Sprint}
         \label{fig llc: apndx 2}
     \end{subfigure}\hfill
     \begin{subfigure}[b]{0.33\textwidth}
         \centering
         \includegraphics[width=0.333\textwidth]{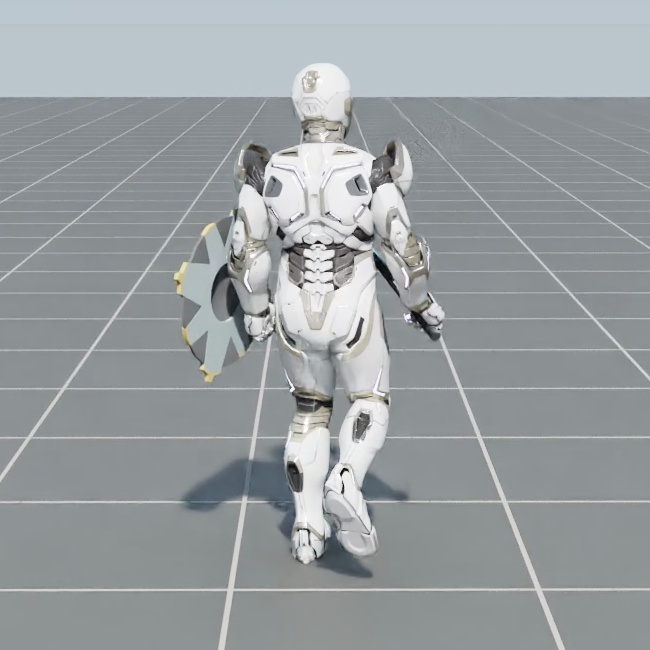}\hfill
         \includegraphics[width=0.333\textwidth]{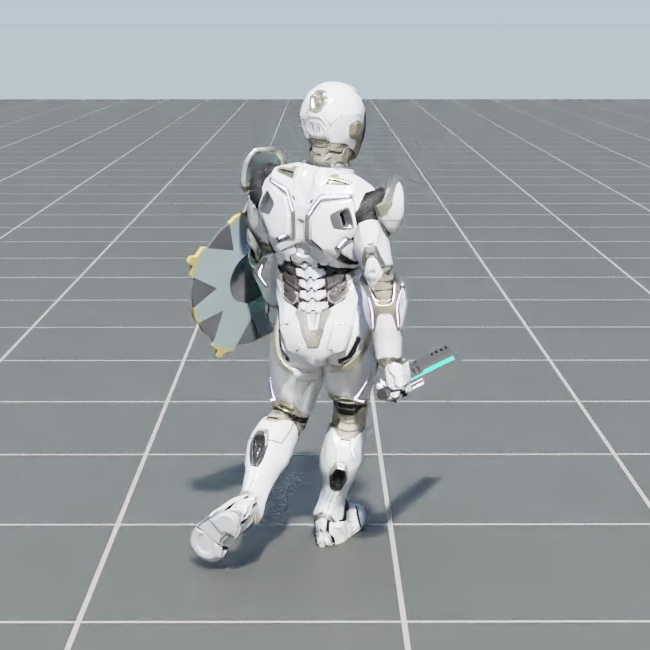}\hfill
         \includegraphics[width=0.333\textwidth]{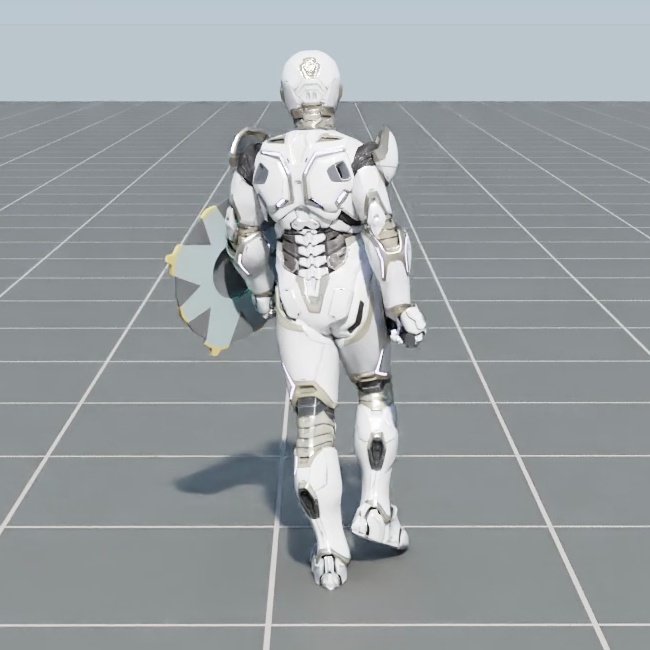}
         \caption{Walk slow}
         \label{fig llc: apndx 3}
     \end{subfigure}\\
     \begin{subfigure}[b]{0.33\textwidth}
         \centering
         \includegraphics[width=0.333\textwidth]{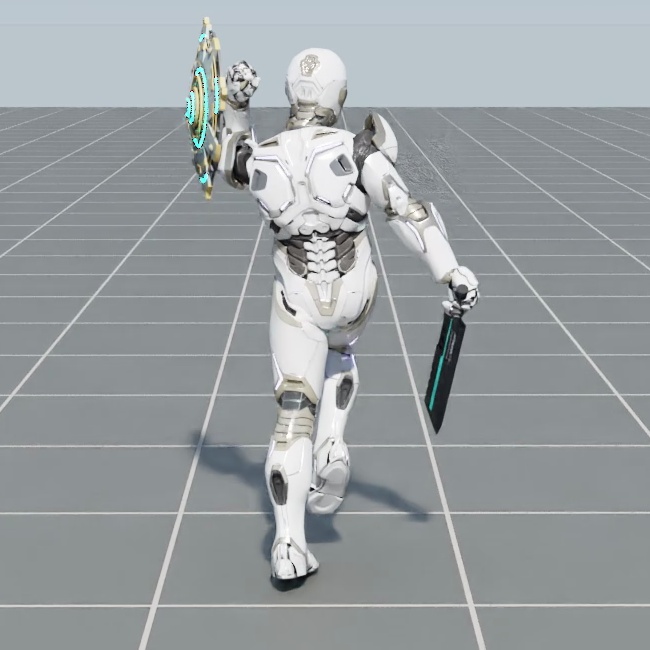}\hfill
         \includegraphics[width=0.333\textwidth]{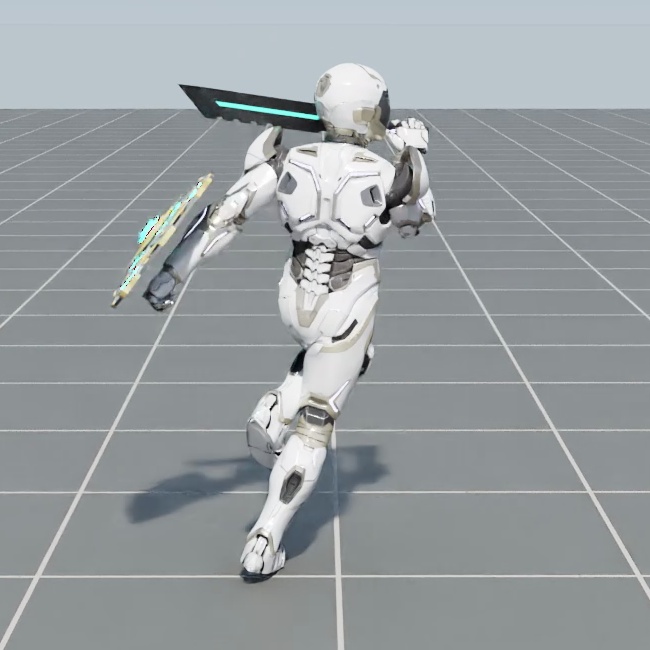}\hfill
         \includegraphics[width=0.333\textwidth]{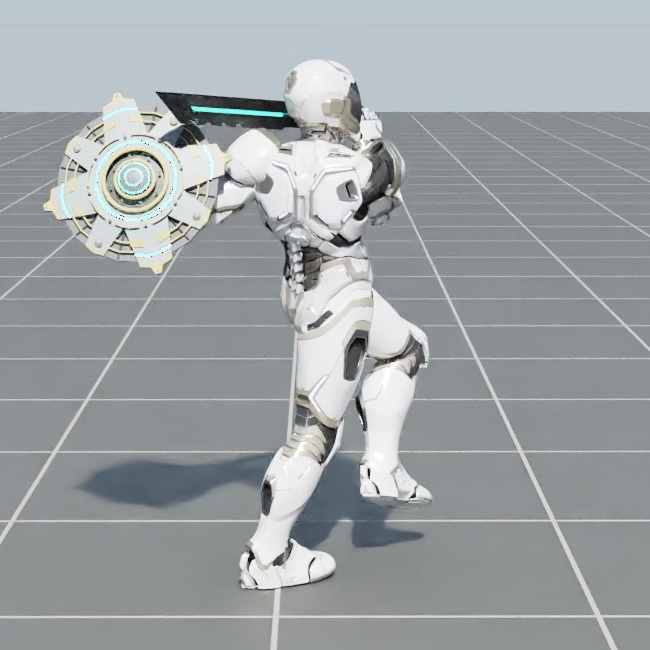}
         \caption{Bounce walking}
         \label{fig llc: apndx 4}
     \end{subfigure}\hfill
     \begin{subfigure}[b]{0.33\textwidth}
         \centering
         \includegraphics[width=0.333\textwidth]{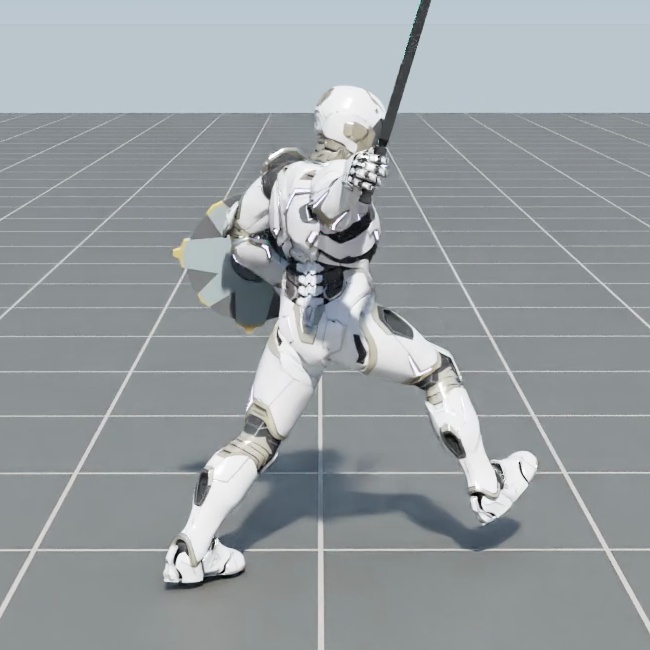}\hfill
         \includegraphics[width=0.333\textwidth]{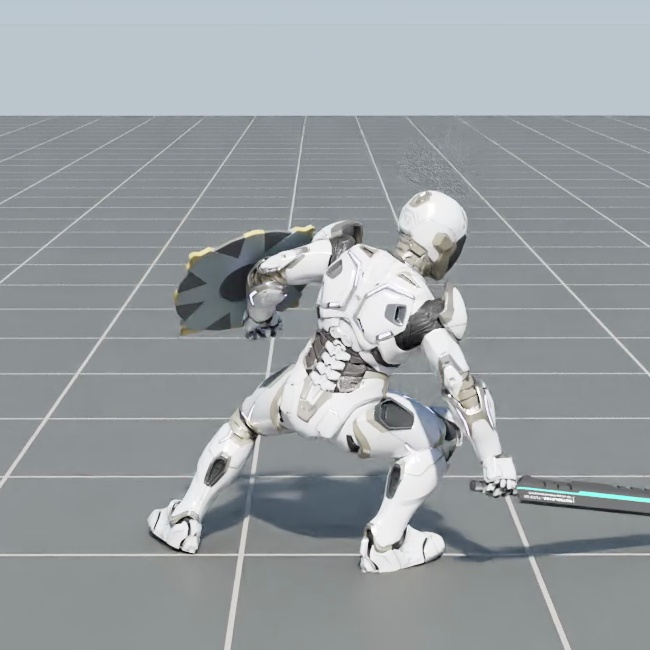}\hfill
         \includegraphics[width=0.333\textwidth]{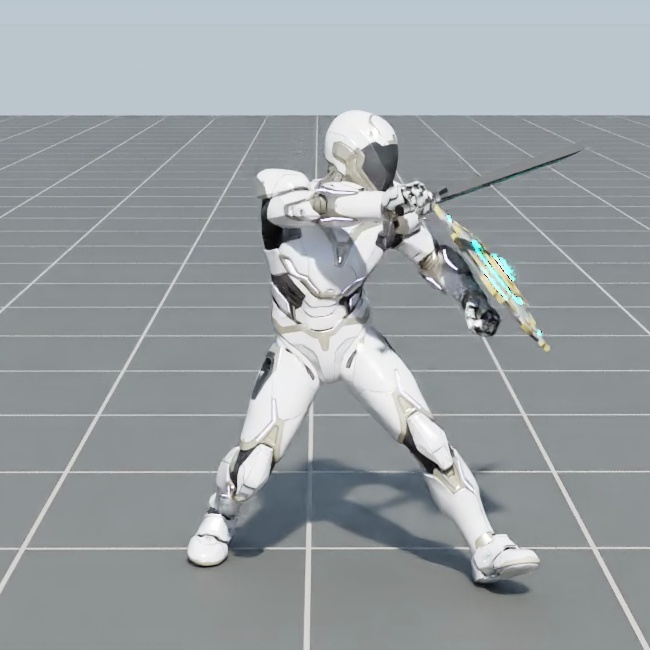}
         \caption{Attack combo}
         \label{fig llc: apndx 5}
     \end{subfigure}\hfill
     \begin{subfigure}[b]{0.33\textwidth}
         \centering
         \includegraphics[width=0.333\textwidth]{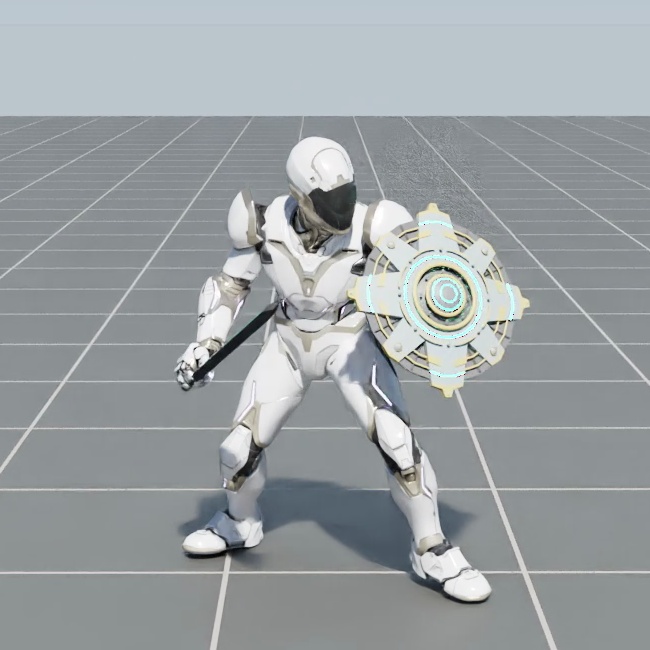}\hfill
         \includegraphics[width=0.333\textwidth]{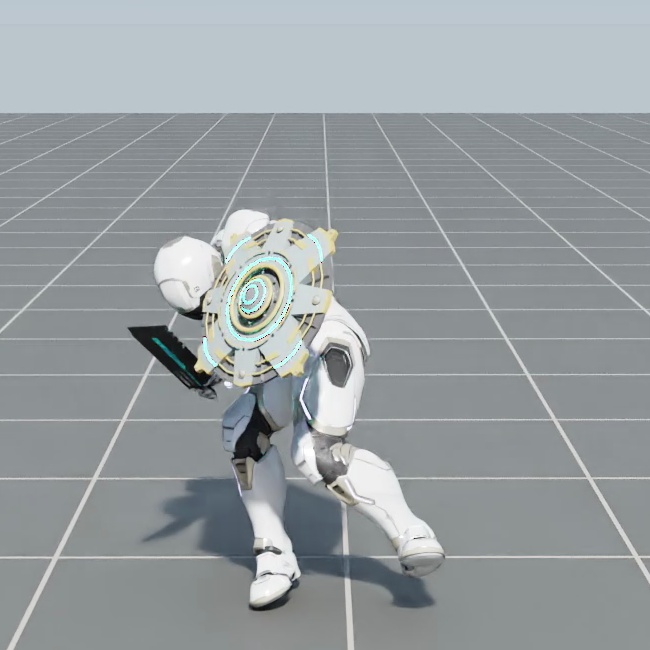}\hfill
         \includegraphics[width=0.333\textwidth]{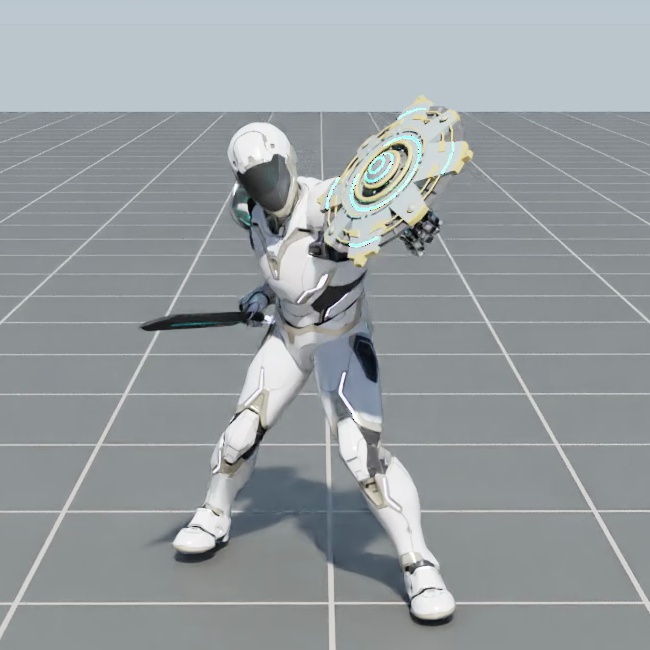}
         \caption{Shield charge}
         \label{fig llc: apndx 6}
     \end{subfigure}\\
     \begin{subfigure}[b]{0.33\textwidth}
         \centering
         \includegraphics[width=0.333\textwidth]{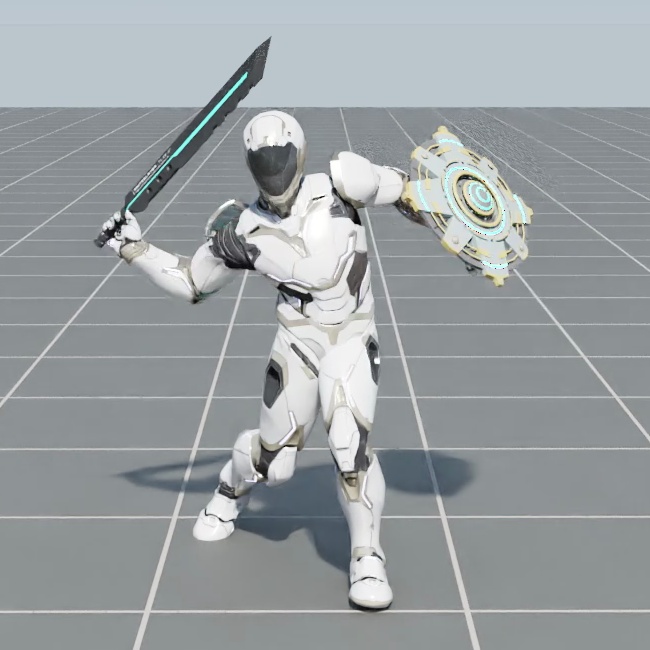}\hfill
         \includegraphics[width=0.333\textwidth]{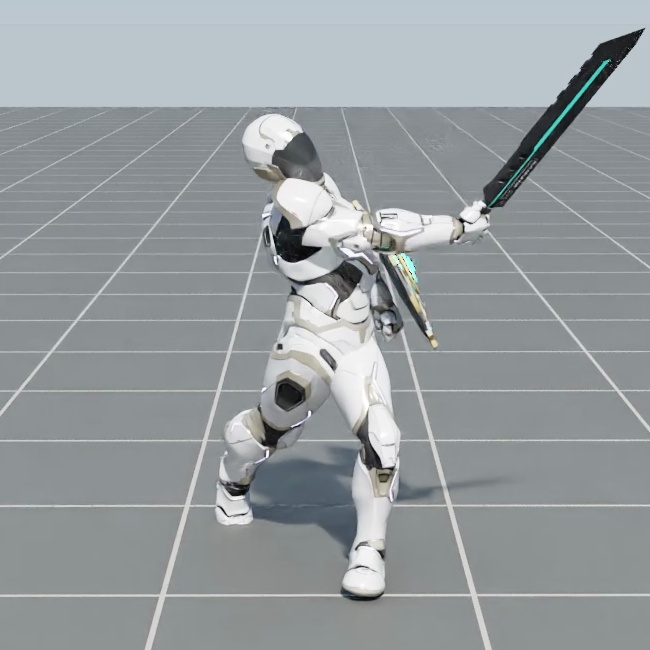}\hfill
         \includegraphics[width=0.333\textwidth]{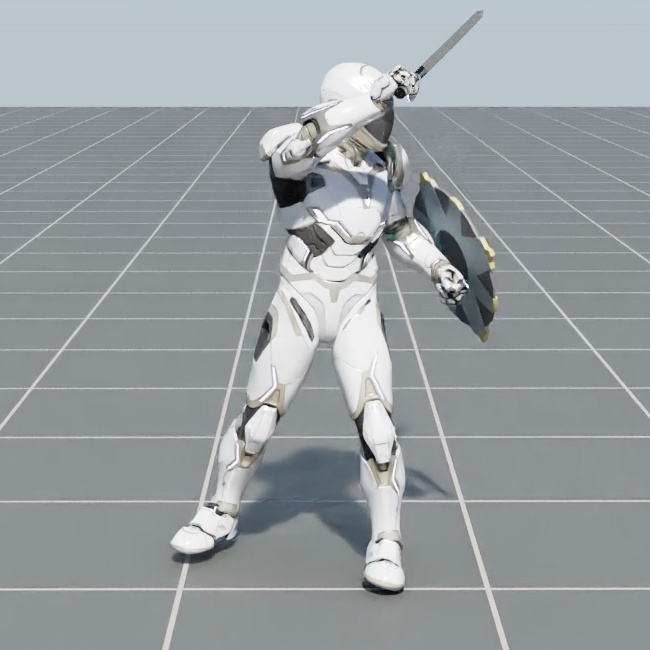}
         \caption{Sword swipe left}
         \label{fig llc: apndx 7}
     \end{subfigure}\hfill
     \begin{subfigure}[b]{0.33\textwidth}
         \centering
         \includegraphics[width=0.333\textwidth]{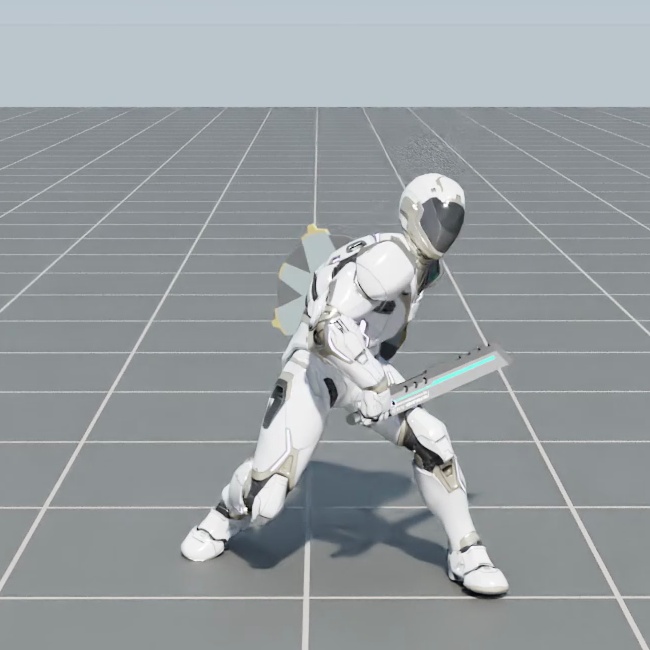}\hfill
         \includegraphics[width=0.333\textwidth]{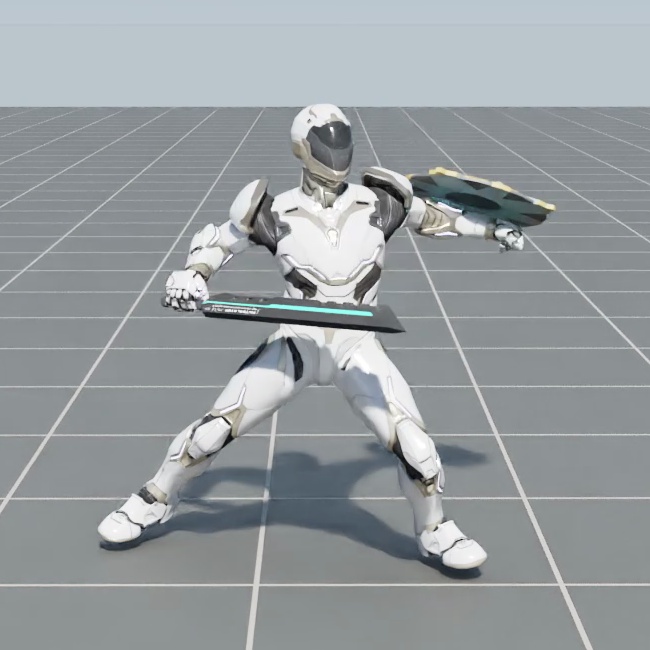}\hfill
         \includegraphics[width=0.333\textwidth]{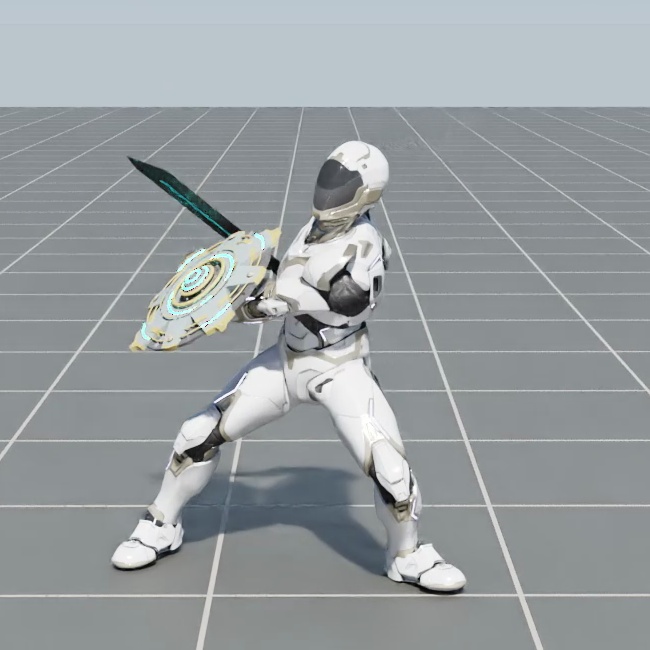}
         \caption{Attack combo}
         \label{fig llc: apndx 8}
     \end{subfigure}\hfill
     \begin{subfigure}[b]{0.33\textwidth}
         \centering
         \includegraphics[width=0.333\textwidth]{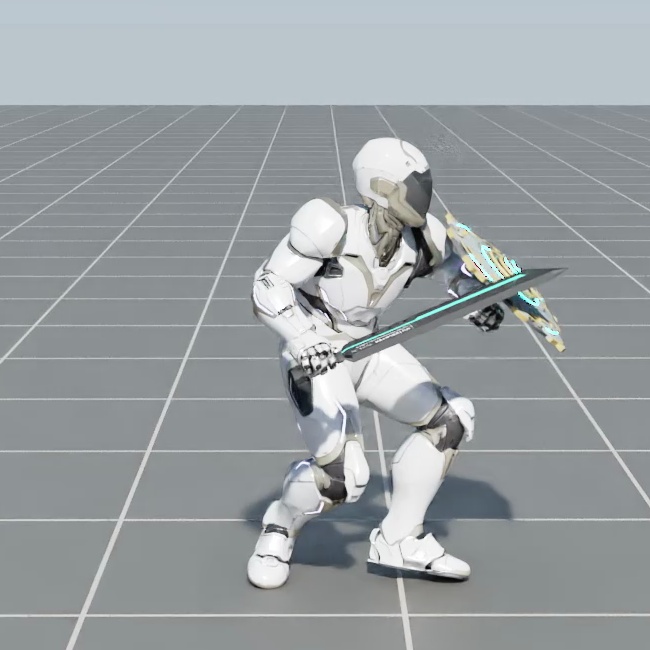}\hfill
         \includegraphics[width=0.333\textwidth]{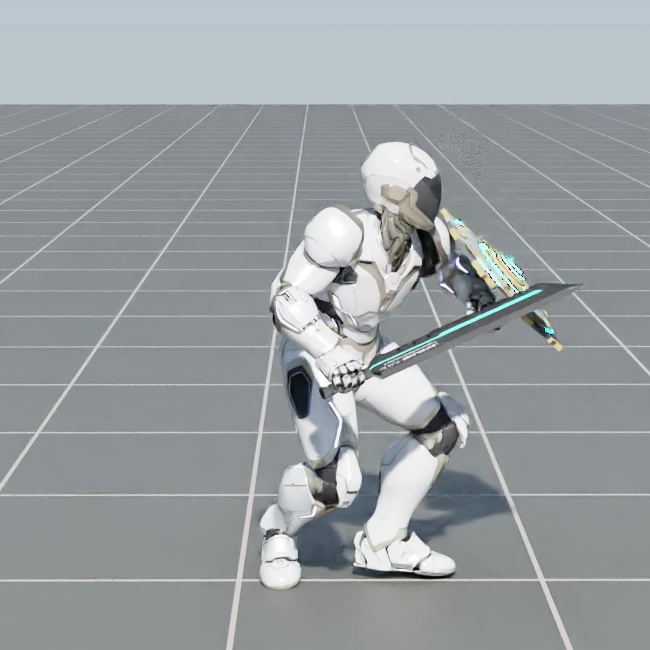}\hfill
         \includegraphics[width=0.333\textwidth]{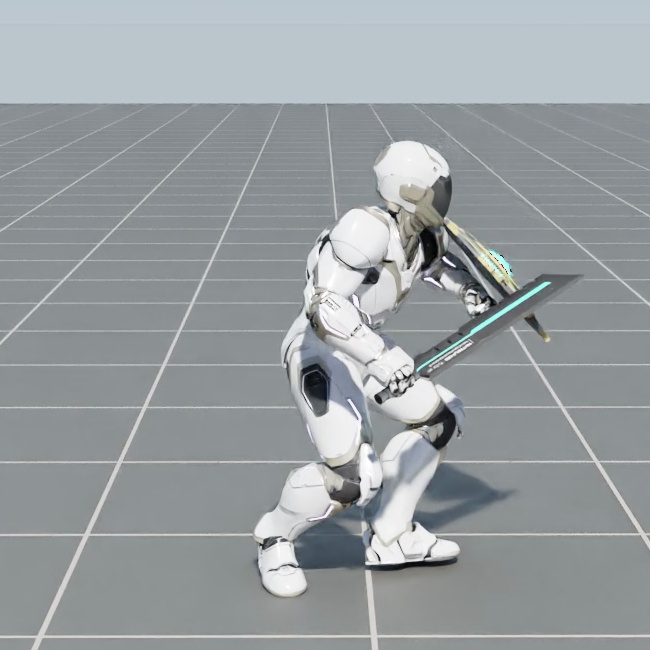}
         \caption{Walk backward}
         \label{fig llc: apndx 9}
     \end{subfigure}\\
     \begin{subfigure}[b]{0.33\textwidth}
         \centering
         \includegraphics[width=0.333\textwidth]{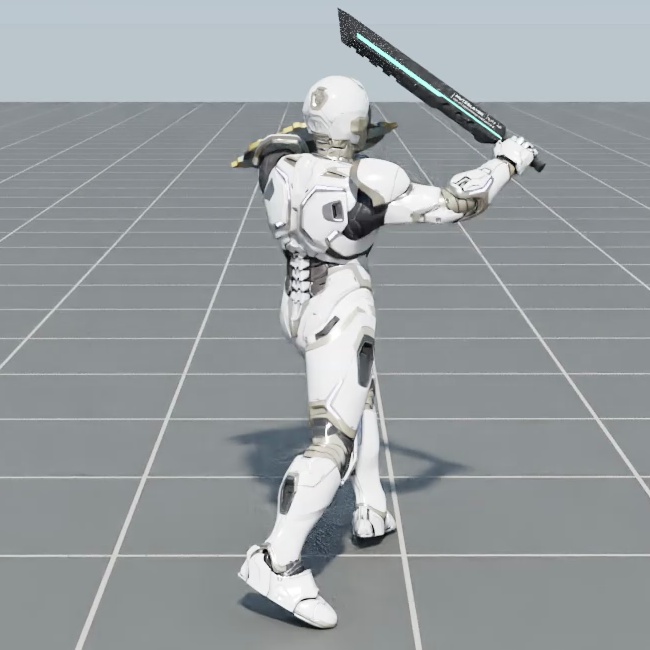}\hfill
         \includegraphics[width=0.333\textwidth]{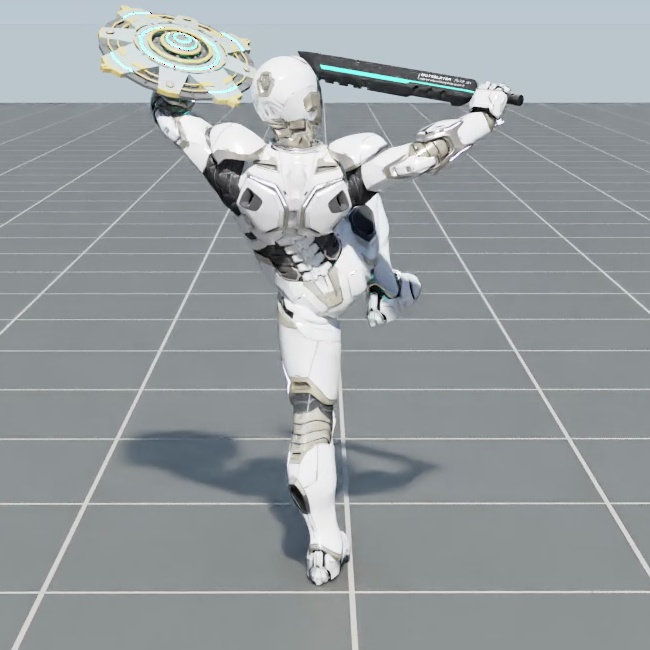}\hfill
         \includegraphics[width=0.333\textwidth]{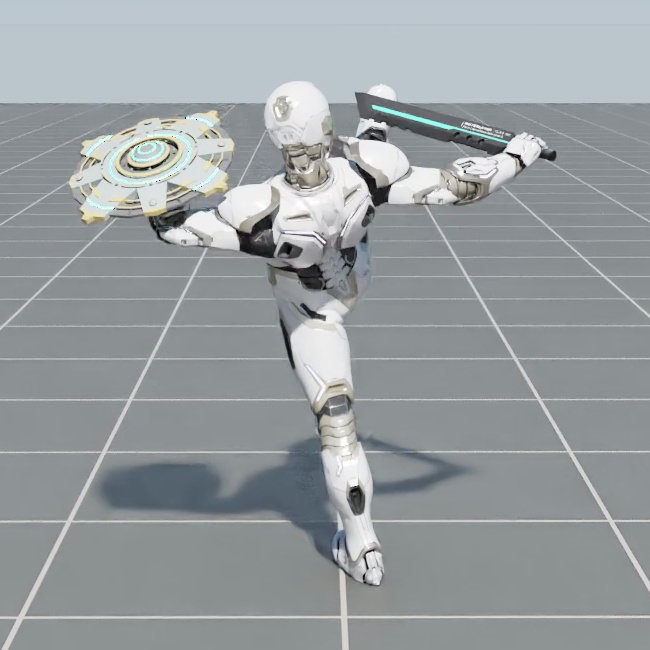}
         \caption{Kick}
         \label{fig llc: apndx 10}
     \end{subfigure}\hfill
     \begin{subfigure}[b]{0.33\textwidth}
         \centering
         \includegraphics[width=0.333\textwidth]{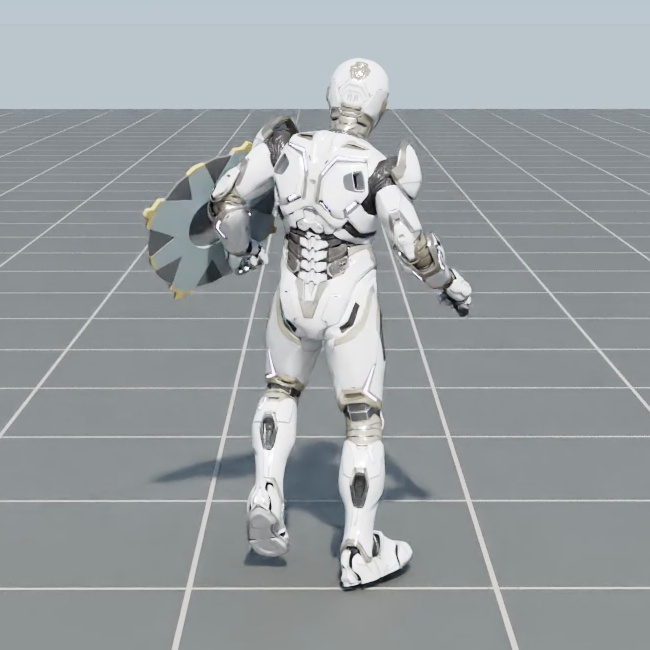}\hfill
         \includegraphics[width=0.333\textwidth]{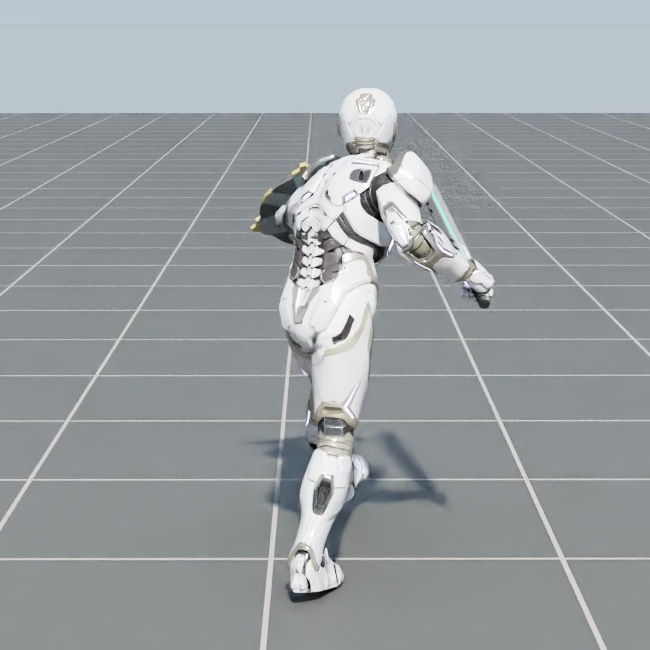}\hfill
         \includegraphics[width=0.333\textwidth]{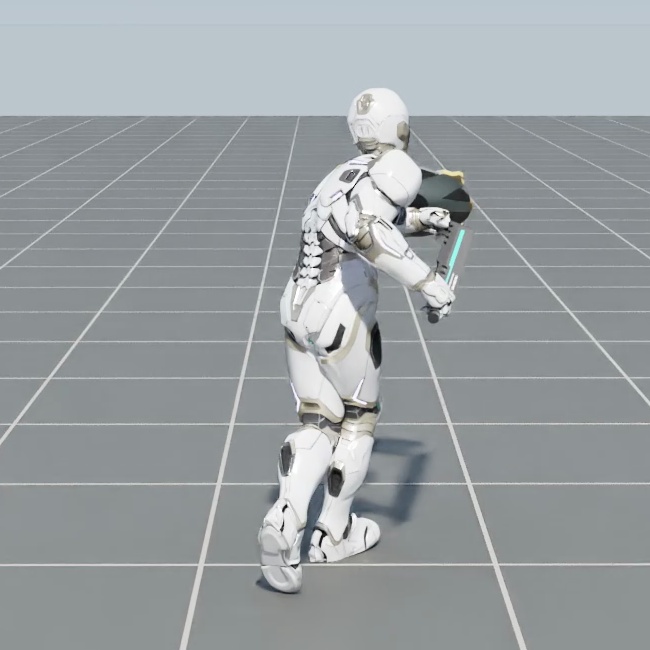}
         \caption{Run forward, shield raised}
         \label{fig llc: apndx 11}
     \end{subfigure}\hfill
     \begin{subfigure}[b]{0.33\textwidth}
         \centering
         \includegraphics[width=0.333\textwidth]{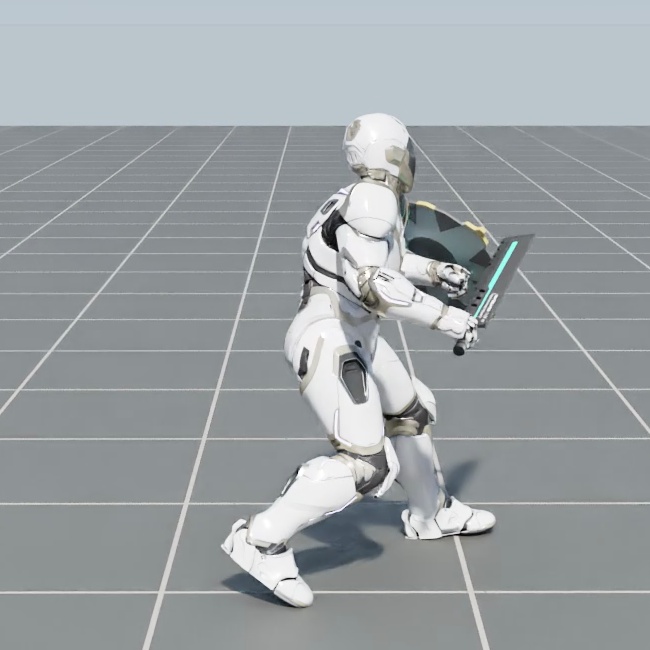}\hfill
         \includegraphics[width=0.333\textwidth]{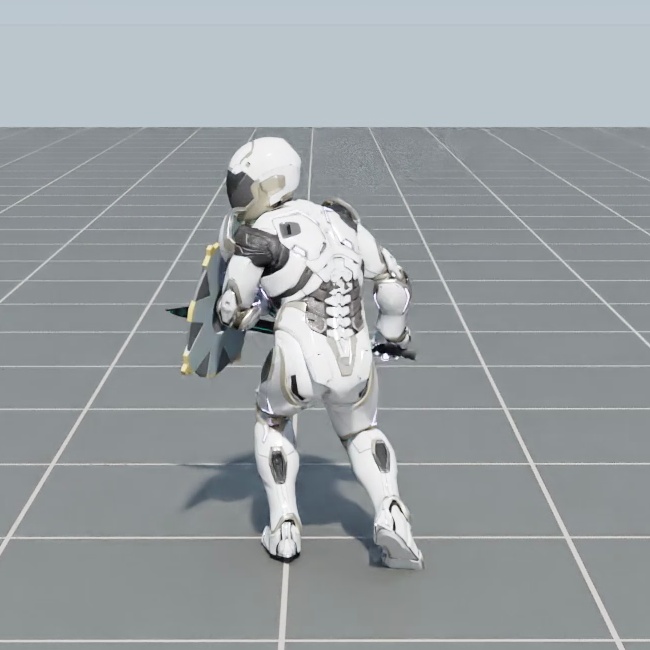}\hfill
         \includegraphics[width=0.333\textwidth]{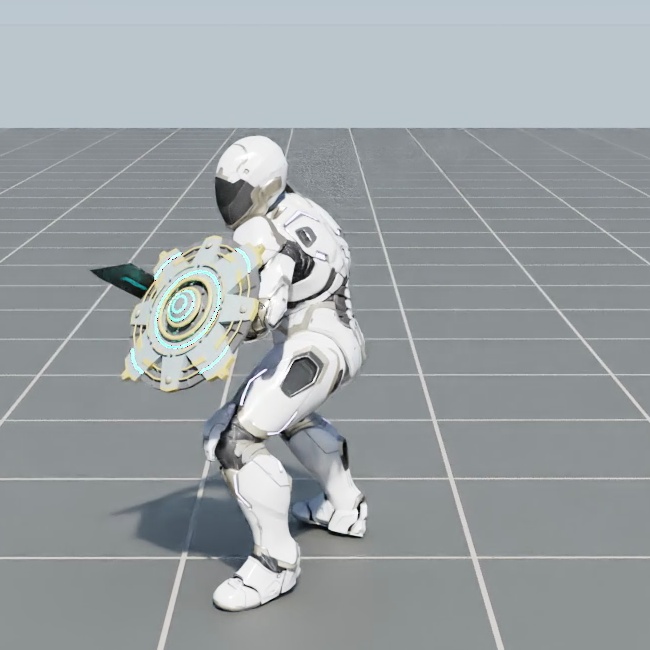}
         \caption{Turn left}
         \label{fig llc: apndx 12}
     \end{subfigure}\\
     \begin{subfigure}[b]{0.33\textwidth}
         \centering
         \includegraphics[width=0.333\textwidth]{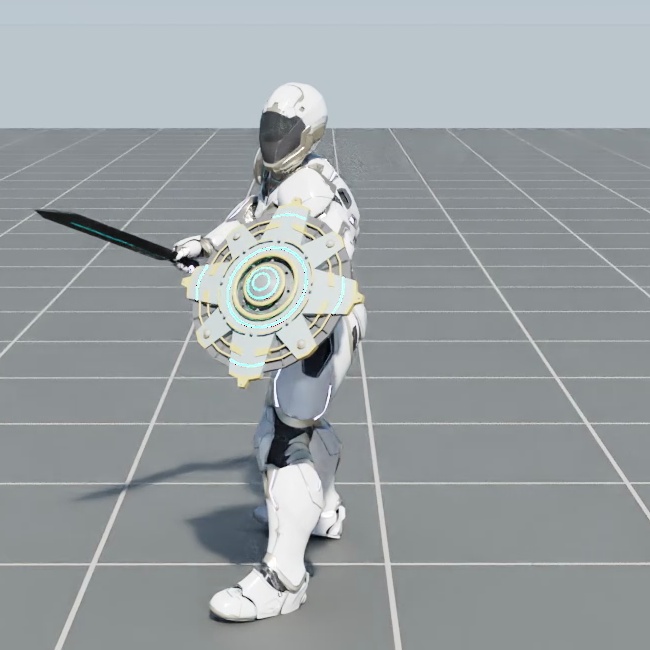}\hfill
         \includegraphics[width=0.333\textwidth]{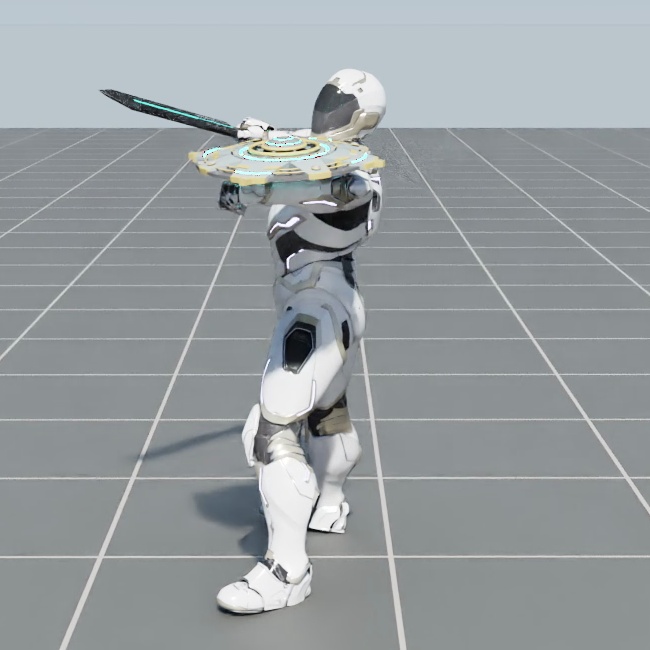}\hfill
         \includegraphics[width=0.333\textwidth]{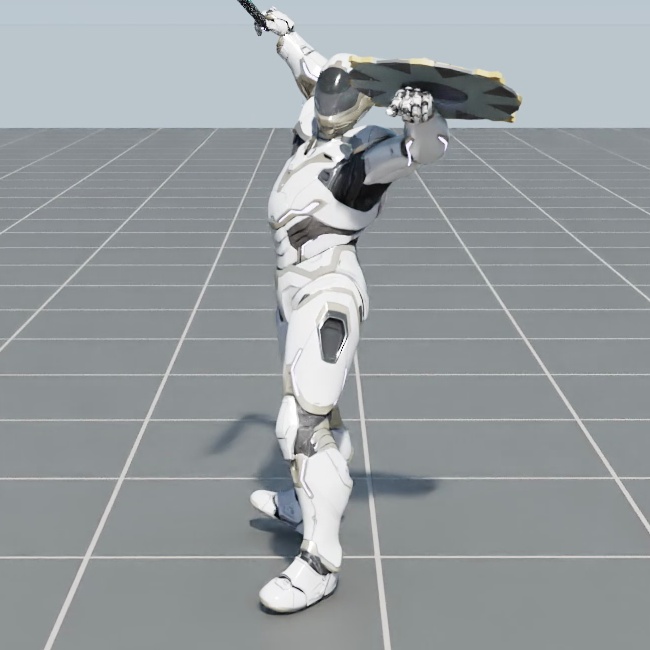}
         \caption{Roar, hands up}
         \label{fig llc: apndx 13}
     \end{subfigure}\hfill
     \begin{subfigure}[b]{0.33\textwidth}
         \centering
         \includegraphics[width=0.333\textwidth]{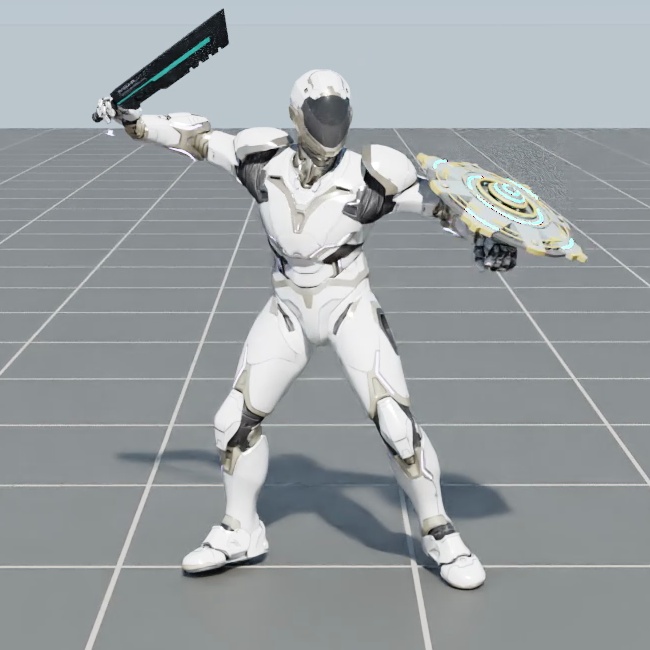}\hfill
         \includegraphics[width=0.333\textwidth]{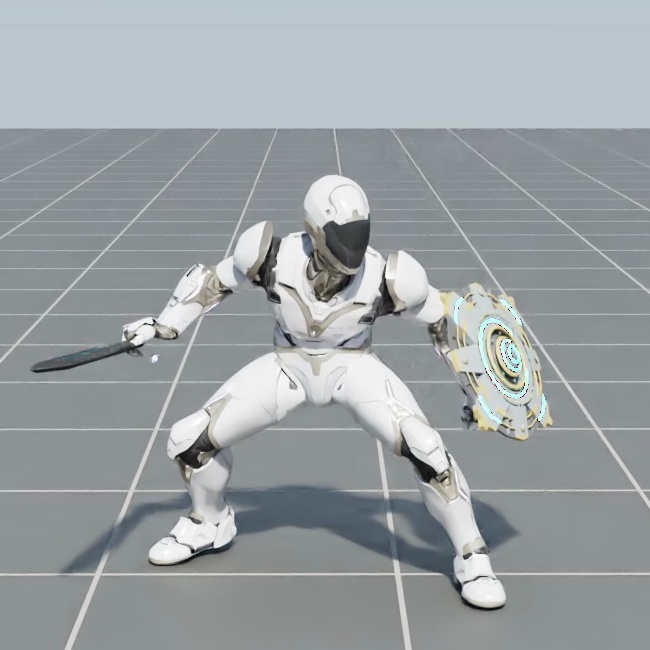}\hfill
         \includegraphics[width=0.333\textwidth]{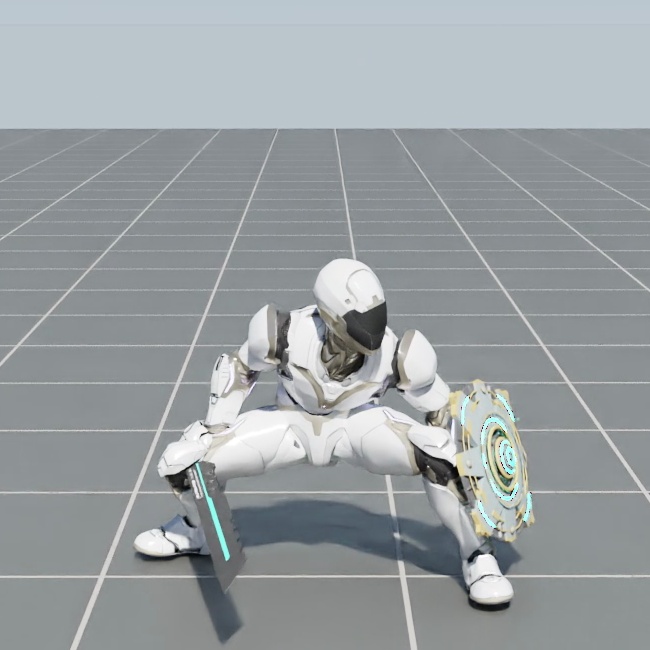}
         \caption{Crouch, idle}
         \label{fig llc: apndx 14}
     \end{subfigure}\hfill
     \begin{subfigure}[b]{0.33\textwidth}
         \centering
         \includegraphics[width=0.333\textwidth]{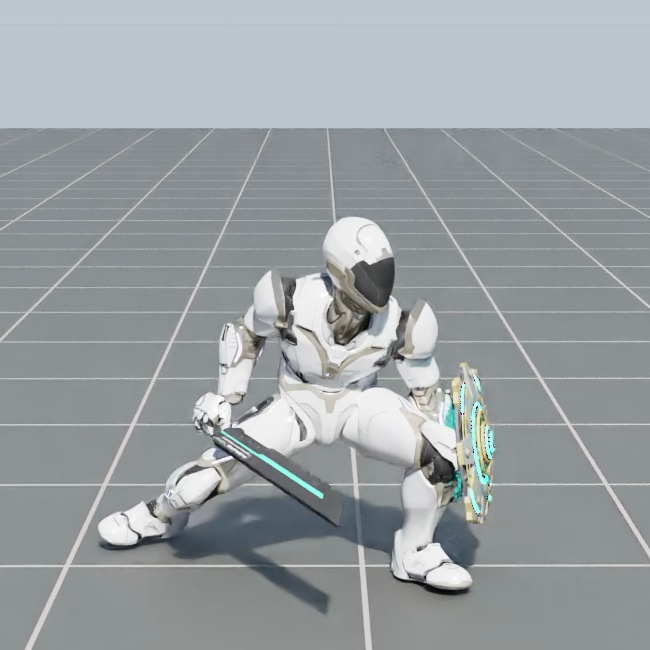}\hfill
         \includegraphics[width=0.333\textwidth]{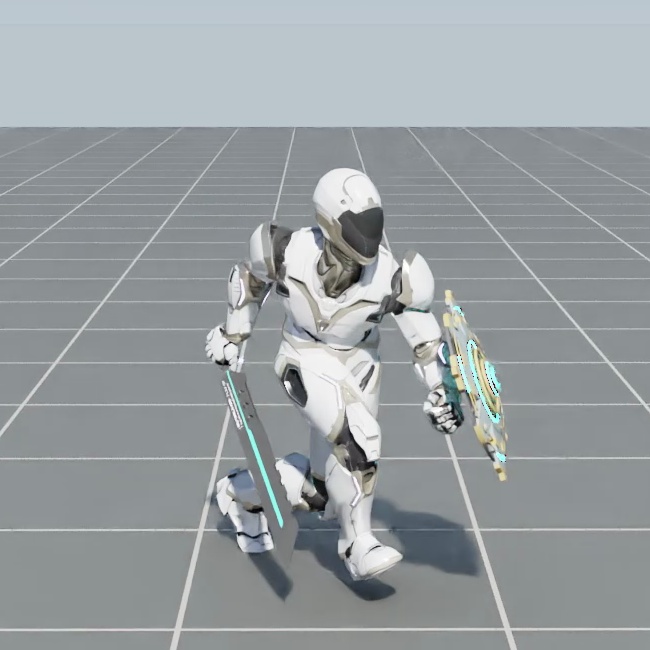}\hfill
         \includegraphics[width=0.333\textwidth]{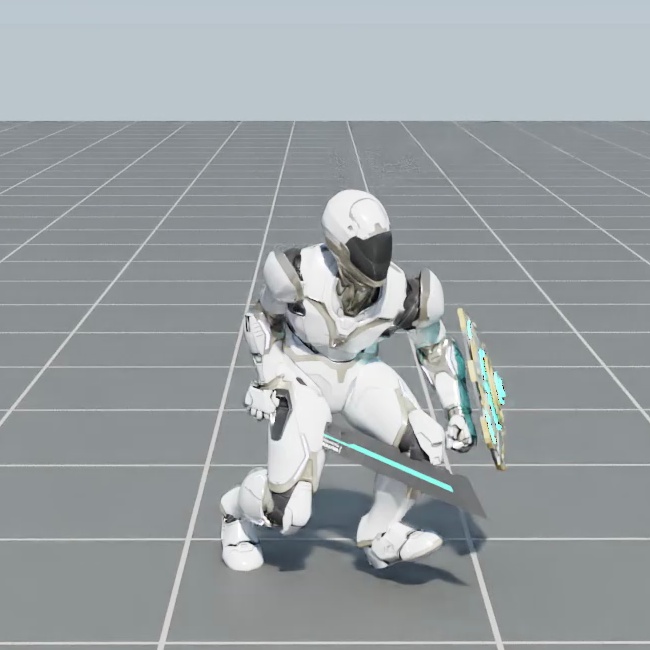}
         \caption{Crouch, walk}
         \label{fig llc: apndx 15}
     \end{subfigure}
    \caption{Skills generated by a low-level controller conditioned on the encoding of a demonstrated motion. Every 3 seconds, a new motion is sampled and the low-level policy is conditioned on the corresponding latent representation. These figures are generated from a single, long, episode during which the motions were performed sequentially.}
    \label{fig: motion transitioning}
\end{figure*}

As can be seen in \cref{fig: motion transitioning}, the agent learns to transition between complex motions.

\begin{figure*}[!ht]
    \centering
         \begin{subfigure}[b]{0.495\textwidth}
         \centering
         \includegraphics[width=0.245\textwidth]{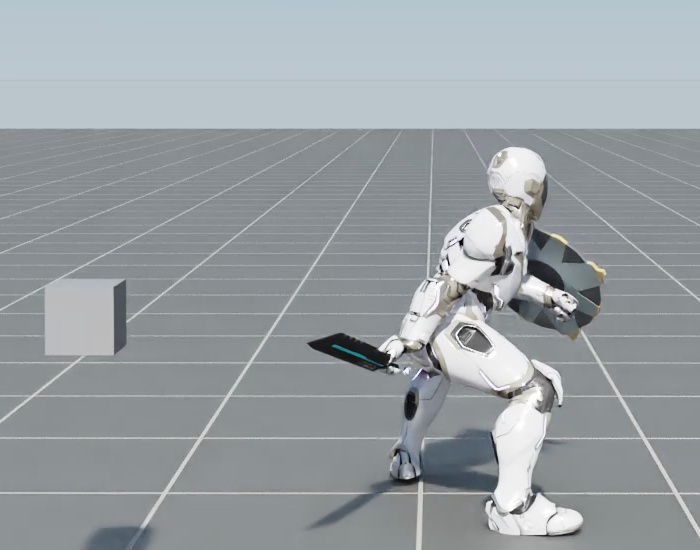}\hfill
         \includegraphics[width=0.245\textwidth]{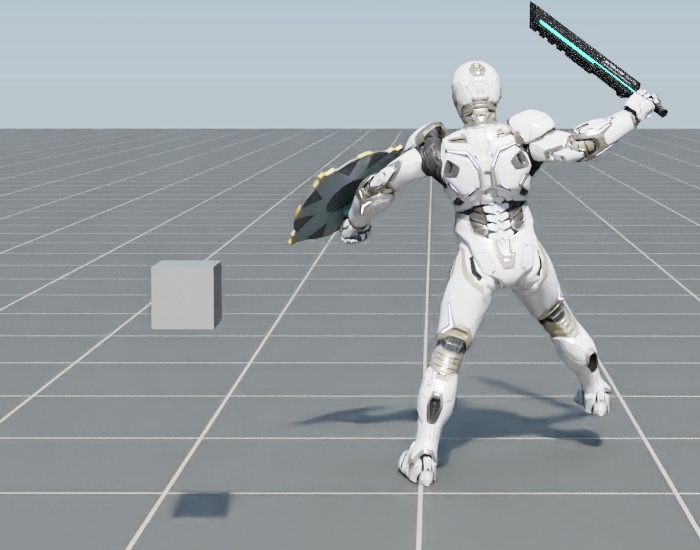}\hfill
         \includegraphics[width=0.245\textwidth]{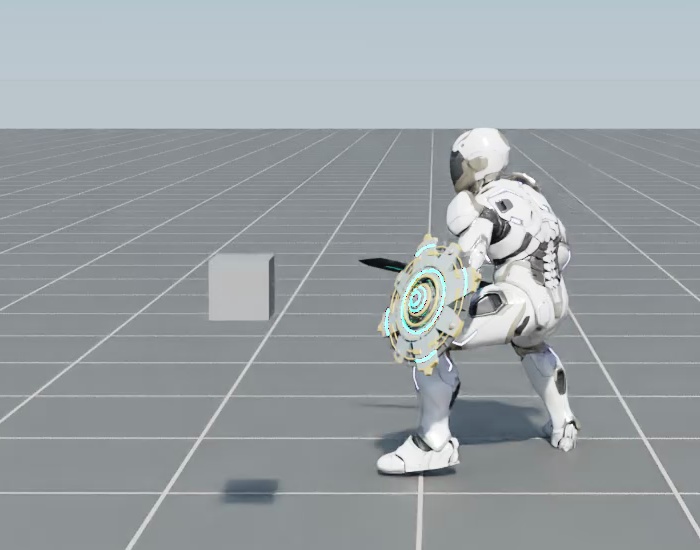}\hfill
         \includegraphics[width=0.245\textwidth]{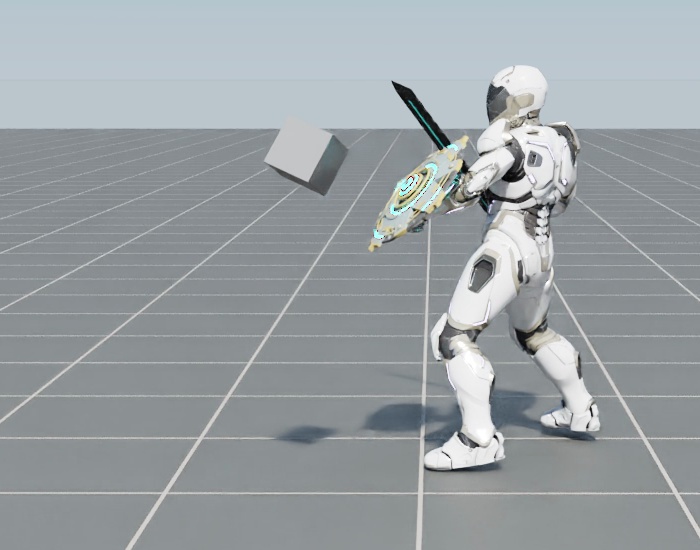}
         \caption{Block}
         \label{fig hrl: block}
     \end{subfigure}
     \hfill
     \begin{subfigure}[b]{0.495\textwidth}
         \centering
         \includegraphics[width=0.245\textwidth]{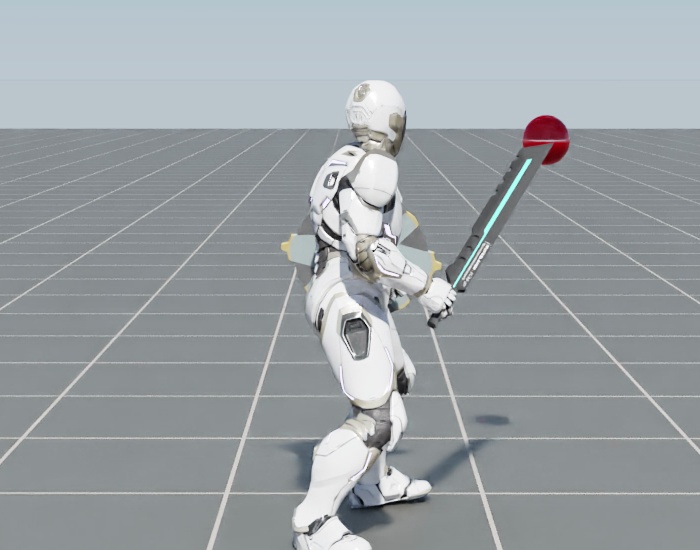}\hfill
         \includegraphics[width=0.245\textwidth]{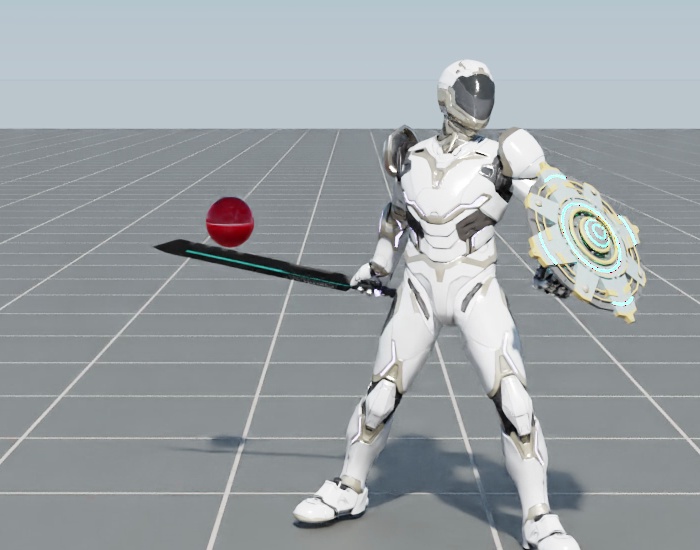}\hfill
         \includegraphics[width=0.245\textwidth]{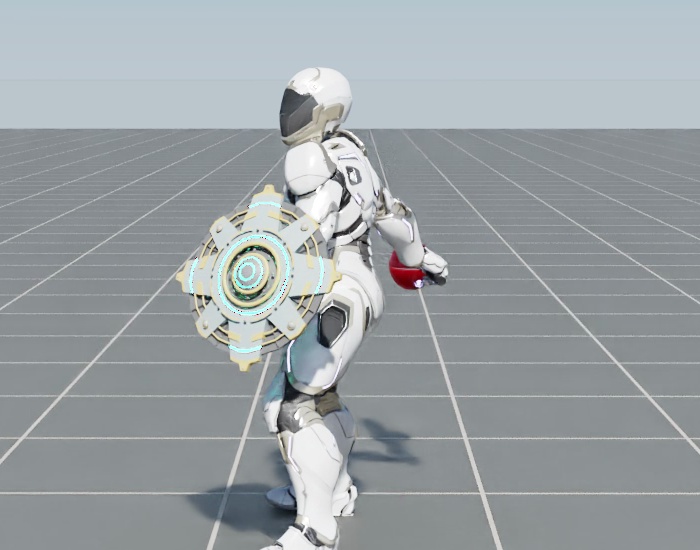}\hfill
         \includegraphics[width=0.245\textwidth]{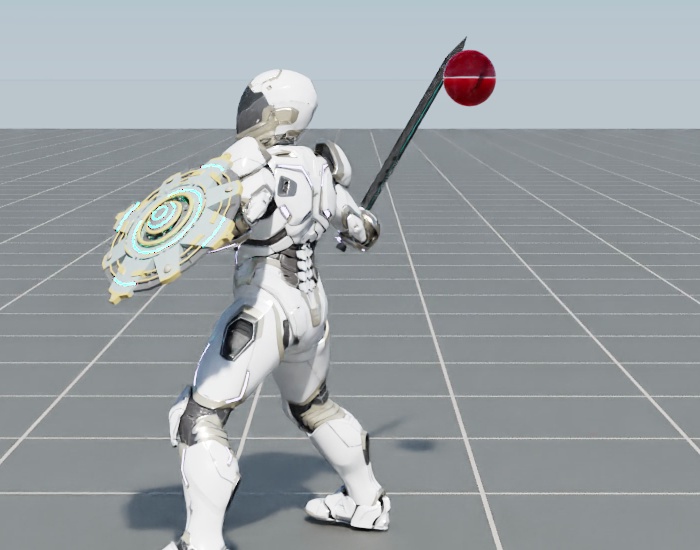}
         \caption{Reach}
         \label{fig hrl: reach}
     \end{subfigure}\\
    \begin{subfigure}[b]{0.498\textwidth}
         \centering
         \includegraphics[width=\textwidth]{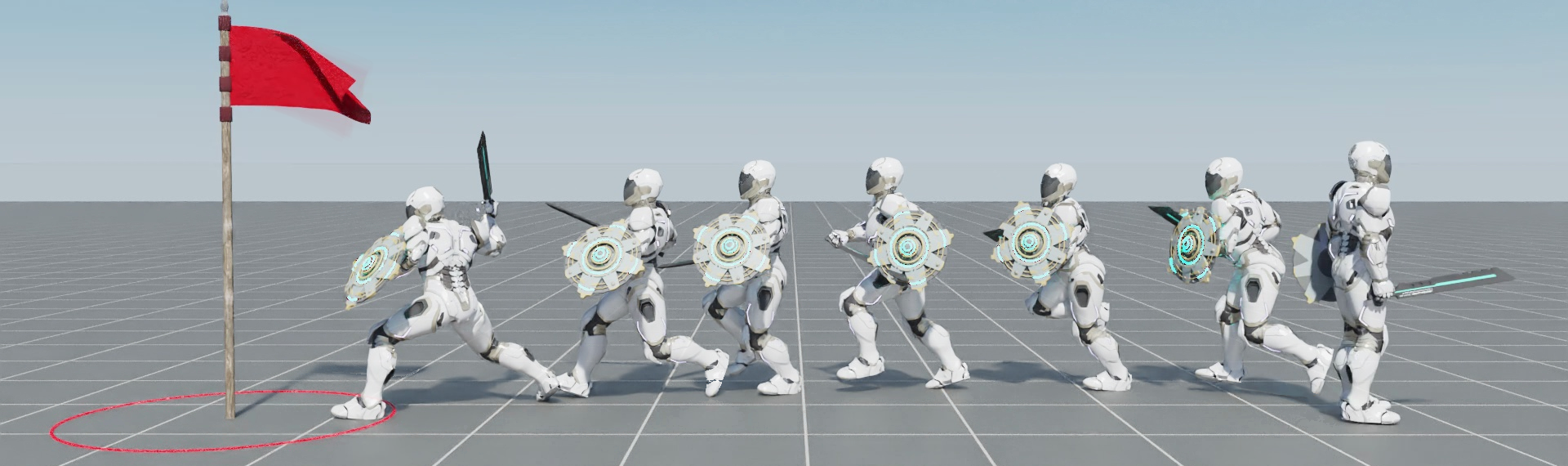}
         \caption{Location}
         \label{fig hrl-reg: location}
     \end{subfigure}
     \begin{subfigure}[b]{0.498\textwidth}
         \centering
         \includegraphics[width=\textwidth]{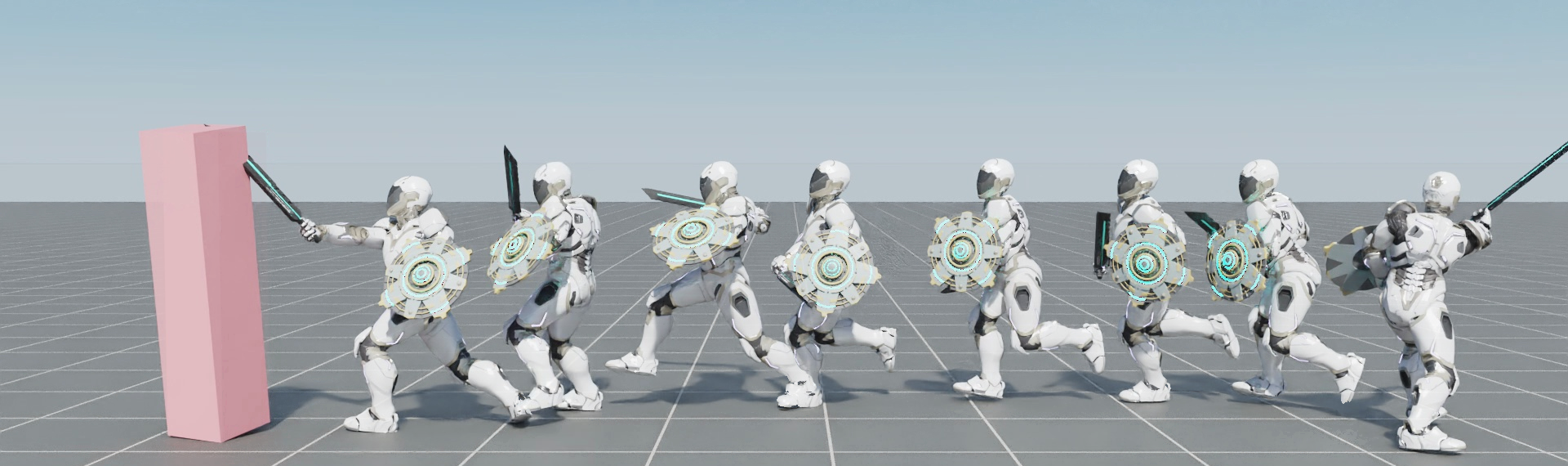}
         \caption{Strike}
         \label{fig hrl-reg: strike}
     \end{subfigure}
    \caption{Standard hierarchical task solving. A high-level policy directly controls the low-level policy and optimizes a task-specific objective.}
    \label{fig: downstream tasks}
\end{figure*}

\subsection{Downstream tasks}

In the paper we focused on the unique benefits of our method, namely, enabling the solution of unseen tasks without any additional training. Despite zero-shot applications being of great interest, sometimes an automated solution is needed.

In \cref{fig: downstream tasks} we present scenarios where the high-level policy is trained directly to solve downstream tasks without style conditioning. 

\subsubsection{Block} In the block task, a projectile is thrown at the agent and it is required to block it with the shield. To solve the task, the high-level policy is provided with a positive reward when a projectile hits its shield
\begin{equation}
    r_t^\text{block} = \mathbf{1} (\text{projectile blocked by shield}) \,.
\end{equation}

\subsubsection{Reach} Since the `Block` task tests control over the shield-bearing arm, we also showcase the `Reach` task, which requires precise sword motions. Here, the agent is tasked with bringing the tip of its sword to a specified $x^*$ location. To achieve this, it is provided a reward
\begin{equation}
    r_t^\text{reach} = \text{exp} \left(- 4 || x_t^\text{sword} - x^* ||_2^2 \right) \,,
\end{equation}
where $x_t^\text{sword}$ is the location of the tip of the sword at time t.

\subsubsection{Location} During `Block` and `Reach` the character remains rooted in the same spot, yet requires accurate low-level motions. To showcase the ability to compose motions directly from reward, as shown in \citet{peng2022ase}, we also train a `Location` and `Strike` task.

In the location task, the character needs to reach a target location. The high-level policy receives a reward
\begin{equation}
    r_t^\text{location} = \text{exp} \left( -0.5 || x^* - x_t^\text{root} ||^2 \right) \,,
\end{equation}
where $x^*$ is the goal location and $x_t^\text{root}$ is the location of the character's root. This reward urges the high-level controller to produce motions that reach the goal location as fast as possible. As seen in \cref{fig hrl-reg: location}, the controller learns to select latent variables $z_t$ that result in running.

\subsubsection{Strike} The `Strike` task tests the ability of the high-level controller to transition between motions within a complex multi-step task. Here, the agent is required to strike down a target. To do so, the high-level policy receives a reward
\begin{equation}
    r_t^\text{strike} = 1 - \mathbf{u}^\text{up} \cdot \mathbf{u}_t^* \,,
\end{equation}
where $\mathbf{u}^\text{up}$ is the global up vector, and $\mathbf{u}_t^*$ is the local up vector of the target object, expressed in the global coordinate frame. In addition, to prevent the character from just crashing into the target, \citet{peng2022ase} provides a termination condition in which the environment terminates if the character touches the target with any element that isn't the sword.

The reward urges to character to quickly reach the target and the termination makes sure it only does so by using the sword. As seen in \cref{fig hrl-reg: strike}, the character runs towards the target and then performs a sword strike to hit it.

\subsubsection{Tasks conclusions} Despite the low-level policy being trained on generating long-term behaviors, through iterative latent control, the high-level policy is capable of producing new motions to solve the task. For instance, to block the projectile it learns to control where the shield is aimed, and in the reach task, it learns to control the location of the sword and maintain a relatively static position around the goal location. These specific behaviors showcase the benefit of adversarial training schemes in their ability to generalize beyond what was observed in the reference dataset. Finally, when tasked with complex long horizon tasks, the character learns to utilize human-like motions while also transitioning naturally between them.

\subsubsection{FSM versus Reward design} In the paper we presented a way for solving tasks, such as location and strike, without performing task-specific training. This was done by leveraging an FSM design and the fact that the low-level policy can generate specified motions on demand. The benefit compared to reward design is clear. Learning to crouch-walk towards the target and then kick it, followed by a celebrative roaring motion -- requires a delicate reward and termination design. On the other hand, by utilizing the FSM approach, the character can be directed to perform specific motions in a specified direction, resulting in diverse solutions to tasks without reward design and without any task-specific training.

\section{Latent space analysis}

\begin{figure*}
    \centering
    \begin{subfigure}[b]{0.45\textwidth}
         \centering
         \includegraphics[width=\textwidth]{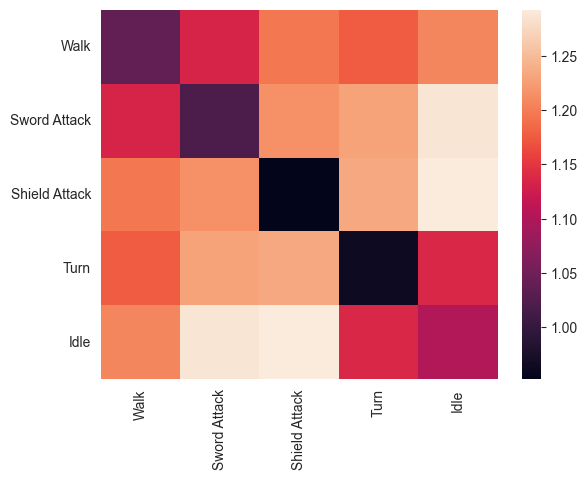}
         \caption{\alg}
         \label{fig latent: cat}
     \end{subfigure}
     \begin{subfigure}[b]{0.45\textwidth}
         \centering
         \includegraphics[width=\textwidth]{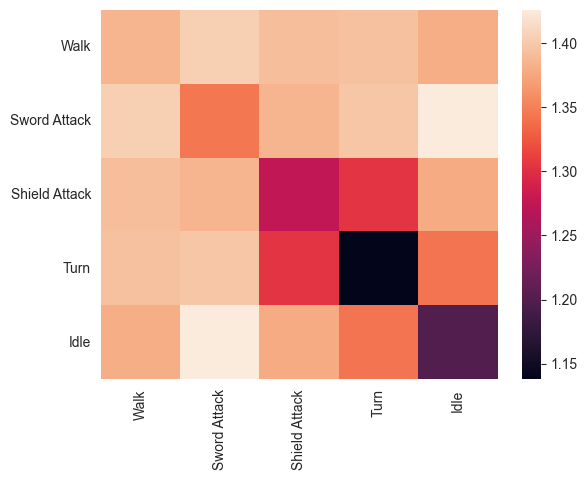}
         \caption{ASE}
         \label{fig latent: ase}
     \end{subfigure}
    \caption{Distance between motion classes within the latent space. Darker colors represent higher pairwise-latent-space proximity between the two motion classes.}
    \label{fig: latent}
\end{figure*}

To gain further insight into the learned representations, we analyze the structure of the latent space. We split the motions, roughly, into 5 categories:
\begin{itemize}
    \item Walking motions
    \item Sword attacks
    \item Shield attacks
    \item Turning motions
    \item and Idle motions
\end{itemize}
For each motion in the reference dataset, we encode it using each method's respective encoder. We then calculate the average pairwise distance between groups. As seen, in \cref{fig: latent}, \alg~ clusters motion groups closer in the latent space. This is observed by a lower distance along the diagonal. The meaning is that \alg~ learned a representation with stronger semantic meaning -- similar motions are clustered closely within the latent space.

Moreover, analyzing the distances within the ASE encodings shows that it does not maintain semantic relations between motion classes. Specifically, motions corresponding to walking are dispersed over the latent space.

In addition, we present a TSNE plot \cite{van2008visualizing} of the learned representation for \alg. While all the data is plotted, due to the vast number of motions (approximately 180), for clarity, we only label a subset.

\begin{figure*}
    \centering
    \includegraphics[width=0.4\linewidth]{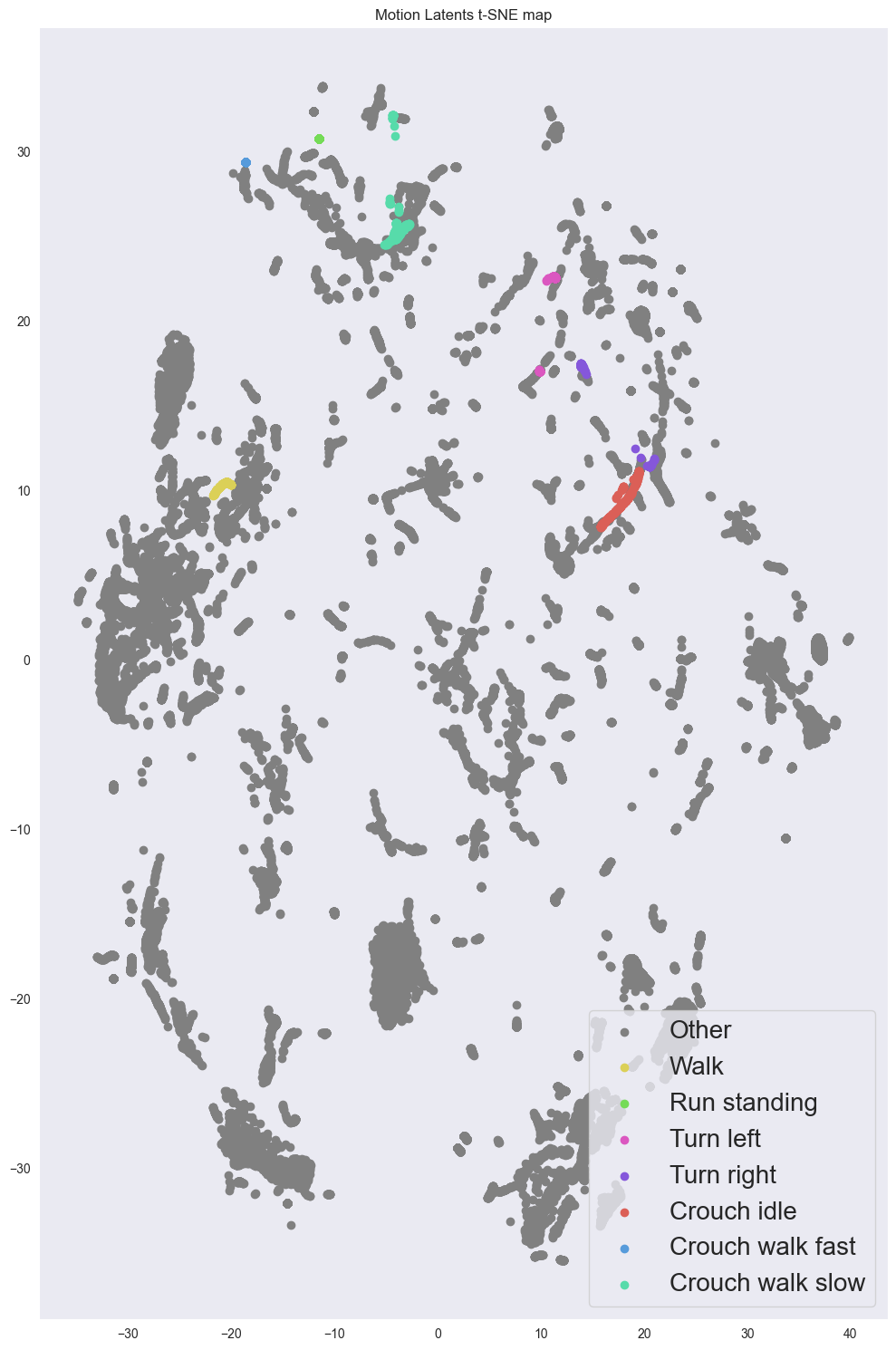}
    \caption{\alg~ TSNE analysis}
    \label{fig: cat tsne}
\end{figure*}

While TSNE does not preserve the global structure, as seen in \cref{fig: cat tsne}, it does suggest that sub-sequences from the same motion clip do indeed receive a similar representation, in the sense that they reside in proximity in the latent space.

\end{document}